\documentclass[11pt]{article}

\usepackage[final]{arxiv}
\usepackage{times}
\usepackage{latexsym}
\usepackage[T1]{fontenc}
\usepackage{inconsolata}
\usepackage{graphicx}
\usepackage{multirow}
\usepackage{url}
\usepackage{times}
\usepackage{latexsym}
\usepackage{colortbl}
\usepackage{xcolor}
\usepackage{threeparttable}
\usepackage[breakable]{tcolorbox}
\usepackage{enumitem}
\usepackage{cuted}
\usepackage{float}
\usepackage{makecell}
\usepackage[T1]{fontenc}
\usepackage{pifont}
\usepackage{amssymb}
\usepackage{wrapfig}

\usepackage{microtype}
\usepackage{tabularray}
\usepackage{graphicx}
\usepackage{subcaption}
\usepackage{booktabs}
\usepackage{amsmath}
\usepackage{inconsolata}

\usepackage{graphicx}
\usepackage{xltabular}  

\usepackage{arydshln}
\usepackage{marvosym}

\newcommand{\TheName}{\texttt{MIMIC}}

\title{The Imitation Game: When LLMs Learn to Reason Like Programs via Code-Centric Reasoning Data Synthesis}

\author{
    \textbf{Jinyang Zhang}\textsuperscript{1}, 
    \textbf{Weibin Liao}\textsuperscript{1},
    \textbf{Keqin Bao}\textsuperscript{2},
    \textbf{Sihang Li}\textsuperscript{2}, 
    \textbf{Shaobo Wang}\textsuperscript{3}, \\ 
    \textbf{Muyang Ye}\textsuperscript{4},
    \textbf{Hongxin Ding}\textsuperscript{1},
    \textbf{Yue Fang}\textsuperscript{1},
    \textbf{Tianyi Tang}\textsuperscript{2},
    \textbf{Fei Huang}\textsuperscript{2}, \\
    \textbf{Kexin Yang}\textsuperscript{2}\thanks{Corresponding author. Work done when Jinyang Zhang is intern at Qwen Team, Alibaba Group.},
    \textbf{Xingzhang Ren}\textsuperscript{2},
    \textbf{Dayiheng Liu}\textsuperscript{2}
    \\
    \textsuperscript{1}School of Computer Science, Peking University
    \\
    \textsuperscript{2}Qwen Team, Alibaba Group \ \ \ \ \
    \textsuperscript{3}EPIC Lab, SJTU
    \\
    \textsuperscript{4}College of Computer Science and Technology, Zhejiang University.
    \\
    \textrm{\Letter}~\texttt{jinyangzhang25@stu.pku.edu.cn}, \texttt{yangkexin.ykx@alibaba-inc.com}
}

\begin{document}
\maketitle
\begin{abstract}

Large Language Models (LLMs) excel at programming tasks but frequently fail at deterministic, fine-grained reasoning in natural language, relying heavily on semantic approximations rather than robust symbolic execution. To bridge this gap, we propose \TheName{}, a framework that leverages executable code as a rigorous medium for reasoning data synthesis. \TheName{} fundamentally transforms algorithms into verifiable reasoning trajectories through narrative fusion, code-guided test synthesis, and dynamic code instrumentation. Crucially, these explicit intermediate execution states naturally form a Code-Instrumented Reward (CIR), providing dense, high-fidelity process supervision for reinforcement learning without external reward models. Extensive evaluations reveal that models trained via SFT and GRPO on our synthesized dataset achieve substantial, consistent gains. Our method significantly elevates accuracy across general reasoning, complex mathematical benchmarks, and fine-grained deterministic tasks, demonstrating that the procedural rigor of executable code can effectively unlock and enhance the generalized reasoning capabilities of LLMs. Our code and data are available at \url{https://github.com/zjy1298/MIMIC}.

\end{abstract}

\section{Introduction}\label{sec:intro}
\begin{quote}
    \footnotesize
    \textit{``A person who is not good at calculating can still become a first-rate mathematician, whereas someone who has no sense for mathematics at all will at best become a great calculator.''}
    \hfill --- Novalis
\end{quote}

\begin{figure}[!t]
    \centering
    \includegraphics[width=\columnwidth]{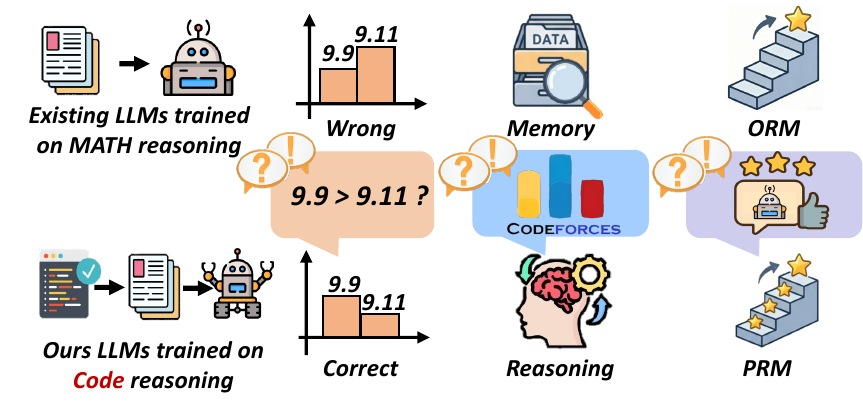}
    \caption{The paradox of LLMs: while they excel at complex programming logic, they often fail at basic deterministic natural language reasoning. \TheName{} bridges this gap by leveraging executable code to generate hallucination-mitigated reasoning trajectories.}
    \label{fig:teaser}
    \vspace{-3mm}
\end{figure}

Large Language Models (LLMs)~\cite{xu2025toward,liao2025tpo,zhang2026adept,ding20253ds} have demonstrated remarkable proficiency in formal environments, frequently surpassing human baselines on competitive programming benchmarks and code synthesis tasks~\cite{liu2024deepseek,hurst2024gpt,NOVA-63}. Yet, a stark paradox persists: the very same models that can author complex dynamic programming solutions routinely fail on natural language tasks requiring rudimentary, fine-grained symbolic fidelity~\cite{zhou2024ethical,yu2024breaking}. They struggle with deterministic operations that any trivial program executes flawlessly, such as comparing the numerical values $9.9$ and $9.11$~\cite{huang2026think,zhang2026tell}, tracking sequential state transitions, or counting the exact occurrences of a character in a word (as illustrated in Figure~\ref{fig:teaser}).

This asymmetry suggests that current ``reasoning'' capabilities in LLMs are largely driven by approximate semantic pattern-matching rather than internalized symbolic execution~\cite{dziri2023faith, gsm_sym}. While a dominant line of research attempts to enhance reasoning purely within the natural language domain (e.g., via chain-of-thought prompting or verifiers) to improve coding
  capabilities~\cite{codet, cot, self_debug, liao2026hyfunc}, the reverse direction---leveraging a model's robust \emph{coding} capabilities to bootstrap its \emph{general reasoning}---remains underexplored~\cite{yang2025code}. The underlying premise is that accepted, executable code inherently encapsulates the rigorous, step-by-step logical deduction that LLMs lack in prose. 

Unlike prior work that mainly studies how to transfer code or execution signals into reasoning, we argue that the surface form of competitive programming itself induces shortcut memorization, causing models to recall programming templates rather than perform transferable natural-language reasoning. This motivates MIMIC as a de-biasing framework: it removes programming-specific cues while preserving the underlying executable reasoning process.

However, translating competitive programming knowledge into generalized reasoning training data introduces significant challenges. Our preliminary investigations reveal that simply providing models with raw programming problems triggers a superficial ``recall illusion.''~\cite{janiak2025illusion,sun2026vlms} Instead of reasoning through the logic, models recognize the structural formatting of competitive programming tasks and attempt to reconstruct memorized code templates from their pre-training corpora, short-circuiting the actual reasoning process~\cite{guo2024learning,wang2026search}. Furthermore, standard data synthesis pipelines often rely on stronger LLMs to fabricate reasoning trajectories, which inevitably injects silent factual errors and computational hallucinations into the intermediate steps~\cite{gomez2022synthesis}.

This leads to our primary \textbf{research question}: \textit{How can we systematically distill the deterministic, procedural rigor of executable code into natural language reasoning trajectories, while simultaneously preventing models from bypassing reasoning via memorized coding patterns?}

To address this question, we propose \TheName{}, \textbf{M}ining \textbf{I}ntelligence from \textbf{M}achine-executed \textbf{I}nstrumented \textbf{C}ode, an end-to-end synthesis framework designed to bridge the ``Coding $\to$ Reasoning'' gap. \TheName{} guarantees computational correctness by strictly dividing labor: \textit{logic is articulated by the language model, but computation is guaranteed by deterministic code execution}. Specifically, our framework solves the aforementioned challenges through three core methodological innovations:

\begin{enumerate}
[leftmargin=*,itemsep=0pt,parsep=0.5em,topsep=0.3em,partopsep=0.3em]
    \item \textbf{Preventing Shortcut Memorization via Narrative Fusion:} To break the coding-problem surface patterns that trigger recall, \TheName{} removes all programmatic artifacts from source problems and fuses their core mathematical logic with diverse, real-world corpora. This transforms algorithmic challenges into self-contained, realistic natural-language exam questions.
    \item \textbf{Ensuring Generalizable Coverage via Code-Guided Synthesis:} Rather than relying on blind random sampling, \TheName{} directly analyzes the control-flow boundaries of the accepted code to synthesize targeted adversarial and guided-random test cases. This ensures that the generated reasoning scenarios probe deep logical branches rather than superficial scale.
    \item \textbf{Eliminating Hallucinations via Trace-Grounded CoT:} To generate verifiable reasoning trajectories, \TheName{} dynamically instruments the accepted code with semantic probes. Executing this instrumented code yields a deterministic sequence of intermediate states. The natural language chain-of-thought is then strictly anchored to these explicit execution traces, ensuring every quantitative claim is grounded in computational reality. These intermediate states further serve as a natural source of \emph{process-level reward signals} during reinforcement learning, providing denser supervision than binary outcome rewards without requiring a separate reward model.
\end{enumerate}

By solving these structural issues, \TheName{} produces high-quality, hallucination-mitigated reasoning data that systematically improves LLM performance across both general reasoning and mathematical benchmarks, demonstrating a profound bidirectional alignment between code understanding and natural-language reasoning.

In summary, our main contributions are threefold: 
\begin{itemize}[leftmargin=*,itemsep=0pt,parsep=0.5em,topsep=0.3em,partopsep=0.3em] 
    \item \textbf{Conceptual Insight:} We identify the critical gap between LLMs' strong formal coding capabilities and their failures in natural language reasoning. We expose the ``recall illusion'' and demonstrate how deterministic code execution can serve as a reliable anchor for true reasoning. 
    \item \textbf{Methodology:} We propose \TheName{} that translates code into hallucination-mitigated reasoning trajectories. Furthermore, we introduce a \textit{Code-Instrumented Reward} that leverages execution checkpoints to provide dense process-level supervision for reinforcement learning. 
    \item \textbf{Empirical Effectiveness:} Extensive evaluations demonstrate that both SFT and GRPO models trained on our synthesized data achieve significant, consistent gains across diverse mathematical, general reasoning, and fine-grained symbolic benchmarks. 
\end{itemize}

\section{Related Work}
\label{sec:related_work}

\paragraph{Synthetic reasoning data and verifiable-reward training.}
Recent work studies synthetic reasoning data through augmentation, large-scale curation, or problem generation, primarily in mathematical domains~\cite{MetaMath,WizardMath,numina_math_datasets,OpenMathInstruct,ScaleQuest,s1,LIMO}. Related efforts also construct verifiable logic and puzzle tasks with rule-based generators~\cite{synlogic,chen2026enigmata,ding2026evorubrics}. In parallel, reinforcement learning from verifiable rewards has become a promising recipe for reasoning, ranging from pure outcome-based RL to process supervision with reward models, human annotations, Monte Carlo estimation, or search~\cite{guo2025deepseek,SimpleRL,OpenReasonerZero,prmlessons,processbench,wang2024math,ding2026promed,liao2026learnat}. However, prior work typically relies on mathematical verifiers or separately trained process reward models, whereas we use programs as executable oracles that provide both gold outputs and verifiable intermediate signals.

\paragraph{Code and reasoning.}
A separate line of work uses code to support reasoning~\cite{yang2025code}, either as an inference-time tool~\cite{PAL,PoT,lv2024codeact}, as an evaluation substrate in programming benchmarks~\cite{humaneval,austin2021program_mbpp}, or as a source of input--output reasoning tasks~\cite{gu2024cruxeval,tear}. Recent works also use algorithmic problems, execution supervision, or synthetic verifiable data to improve reasoning~\cite{tear,tracepile,execgrounded,synlogic,chen2026enigmata}. \textbf{In contrast}, we do not use algorithmic problems in their original or lightly transformed forms as supervision. Instead, we transplant their executable logic into natural narratives, reducing programming-specific biases while preserving verifiable reasoning structure.

\paragraph{Reasoning failures on some trivial tasks.}
Recent benchmarks show that LLMs remain weak on tasks that are trivial for deterministic programs and human~\cite{malek2025frontier}, including basic string manipulation, character-level reasoning, and text-based maze navigation~\cite{wang2025stringllm,uzan2026charbench,dao2025alphamaze}. Several studies attribute these failures partly to tokenization-induced mismatches between symbolic units and model processing units~\cite{counting_tokenization,tokenization_arithmetic,why_count_letters}. \textbf{This gap motivates our setting}: we target such code-trivial but reasoning-fragile tasks and improve them with program-grounded supervision.

\section{Methodology}
\label{sec:methodology}

\begin{figure*}[t]
    \centering
    \includegraphics[width=\textwidth]{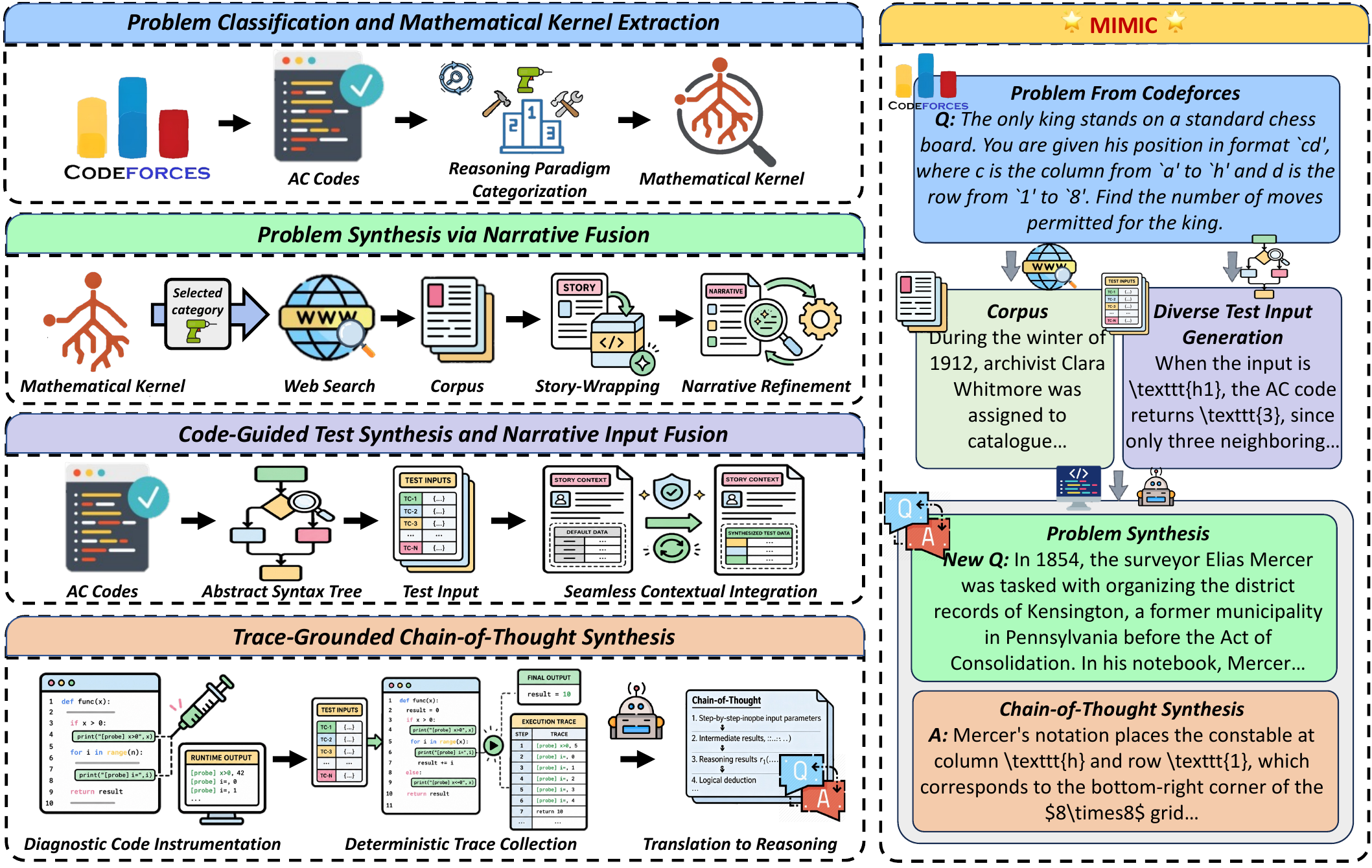}
    \caption{Overview of the \TheName{} framework. It transforms raw algorithmic problems into verifiable reasoning data through a four-stage pipeline: (1) Problem Classification and Mathematical Kernel Extraction, (2) Problem Synthesis via Narrative Fusion, (3) Code-Guided Test Synthesis and Narrative Input Fusion, and (4) Trace-Grounded Chain-of-Thought (CoT) Synthesis.}
    \label{fig:framework}
    \vspace{-3mm}
\end{figure*}

\subsection{Framework Overview and Notations}
We formally define the components of our proposed framework, \TheName{}. Let $\mathbb{P}$ denote the domain of raw competitive programming problems. Each problem $p \in \mathbb{P}$ is parameterized as a tuple $p = (\mathcal{S}_p, \mathcal{I}_p, \mathcal{O}_p, f_p)$, where $\mathcal{S}_p \in \Sigma$ represents the raw natural language problem statement within the universal textual space $\Sigma$, $\mathcal{I}_p$ is the valid input space, $\mathcal{O}_p$ is the target output space, and $f_p: \mathcal{I}_p \to \mathcal{O}_p \in \mathbb{F}$ denotes the accepted code (AC) representing the ground-truth deterministic execution mapping within the universal program space $\mathbb{F}$. To transform the raw algorithmic tuple $p$ into a robust, hallucination-mitigated set of natural-language reasoning pairs, \TheName{} executes a sequential, four-stage pipeline (see Figure~\ref{fig:framework}): (1) Problem Collection and Mathematical Kernel Extraction, (2) Problem Synthesis via Narrative Fusion, (3) Code-Guided Test Synthesis and Narrative Input Fusion, and (4) Trace-Grounded Chain-of-Thought (CoT) Synthesis. 

Structurally, this sequential pipeline directly resolves the three core challenges identified in Section \ref{sec:intro}. To prevent shortcut memorization, Stages 1 and 2 project the algorithmic kernel away from recognizable programmatic artifacts and fuse it into an external, real-world narrative corpus, effectively decorrelating the underlying mathematical logic from its surface phrasing. To ensure generalizable coverage, Stage 3 bypasses naive random sampling by introspecting the control-flow boundaries of the executable code $f_p$ to strategically generate targeted adversarial and guided-random test inputs. Finally, to eliminate hallucinations, Stage 4 dynamically instruments the accepted code to expose deterministic intermediate execution states, ensuring that every generated quantitative deduction is verifiably grounded in computational reality rather than autoregressive prior distributions.

\subsection{Stage 1: Problem Classification and Mathematical Kernel Extraction}
\paragraph{Reasoning Paradigm Categorization.} In the initial step, \TheName{} standardizes the raw problem $p \in \mathbb{P}$ by mapping it into a structured conceptual space. Let $\mathbb{T} = \{\tau_1, \tau_2, \dots, \tau_k\}$ denote a predefined taxonomy of reasoning paradigms (e.g., precise mathematical computation, strict state simulation). \TheName{} applies a classification operator $\Phi_{\text{cls}}: \mathbb{P} \to \mathbb{T}$ to assign an optimal reasoning category $t_p = \Phi_{\text{cls}}(p) \in \mathbb{T}$ to each problem, ensuring type-aware processing in all subsequent operations.

\paragraph{Algorithmic Kernel Isolation.} Concurrently, to eliminate contest-specific narrative framing, \TheName{} isolates the procedural semantics from the raw problem statement $\mathcal{S}_p$. We utilize a semantic refinement mapping $\Phi_{\text{ref}}: \Sigma \to \mathbb{K}$, where $\mathbb{K}$ is defined as the abstract space of concise mathematical kernels. The extracted kernel $K_p = \Phi_{\text{ref}}(\mathcal{S}_p)$ strictly preserves the algorithmic constraints, boundary conditions, and objective function operationalized by $f_p$, whilst discarding extraneous textual metadata. Stage 1 thus formally projects the raw tuple $p$ into a normalized semantic representation $(K_p, t_p, f_p)$. Further details on the taxonomy $\mathbb{T}$ and the extraction procedure are provided in Appendix~\ref{appendix:pipeline}.

\subsection{Stage 2: Problem Synthesis via Narrative Fusion}
\paragraph{Type-Aware Corpus Sampling.} To synthesize problems that authentically emulate real-world human-written exams without triggering memorized code templates, \TheName{} relies on an external, large-scale educational corpus $\mathbb{C}$. To maintain semantic coherence, \TheName{} partitions $\mathbb{C}$ into type-compatible subsets $\mathbb{C}_{\tau} \subset \mathbb{C}$ for each $\tau \in \mathbb{T}$. Given the problem type $t_p$, \TheName{} samples an appropriate context passage $c \sim \mathbb{C}_{t_p}$.

\paragraph{Story-Wrapping Generation.} Next, \TheName{} fuses the extracted kernel $K_p$ with the sampled context $c$. We formulate a story-wrapping generator $\Psi_{\text{wrap}}: \mathbb{K} \times \mathbb{C} \to \mathbb{Q}$, where $\mathbb{Q}$ is the space of natural-language exam questions. This generator integrates the abstract mathematical logic of $K_p$ into the narrative environment of $c$, producing a base story-wrapped question $\tilde{Q}_{p, c} = \Psi_{\text{wrap}}(K_p, c)$. This initial projection inherently embeds a default valid input $\tilde{x} \in \mathcal{I}_p$ implicitly within its prose.

\paragraph{Narrative Refinement Optimization.} \TheName{} applies a narrative smoothing function $\Psi_{\text{smooth}}: \mathbb{Q} \to \mathbb{Q}$ to improve continuity between the retrieved context and problem formulation, producing the refined question template $Q_{p, c}^* = \Psi_{\text{smooth}}(\tilde{Q}_{p, c})$. Details are provided in Appendix~\ref{appendix:stage2}.

\subsection{Stage 3: Code-Guided Test Synthesis and Narrative Input Fusion}
\paragraph{Control-Flow Introspection.} Relying solely on the embedded default input $\tilde{x}$ yields insufficient coverage over the logical domain. To ensure robust generalizability, \TheName{} systematically analyzes the deterministic algorithmic structure of $f_p$. Let $\mathcal{D}(f_p)$ denote the set of salient control-flow decision points (e.g., branch conditions, loop terminations, and variable configurations) identified within the abstract syntax tree of $f_p$.

\paragraph{Targeted Input Synthesis.} Using these decision points, \TheName{} defines an input synthesis operator $\Gamma_{\text{test}}: \mathcal{D}(f_p) \to 2^{\mathcal{I}_p}$. It systematically generates a diverse set of test configurations $\mathcal{X}_p = \{x_1, x_2, \dots, x_N\} \subset \mathcal{I}_p$, designed to stress distinct execution paths through a mixture of adversarial boundary cases and guided-random inputs.

\paragraph{Seamless Contextual Integration.} For each generated input $x_i \in \mathcal{X}_p$, \TheName{} must update the associated narrative without violating the underlying story context. We formulate a narrative input fusion function $\Omega_{\text{fuse}}: \mathbb{Q} \times \mathcal{I}_p \times \mathcal{I}_p \to \mathbb{Q}$, which structurally replaces the original default parameters $\tilde{x}$ within $Q_{p, c}^*$ with the newly synthesized input $x_i$. This mapping yields an instantiated, input-specific natural language question $Q_{p, c}^{(i)} = \Omega_{\text{fuse}}(Q_{p, c}^*, \tilde{x}, x_i)$. The comprehensive methodology for decision-point extraction and narrative input fusion is detailed in Appendix \ref{appendix:stage3}.

\subsection{Stage 4: Trace-Grounded Chain-of-Thought Synthesis}
\paragraph{Diagnostic Code Instrumentation.} In the final stage, \TheName{} aims to generate hallucination-mitigated reasoning trajectories. Rather than relying on autoregressive prior distributions to hallucinate intermediate steps, \TheName{} structurally instruments the accepted code $f_p$. Let $\Lambda_{\text{inst}}: \mathbb{F} \to \mathbb{F}^*$ be a read-only instrumentation mapping that injects sparse diagnostic probes at semantically meaningful breakpoints within $f_p$, yielding the instrumented program $f^*_p = \Lambda_{\text{inst}}(f_p)$.

\paragraph{Deterministic Trace Collection.} For each instantiated input $x_i \in \mathcal{X}_p$, executing the instrumented program $f^*_p(x_i)$ deterministically produces both the ground-truth output $y_i \in \mathcal{O}_p$ (where $y_i = f_p(x_i)$) and a corresponding sequence of intermediate execution states $E_i = (s_1, s_2, \dots, s_m) \in \mathbb{E}$, where $\mathbb{E}$ is the sequential space of execution traces.

\paragraph{Procedural Translation to Reasoning.} Finally, \TheName{} utilizes a reasoning synthesis function $\Theta_{\text{CoT}}: \mathbb{Q} \times \mathbb{E} \times \mathcal{O}_p \to \mathbb{R}$ which translates the procedural scaffold $E_i$ into a natural-language chain-of-thought $R_{p, c}^{(i)} \in \mathbb{R}$. The resulting reasoning sequence $R_{p, c}^{(i)}$ bridges the premise $Q_{p, c}^{(i)}$ and the conclusion $y_i$, ensuring that every deductive step is grounded in computational reality. The final training instance is the verified reasoning pair $(Q_{p, c}^{(i)}, R_{p, c}^{(i)})$. Details on trace collection and CoT generation are provided in Appendix~\ref{appendix:stage4}.

\section{Experiment}
\label{sec:experiment}

\newcommand{\up}[1]{\textcolor{green!60!black}{\scriptsize$\uparrow$\!#1}}
\newcommand{\dn}[1]{\textcolor{red}{\scriptsize$\downarrow$\!#1}}

In this section, we conduct extensive experiments to answer the following research questions (RQs):
\begin{itemize}[leftmargin=*,itemsep=0pt,parsep=0.5em,topsep=0.3em,partopsep=0.3em]
    \item \textbf{RQ1:} Can our method enhance LLMs' reasoning capabilities, and does it generalize to mathematical and other general reasoning tasks?
    \item \textbf{RQ2:} Can our method effectively solve fine-grained, rule-based deterministic problems (e.g., character-level perception or spatial reasoning)?
    \item \textbf{RQ3:} Does our data synthesis pipeline genuinely elicit reasoning rather than triggering superficial pattern matching or memorized recall?
    \item \textbf{RQ4:} Is our code-instrumented process reward effective for reinforcing reasoning?
\end{itemize}

\begin{table*}[t]
\centering
\resizebox{\textwidth}{!}{
\begin{tabular}{@{}ll ccccccc ccccc cccc @{}}
\toprule
\multirow{2}{*}{\textbf{Model}} & \multirow{2}{*}{\textbf{Setting}}
  & \multicolumn{7}{c}{\textbf{General Reasoning}}
  & \multicolumn{5}{c}{\textbf{Math}}
  & \multicolumn{4}{c}{\textbf{Code}} \\
\cmidrule(lr){3-9} \cmidrule(lr){10-14} \cmidrule(lr){15-18}
& & ARC & BBH & MIMIC & GPQA & MMLU & DROP & \cellcolor{blue!8}\textbf{Avg}
  & GSM8K & GSM+ & MATH & AIME & \cellcolor{blue!8}\textbf{Avg}
  & HEval & CruxE & MBPP & \cellcolor{blue!8}\textbf{Avg} \\
\midrule
\multicolumn{18}{c}{\textit{(A) Main Results: Baseline $\rightarrow$ Ours (SFT) / Ours (GRPO)}} \\
\midrule
\multirow{3}{*}{Qwen3-4B}
 & Base             & 79.6 & 76.0 & 40.6 & 40.1 & 66.3 & 50.5 & \cellcolor{blue!8}58.9 & 89.8 & 73.2 & 78.3 & 15.6 & \cellcolor{blue!8}64.2 & 31.7 & 69.3 & 39.7 & \cellcolor{blue!8}46.9 \\
 & Ours (SFT)       & \textbf{85.9} & 77.0 & 43.5 & 40.4 & 67.3 & 55.5 & \cellcolor{blue!8}61.6\up{2.7} & \textbf{91.0} & 74.9 & 77.6 & 16.7 & \cellcolor{blue!8}65.0\up{0.8} & \textbf{32.1} & 70.8 & 38.8 & \cellcolor{blue!8}47.2\up{0.3} \\
 & Ours (GRPO)      & 80.6 & \textbf{77.9} & \textbf{51.3} & \textbf{40.8} & \textbf{67.5} & \textbf{58.3} & \cellcolor{blue!8}\textbf{62.7}\up{3.8} & \textbf{91.0} & \textbf{76.4} & \textbf{80.3} & \textbf{28.3} & \cellcolor{blue!8}\textbf{69.0}\up{4.8} & 31.7 & \textbf{72.7} & \textbf{43.6} & \cellcolor{blue!8}\textbf{49.3}\up{2.4} \\
\midrule
\multirow{3}{*}{Qwen3-8B}
 & Base             & 79.8 & 78.8 & 41.2 & \textbf{44.4} & 70.5 & 46.6 & \cellcolor{blue!8}60.2 & 90.6 & 73.7 & 79.2 & 23.3 & \cellcolor{blue!8}66.7 & 22.6 & 70.6 & 40.5 & \cellcolor{blue!8}44.6 \\
 & Ours (SFT)       & \textbf{85.3} & 78.5 & 44.5 & 42.0 & 69.5 & 50.4 & \cellcolor{blue!8}61.7\up{1.5} & 90.1 & 74.5 & 80.2 & \textbf{31.7} & \cellcolor{blue!8}69.1\up{2.4} & \textbf{27.4} & 72.8 & 34.2 & \cellcolor{blue!8}44.8\up{0.2} \\
 & Ours (GRPO)      & 81.2 & \textbf{81.7} & \textbf{52.2} & 42.6 & \textbf{72.2} & \textbf{53.1} & \cellcolor{blue!8}\textbf{63.8}\up{3.6} & \textbf{91.7} & \textbf{78.1} & \textbf{81.0} & 26.7 & \cellcolor{blue!8}\textbf{69.4}\up{2.7} & 23.8 & \textbf{76.3} & \textbf{44.0} & \cellcolor{blue!8}\textbf{48.0}\up{3.4} \\
\midrule
\multirow{3}{*}{\shortstack[l]{DeepSeek-R1\\-Distill-8B}}
 & Base             & 83.0 & 69.1 & 38.3 & 24.0 & 52.8 & 18.1 & \cellcolor{blue!8}47.6 & 82.5 & 60.3 & 58.4 & 16.7 & \cellcolor{blue!8}54.5 & 22.6 & 44.8 & 17.5 & \cellcolor{blue!8}28.3 \\
 & Ours (SFT)       & \textbf{86.1} & \textbf{75.7} & 40.8 & 25.2 & 55.8 & \textbf{23.6} & \cellcolor{blue!8}\textbf{51.2}\up{3.6} & \textbf{88.5} & \textbf{70.4} & 66.2 & \textbf{26.7} & \cellcolor{blue!8}63.0\up{8.5} & 23.9 & 38.9 & \textbf{24.8} & \cellcolor{blue!8}29.2\up{0.9} \\
 & Ours (GRPO)      & 83.5 & 68.5 & \textbf{43.5} & \textbf{27.7} & \textbf{59.3} & 19.1 & \cellcolor{blue!8}50.2\up{2.6} & 88.4 & 66.7 & \textbf{78.2} & 23.3 & \cellcolor{blue!8}\textbf{64.2}\up{9.7} & \textbf{26.8} & \textbf{48.3} & 21.8 & \cellcolor{blue!8}\textbf{32.3}\up{4.0} \\
\midrule
\multicolumn{18}{c}{\textit{(B) Synthesis Ablation: Ours (Contextualized) vs.\ Raw Codeforces}} \\
\midrule
\multirow{4}{*}{Qwen3-4B}
 & Ours (SFT)         & \textbf{85.9} & \textbf{77.0} & \textbf{43.5} & \textbf{40.4} & \textbf{67.3} & \textbf{55.5} & \cellcolor{blue!8}\textbf{61.6} & \textbf{91.0} & \textbf{74.9} & \textbf{77.6} & \textbf{16.7} & \cellcolor{blue!8}\textbf{65.0} & 32.1 & \textbf{70.8} & \textbf{35.8} & \cellcolor{blue!8}\textbf{46.2} \\
 & Codeforces (SFT)   & 85.2 & 43.9 & 28.9 & 25.4 & 44.8 & 42.6 & \cellcolor{blue!8}45.1\dn{16.5} & 71.5 & 54.9 & 40.4 & 6.7 & \cellcolor{blue!8}43.4\dn{21.6} & \textbf{34.2} & 65.6 & 34.6 & \cellcolor{blue!8}44.8\dn{1.4} \\
\cmidrule(lr){2-18}
 & Ours (GRPO)        & \textbf{80.6} & \textbf{77.9} & \textbf{51.3} & \textbf{40.8} & \textbf{67.5} & \textbf{58.3} & \cellcolor{blue!8}\textbf{62.7} & \textbf{91.0} & \textbf{76.4} & \textbf{80.3} & \textbf{28.3} & \cellcolor{blue!8}\textbf{69.0} & \textbf{31.7} & \textbf{72.7} & \textbf{43.6} & \cellcolor{blue!8}\textbf{49.3} \\
 & Codeforces (GRPO)  & 80.1 & 75.1 & 45.1 & 32.4 & 65.1 & 41.2 & \cellcolor{blue!8}56.5\dn{6.2} & 89.4 & 75.6 & 75.5 & 20.0 & \cellcolor{blue!8}65.1\dn{3.9} & 25.6 & 66.8 & 33.9 & \cellcolor{blue!8}42.1\dn{7.2} \\
\midrule
\multirow{4}{*}{Qwen3-8B}
 & Ours (SFT)         & \textbf{85.3} & \textbf{78.5} & \textbf{44.5} & \textbf{42.0} & \textbf{69.5} & \textbf{50.4} & \cellcolor{blue!8}\textbf{61.7} & \textbf{91.6} & \textbf{74.5} & \textbf{80.2} & \textbf{32.7} & \cellcolor{blue!8}\textbf{67.0} & 27.4 & 72.8 & 34.2 & \cellcolor{blue!8}44.8 \\
 & Codeforces (SFT)   & 81.7 & 44.1 & 28.7 & 24.7 & 39.4 & 31.3 & \cellcolor{blue!8}41.7\dn{20.0} & 74.2 & 57.1 & 41.9 & 6.7 & \cellcolor{blue!8}45.0\dn{22.0} & \textbf{32.3} & \textbf{72.8} & \textbf{35.8} & \cellcolor{blue!8}\textbf{47.0}\up{2.2} \\
\cmidrule(lr){2-18}
 & Ours (GRPO)        & \textbf{81.2} & \textbf{81.7} & \textbf{52.2} & \textbf{42.6} & \textbf{72.2} & \textbf{53.1} & \cellcolor{blue!8}\textbf{63.8} & \textbf{91.7} & \textbf{78.1} & 81.0 & \textbf{26.7} & \cellcolor{blue!8}\textbf{69.4} & \textbf{23.8} & \textbf{76.3} & \textbf{44.0} & \cellcolor{blue!8}\textbf{48.0} \\
 & Codeforces (GRPO)  & 79.4 & 81.1 & 45.8 & 40.2 & 67.3 & 36.7 & \cellcolor{blue!8}58.4\dn{5.4} & \textbf{91.7} & 76.5 & \textbf{81.2} & 20.0 & \cellcolor{blue!8}67.3\dn{2.1} & 21.3 & 76.0 & 42.4 & \cellcolor{blue!8}46.6\dn{1.4} \\
\midrule
\multirow{4}{*}{\shortstack[l]{DeepSeek-R1\\-Distill-8B}}
 & Ours (SFT)         & \textbf{86.1} & \textbf{75.7} & \textbf{40.8} & \textbf{25.2} & \textbf{55.8} & \textbf{23.6} & \cellcolor{blue!8}\textbf{51.2} & \textbf{88.5} & \textbf{70.4} & \textbf{66.2} & \textbf{26.7} & \cellcolor{blue!8}\textbf{63.0} & 23.9 & \textbf{38.9} & \textbf{24.8} & \cellcolor{blue!8}29.2 \\
 & Codeforces (SFT)   & 83.8 & 43.8 & 36.6 & 21.5 & 43.0 & 22.7 & \cellcolor{blue!8}41.9\dn{9.3} & 62.0 & 49.1 & 55.4 & 23.3 & \cellcolor{blue!8}47.5\dn{15.5} & \textbf{27.4} & 35.1 & 14.8 & \cellcolor{blue!8}25.8\dn{3.4} \\
\cmidrule(lr){2-18}
 & Ours (GRPO)        & \textbf{83.5} & \textbf{68.5} & \textbf{43.5} & \textbf{27.7} & \textbf{59.3} & \textbf{19.1} & \cellcolor{blue!8}\textbf{50.2} & \textbf{88.4} & 66.7 & \textbf{78.2} & \textbf{23.3} & \cellcolor{blue!8}\textbf{64.2} & \textbf{26.8} & \textbf{48.3} & \textbf{21.8} & \cellcolor{blue!8}\textbf{32.3} \\
 & Codeforces (GRPO)  & 80.8 & 61.2 & 41.1 & 24.3 & 57.8 & 18.7 & \cellcolor{blue!8}47.3\dn{2.9} & 74.8 & \textbf{68.1} & 74.7 & 15.0 & \cellcolor{blue!8}58.2\dn{6.0} & \textbf{26.8} & 38.4 & 16.7 & \cellcolor{blue!8}27.3\dn{5.0} \\
\midrule
\multicolumn{18}{c}{\textit{(C) Cross-Domain Mixing: Composability with Mathematical Data}} \\
\midrule
\multirow{5}{*}{Qwen3-4B}
 & Math (SFT)                     & 85.2 & \textbf{77.6} & 39.4 & 38.3 & 66.3 & 51.3 & \cellcolor{blue!8}59.7 & 90.2 & \textbf{75.3} & 75.9 & \textbf{16.7} & \cellcolor{blue!8}64.5 & \textbf{34.2} & 69.5 & 31.1 & \cellcolor{blue!8}44.9 \\
 & Ours (SFT)                     & \textbf{85.9} & 77.0 & \textbf{43.5} & \textbf{40.4} & \textbf{67.3} & \textbf{55.5} & \cellcolor{blue!8}\textbf{61.6}\up{1.9} & \textbf{91.0} & 74.9 & \textbf{77.6} & \textbf{16.7} & \cellcolor{blue!8}\textbf{65.0}\up{0.5} & 32.1 & \textbf{70.8} & \textbf{35.8} & \cellcolor{blue!8}\textbf{46.2}\up{1.3} \\
\cmidrule(lr){2-18}
 & $\mathcal{D}_\text{math}$ (GRPO) & 80.2 & 77.8 & 41.4 & 38.6 & \textbf{68.7} & 39.9 & \cellcolor{blue!8}57.8 & 91.7 & 76.6 & \textbf{81.7} & 23.3 & \cellcolor{blue!8}68.3 & 27.4 & 70.5 & 39.3 & \cellcolor{blue!8}45.8 \\
 & Ours (GRPO)                    & \textbf{80.6} & \textbf{77.9} & \textbf{51.3} & \textbf{40.8} & 67.5 & \textbf{58.3} & \cellcolor{blue!8}\textbf{62.7}\up{4.9} & 91.0 & 76.4 & 80.3 & 28.3 & \cellcolor{blue!8}69.0\up{0.7} & \textbf{31.7} & 72.7 & \textbf{43.6} & \cellcolor{blue!8}\textbf{49.3}\up{3.5} \\
 & $\mathcal{D}_\text{mixed}$ (GRPO) & 79.3 & 77.6 & 51.0 & 39.1 & 67.8 & 41.6 & \cellcolor{blue!8}59.4\up{1.6} & \textbf{93.2} & \textbf{76.7} & 78.6 & \textbf{30.0} & \cellcolor{blue!8}\textbf{69.6}\up{1.3} & 27.4 & \textbf{75.7} & 39.9 & \cellcolor{blue!8}47.7\up{1.9} \\
\cmidrule(lr){1-18}
\multirow{5}{*}{Qwen3-8B}
 & Math (SFT)                     & 84.0 & \textbf{78.5} & 39.9 & 40.8 & 69.0 & 48.8 & \cellcolor{blue!8}60.2 & \textbf{91.7} & \textbf{75.2} & \textbf{82.0} & 23.3 & \cellcolor{blue!8}68.1 & 20.1 & 66.5 & \textbf{37.0} & \cellcolor{blue!8}41.2 \\
 & Ours (SFT)                     & \textbf{85.3} & \textbf{78.5} & \textbf{44.5} & \textbf{42.0} & \textbf{69.5} & \textbf{50.4} & \cellcolor{blue!8}\textbf{61.7}\up{1.5} & 91.6 & 74.5 & 80.2 & \textbf{32.7} & \cellcolor{blue!8}\textbf{69.8}\up{1.7} & \textbf{27.4} & \textbf{72.8} & 34.2 & \cellcolor{blue!8}\textbf{44.8}\up{3.6} \\
\cmidrule(lr){2-18}
 & $\mathcal{D}_\text{math}$ (GRPO) & 78.6 & 81.2 & 42.0 & \textbf{44.4} & 68.4 & 36.2 & \cellcolor{blue!8}58.5 & 91.0 & \textbf{78.2} & \textbf{83.7} & 26.7 & \cellcolor{blue!8}69.9 & 19.5 & 73.5 & 45.9 & \cellcolor{blue!8}46.3 \\
 & Ours (GRPO)                    & 81.2 & \textbf{81.7} & \textbf{52.2} & 42.6 & \textbf{72.2} & \textbf{53.1} & \cellcolor{blue!8}\textbf{63.8}\up{5.3} & 91.7 & 78.1 & 81.0 & 26.7 & \cellcolor{blue!8}69.4\dn{0.5} & \textbf{23.8} & 76.3 & 44.0 & \cellcolor{blue!8}48.0\up{1.7} \\
 & $\mathcal{D}_\text{mixed}$ (GRPO) & \textbf{81.9} & 80.7 & 51.0 & 43.9 & 69.1 & 37.3 & \cellcolor{blue!8}60.6\up{2.1} & \textbf{92.1} & 77.7 & 81.6 & \textbf{30.0} & \cellcolor{blue!8}\textbf{70.3}\up{0.4} & 23.2 & \textbf{77.2} & \textbf{49.4} & \cellcolor{blue!8}\textbf{49.9}\up{3.6} \\
\cmidrule(lr){1-18}
\multirow{5}{*}{\shortstack[l]{DeepSeek-R1\\-Distill-8B}}
 & Math (SFT)                     & 85.1 & \textbf{75.8} & 36.3 & 24.0 & \textbf{57.4} & \textbf{26.5} & \cellcolor{blue!8}50.9 & \textbf{89.6} & \textbf{71.2} & \textbf{67.2} & 3.3 & \cellcolor{blue!8}57.8 & \textbf{26.3} & 32.8 & \textbf{25.3} & \cellcolor{blue!8}28.1 \\
 & Ours (SFT)                     & \textbf{86.1} & 75.7 & \textbf{40.8} & \textbf{25.2} & 55.8 & 23.6 & \cellcolor{blue!8}\textbf{51.2}\up{0.3} & 88.5 & 70.4 & 66.2 & \textbf{26.7} & \cellcolor{blue!8}\textbf{63.0}\up{5.2} & 23.9 & \textbf{38.9} & 24.8 & \cellcolor{blue!8}\textbf{29.2}\up{1.1} \\
\cmidrule(lr){2-18}
 & $\mathcal{D}_\text{math}$ (GRPO) & 81.1 & \textbf{75.0} & 38.4 & 23.9 & 58.9 & \textbf{20.8} & \cellcolor{blue!8}49.7 & 91.4 & \textbf{74.3} & 73.4 & 20.0 & \cellcolor{blue!8}64.8 & 22.4 & 45.4 & 11.6 & \cellcolor{blue!8}26.5 \\
 & Ours (GRPO)                    & 83.5 & 68.5 & \textbf{43.5} & \textbf{27.7} & 59.3 & 19.1 & \cellcolor{blue!8}50.2\up{0.5} & 88.4 & 66.7 & 78.2 & \textbf{23.3} & \cellcolor{blue!8}64.2\dn{0.6} & \textbf{26.8} & \textbf{48.3} & \textbf{21.8} & \cellcolor{blue!8}\textbf{32.3}\up{5.8} \\
 & $\mathcal{D}_\text{mixed}$ (GRPO) & \textbf{84.2} & 70.6 & 42.4 & 26.6 & \textbf{59.7} & 19.4 & \cellcolor{blue!8}\textbf{50.5}\up{0.8} & \textbf{91.7} & 73.1 & \textbf{80.7} & 20.0 & \cellcolor{blue!8}\textbf{66.4}\up{1.6} & 21.8 & 46.3 & 11.9 & \cellcolor{blue!8}26.7\up{0.2} \\
\midrule
\multicolumn{18}{c}{\textit{(D) Robustness across RL Algorithms (Qwen3-8B)}} \\
\midrule
 Qwen3-8B & Base & 79.8 & 78.8 & 41.2 & \textbf{44.4} & 70.5 & 46.6 & \cellcolor{blue!8}60.2 & 90.6 & 73.7 & 79.2 & 23.3 & \cellcolor{blue!8}66.7 & 22.6 & 70.6 & 40.5 & \cellcolor{blue!8}44.6 \\
\cmidrule(lr){1-18}
\multirow{4}{*}{Qwen3-8B}
 & GRPO         & 81.2 & \textbf{81.7} & 52.2 & 42.6 & 72.2 & \textbf{53.1} & \cellcolor{blue!8}\textbf{63.8}\up{3.6} & 91.7 & 78.1 & 81.0 & 26.7 & \cellcolor{blue!8}69.4\up{2.7} & \textbf{23.8} & 76.3 & \textbf{44.0} & \cellcolor{blue!8}\textbf{48.0}\up{3.4} \\
 & GPG          & 81.0 & 81.4 & 52.1 & 41.5 & \textbf{72.6} & 53.0 & \cellcolor{blue!8}63.6\up{3.4} & 92.0 & 77.5 & 81.3 & 26.7 & \cellcolor{blue!8}69.4\up{2.7} & 21.3 & 74.4 & 41.2 & \cellcolor{blue!8}45.7\up{1.1} \\
 & RLOO         & 79.3 & 81.4 & \textbf{52.8} & 41.1 & 71.2 & 52.4 & \cellcolor{blue!8}63.0\up{2.8} & \textbf{93.6} & \textbf{78.2} & \textbf{81.7} & \textbf{31.7} & \cellcolor{blue!8}\textbf{71.3}\up{4.6} & 21.3 & \textbf{77.4} & 42.8 & \cellcolor{blue!8}47.2\up{2.6} \\
 & Reinforce++  & \textbf{81.4} & 80.7 & 46.8 & 43.1 & 71.3 & 52.7 & \cellcolor{blue!8}62.7\up{2.5} & 91.4 & 75.1 & 80.8 & 30.0 & \cellcolor{blue!8}69.3\up{2.6} & 22.8 & 70.9 & 40.1 & \cellcolor{blue!8}44.6\up{0} \\
\bottomrule
\end{tabular}}
\caption{\textbf{Main Results and Ablation Studies.} (A)~Main results comparing base models with our \TheName{} data via SFT and GRPO. (B)~Data source ablation: our contextualized synthesis vs.\ raw Codeforces problems. (C)~Cross-domain mixing with mathematical/benchmark data. (D)~RL algorithm robustness. \textbf{Avg} = category mean; subscripts show absolute change vs.\ the reference row. Best per group in \textbf{bold}.}
\label{tab:main_results}
\vspace{-3mm}
\end{table*}

\subsection{Experimental Setup}
We briefly summarize our experimental setup. For base models, we use Qwen3-4B, Qwen3-8B, and DeepSeek-R1-Distill-8B. Our training incorporates both Supervised Fine-Tuning (SFT) and Group Relative Policy Optimization (GRPO). The synthesized \TheName{} dataset is utilized alongside mathematical datasets to study composability. We evaluate across general reasoning (e.g., ARC, BBH, GPQA, MMLU, DROP), mathematical benchmarks (GSM8K, MATH, AIME), coding benchmarks (HumanEval, MBPP), and fine-grained tasks (CharBench, StringBench, MazeBench). Comprehensive details regarding the datasets, training hyperparameters, algorithms, and evaluation metrics are provided in Appendix~\ref{appendix:eval_setting} and~\ref{sec:appendix_training}.

\subsection{RQ1: Generalization of Reasoning Capabilities}

To answer \textbf{RQ1}, we present a comprehensive evaluation across general reasoning, mathematical, and coding benchmarks in Table~\ref{tab:main_results}. The results explicitly confirm that our method significantly enhances reasoning capabilities and generalizes profoundly across domains.

\textbf{Main Results (Panel A):} Across all three base model families, applying our \TheName{} data through SFT and GRPO yields substantial performance leaps. For instance, on Qwen3-4B, GRPO lifts the general reasoning average by 3.8 points and the mathematical reasoning average by 4.8 points, highlighted by a staggering improvement on AIME (15.6 $\rightarrow$ 28.3). Notably, the DeepSeek-R1-Distill-8B model experiences the most dramatic gains in math (+9.7 average), proving that our methodology is highly effective even on already distilled reasoning models. Furthermore, contrary to the typical alignment tax where reasoning improvements degrade coding skills, our models maintain or improve on coding benchmarks (e.g., Qwen3-8B MBPP improves from 40.5 to 44.0).

\textbf{Synthesis Ablation (Panel B):} To isolate the source of improvement, we compared our contextualized synthesis with training on raw Codeforces problems, where the same test cases were incorporated using the same input-fuse procedure. The contrast is stark: directly fine-tuning or applying RL on raw Codeforces data heavily harms general reasoning. For example, Qwen3-4B trained on raw Codeforces (SFT) suffers a 16.5-point drop on general reasoning and a 21.6-point drop on math. This confirms that merely exposing models to code data induces overfitting to syntax at the expense of general logic, whereas our narrative fusion effectively distills algorithmic logic into a universally applicable reasoning format.

\textbf{Composability and RL Robustness (Panels C \& D):} Panel (C) highlights the exceptional composability of our data. When mixing \TheName{} with mathematical data ($\mathcal{D}_\text{mixed}$), Qwen3-4B achieves the best of both worlds, setting peak scores on GSM8K (93.2) and AIME (30.0) without compromising general reasoning. Finally, Panel (D) explores algorithmic robustness. While GRPO, GPG, RLOO, and Reinforce++ all yield solid improvements over the base model, the gains remain largely consistent across different RL optimization strategies. This indicates that the core driver of our success stems from the high-quality, hallucination-mitigated reasoning trajectories synthesized by our pipeline, rather than reliance on any single specific RL algorithm.

\textbf{Comparison with recent baselines:}
Following reviewer suggestions, we additionally compare MIMIC against several closely related methods, including TeaR, CoE, ExecGrounded, SynLogic, and Enigmata. As shown in Table~\ref{tab:related_baselines}, MIMIC consistently outperforms these baselines across the majority of evaluation settings. We believe this result further supports our central claim: preserving executable logic alone is insufficient if the supervision still retains strong programming-specific artifacts, whereas our narrative fusion formulation yields more transferable reasoning signals.

\begin{table*}[t]
\centering
\resizebox{\textwidth}{!}{
\begin{tabular}{@{}ll ccccccc ccccc cccc @{}}
\toprule
\multirow{2}{*}{\textbf{Model}} & \multirow{2}{*}{\textbf{Setting}}
  & \multicolumn{7}{c}{\textbf{General Reasoning}}
  & \multicolumn{5}{c}{\textbf{Math}}
  & \multicolumn{4}{c}{\textbf{Code}} \\
\cmidrule(lr){3-9} \cmidrule(lr){10-14} \cmidrule(lr){15-18}
& & ARC & BBH & MIMIC & GPQA & MMLU & DROP & \cellcolor{blue!8}\textbf{Avg}
  & GSM8K & GSM+ & MATH & AIME & \cellcolor{blue!8}\textbf{Avg}
  & HEval & CruxE & MBPP & \cellcolor{blue!8}\textbf{Avg} \\
\midrule
\multicolumn{18}{c}{\textit{Comparison with Closely Related Baselines}} \\
\midrule
\multicolumn{18}{c}{\textit{(A) SFT}} \\
\midrule
\multirow{4}{*}{Qwen3-4B}
 & Ours (SFT)              & \textbf{85.9} & \textbf{77.0} & \textbf{43.5} & \textbf{40.4} & \textbf{67.3} & \textbf{55.5} & \cellcolor{blue!8}\textbf{61.6} & \textbf{91.0} & \textbf{74.9} & \textbf{77.6} & \textbf{16.7} & \cellcolor{blue!8}\textbf{65.0} & 32.1 & \textbf{70.8} & 35.8 & \cellcolor{blue!8}\textbf{46.2} \\
 & CoE (SFT)               & 85.2 & 43.9 & 28.9 & 25.4 & 44.8 & 42.6 & \cellcolor{blue!8}45.1\dn{16.5} & 71.5 & 54.9 & 40.4 & 6.7 & \cellcolor{blue!8}43.4\dn{21.6} & 34.2 & 65.6 & 34.6 & \cellcolor{blue!8}44.8\dn{1.4} \\
 & ExecGrounded (SFT)      & 75.4 & 66.9 & 39.2 & 35.0 & 40.8 & 26.1 & \cellcolor{blue!8}47.2\dn{14.4} & 77.6 & 63.4 & 74.4 & \textbf{16.7} & \cellcolor{blue!8}58.0\dn{7.0} & 23.8 & 63.0 & \textbf{45.5} & \cellcolor{blue!8}44.1\dn{2.1} \\
 & Enigmata (SFT)          & 72.0 & 49.1 & 21.5 & 34.6 & 30.1 & 37.1 & \cellcolor{blue!8}40.7\dn{20.9} & 60.9 & 53.8 & 69.2 & 13.3 & \cellcolor{blue!8}49.3\dn{15.7} & \textbf{35.4} & 41.9 & 24.9 & \cellcolor{blue!8}34.1\dn{12.1} \\
\midrule
\multirow{4}{*}{\shortstack[l]{DeepSeek-R1\\-Distill-8B}}
 & Ours (SFT)              & \textbf{86.1} & \textbf{75.7} & \textbf{40.8} & 25.2 & \textbf{55.8} & \textbf{23.6} & \cellcolor{blue!8}\textbf{51.2} & \textbf{88.5} & \textbf{70.4} & 66.2 & \textbf{26.7} & \cellcolor{blue!8}\textbf{63.0} & 23.9 & \textbf{38.9} & 24.8 & \cellcolor{blue!8}\textbf{29.2} \\
 & CoE (SFT)               & 83.8 & 43.8 & 36.6 & 21.5 & 43.0 & 22.7 & \cellcolor{blue!8}41.9\dn{9.3} & 62.0 & 49.1 & 55.4 & 23.3 & \cellcolor{blue!8}47.5\dn{15.5} & \textbf{27.4} & 35.1 & 14.8 & \cellcolor{blue!8}25.8\dn{3.4} \\
 & ExecGrounded (SFT)      & 76.3 & 59.3 & 38.9 & 25.5 & 51.3 & 10.0 & \cellcolor{blue!8}43.6\dn{7.6} & 66.5 & 49.1 & \textbf{77.1} & 23.3 & \cellcolor{blue!8}54.0\dn{9.0} & 12.8 & 35.6 & 20.6 & \cellcolor{blue!8}23.0\dn{6.2} \\
 & Enigmata (SFT)          & 75.6 & 37.7 & 18.7 & \textbf{28.7} & 41.1 & 10.0 & \cellcolor{blue!8}35.3\dn{15.9} & 61.1 & 40.7 & 62.3 & 20.0 & \cellcolor{blue!8}46.0\dn{17.0} & 21.3 & 34.9 & \textbf{26.2} & \cellcolor{blue!8}27.5\dn{1.7} \\
\midrule
\multicolumn{18}{c}{\textit{(B) GRPO}} \\
\midrule
\multirow{4}{*}{Qwen3-4B}
 & Ours (GRPO)             & \textbf{80.6} & \textbf{77.9} & \textbf{51.3} & 40.8 & \textbf{67.5} & \textbf{58.3} & \cellcolor{blue!8}\textbf{62.7} & \textbf{91.0} & \textbf{76.4} & 80.3 & \textbf{28.3} & \cellcolor{blue!8}\textbf{69.0} & \textbf{31.7} & \textbf{72.7} & \textbf{43.6} & \cellcolor{blue!8}\textbf{49.3} \\
 & TeaR (GRPO)             & 80.1 & 75.1 & 45.1 & 32.4 & 65.1 & 41.2 & \cellcolor{blue!8}56.5\dn{6.2} & 89.4 & 75.6 & 75.5 & 20.0 & \cellcolor{blue!8}65.1\dn{3.9} & 25.6 & 66.8 & 33.9 & \cellcolor{blue!8}42.1\dn{7.2} \\
 & SynLogic (GRPO)         & 77.1 & 76.2 & 41.0 & 39.5 & 62.3 & 41.7 & \cellcolor{blue!8}56.3\dn{6.4} & 90.2 & 69.0 & \textbf{81.3} & 23.3 & \cellcolor{blue!8}66.0\dn{3.0} & 26.8 & 64.2 & 37.0 & \cellcolor{blue!8}42.7\dn{6.6} \\
 & Enigmata (GRPO)         & 76.8 & 75.6 & 41.0 & \textbf{41.5} & 62.4 & 42.1 & \cellcolor{blue!8}56.6\dn{6.1} & 89.0 & 67.9 & 80.6 & 23.3 & \cellcolor{blue!8}65.2\dn{3.8} & 29.9 & 68.8 & 36.2 & \cellcolor{blue!8}45.0\dn{4.3} \\
\midrule
\multirow{4}{*}{\shortstack[l]{DeepSeek-R1\\-Distill-8B}}
 & Ours (GRPO)             & \textbf{83.5} & \textbf{68.5} & \textbf{43.5} & 27.7 & \textbf{59.3} & \textbf{19.1} & \cellcolor{blue!8}\textbf{50.3} & \textbf{88.4} & 66.7 & 78.2 & \textbf{23.3} & \cellcolor{blue!8}\textbf{64.2} & \textbf{26.8} & \textbf{48.3} & \textbf{21.8} & \cellcolor{blue!8}\textbf{32.3} \\
 & TeaR (GRPO)             & 80.8 & 61.2 & 41.1 & 24.3 & 57.8 & 18.7 & \cellcolor{blue!8}47.3\dn{3.0} & 74.8 & \textbf{68.1} & 74.7 & 15.0 & \cellcolor{blue!8}58.1\dn{6.1} & \textbf{26.8} & 38.4 & 16.7 & \cellcolor{blue!8}27.3\dn{5.0} \\
 & SynLogic (GRPO)         & 76.1 & 62.4 & 40.3 & 27.9 & 44.6 & 14.3 & \cellcolor{blue!8}44.3\dn{6.0} & 83.9 & 58.7 & 76.2 & 20.0 & \cellcolor{blue!8}59.7\dn{4.5} & 20.7 & 33.4 & 12.8 & \cellcolor{blue!8}22.3\dn{10.0} \\
 & Enigmata (GRPO)         & 75.4 & 56.5 & 38.7 & \textbf{30.7} & 47.8 & 12.3 & \cellcolor{blue!8}43.6\dn{6.7} & 76.0 & 57.5 & \textbf{78.3} & \textbf{23.3} & \cellcolor{blue!8}58.8\dn{5.4} & 22.0 & 37.6 & 16.0 & \cellcolor{blue!8}25.2\dn{7.1} \\
\bottomrule
\end{tabular}
}
\caption{Comparison with closely related baselines. We compare MIMIC against representative recent methods under both SFT and GRPO. \textbf{Avg} = category mean; subscripts show absolute change vs.\ the reference row. Best per group in \textbf{bold}.}
\label{tab:related_baselines}
\end{table*}

\subsection{RQ2: Solving Fine-grained, Rule-based Deterministic Problems}

\begin{table}[t]
\centering
\small
\setlength{\tabcolsep}{3pt}
\renewcommand{\arraystretch}{1.15}
\begin{tabular}{@{}llcccc@{}}
\toprule
\textbf{Model} & \textbf{Method} & \textbf{Maze} & \textbf{String} & \textbf{Char} & \textbf{Char+CoT} \\
\midrule
\multirow{3}{*}{Qwen3-4B}
 & Base       & 16.0 & 13.6 & 37.9 & 97.1 \\
 & +SFT  & 18.0 & 22.8 & \textbf{40.4} & 97.8 \\
 & +GRPO & \textbf{22.0} & \textbf{26.8} & 39.1 & \textbf{98.2} \\
\midrule
\multirow{3}{*}{Qwen3-8B}
 & Base       & 22.0 & 26.2 & 38.6 & 98.0 \\
 & +SFT  & 22.0 & 37.5 & 39.3 & 98.6 \\
 & +GRPO & \textbf{40.0} & \textbf{39.5} & \textbf{41.3} & \textbf{99.2} \\
\midrule
\multirow{3}{*}{\shortstack[l]{DeepSeek-R1\\-Distill-8B}}
 & Base       & 10.0 & 14.0 & --- & 89.2 \\
 & +SFT  & 11.0 & \textbf{22.2} & --- & 91.2 \\
 & +GRPO & \textbf{14.0} & 20.0 & --- & \textbf{94.7} \\
\bottomrule
\end{tabular}
\caption{\textbf{Character-level and spatial reasoning benchmarks.} We evaluate on benchmarks targeting fine-grained perception and rules (e.g., comparing string patterns like 9.11 and 9.9): CharBench, StringBench, and MazeBench. Best per model family in \textbf{bold}.}
\label{tab:special_benchmarks}
\vspace{-3mm}
\end{table}

Regarding \textbf{RQ2}, we investigate whether our models can overcome the notorious vulnerabilities LLMs face with tokenization artifacts and fine-grained, deterministic problems. 
To assess this, we tested the models on MazeBench (spatial state tracking), StringBench, and CharBench (character-level perception) in Table~\ref{tab:special_benchmarks}\footnote{Because Deepseek-R1-Distilled-8B only has think mode, the "directly output answer" constraint of CharBench does not apply to it; we only report performance with cot.}. It is crucial to note that \emph{none of these specific benchmarks or their exact formats were included in our training data}.

The results demonstrate profound, zero-shot generalizations to deterministic rules. Qwen3-8B with GRPO demonstrates remarkable improvements across benchmarks, jumping to 40.0\% (+82\% relative gain) on MazeBench, rising to 39.5\% on StringBench to rival larger proprietary models, and achieving a near-perfect 99.2\% accuracy on CharBench.
By grounding the training in explicit execution traces and step-by-step state tracking, \TheName{} fundamentally alters how the model perceives inputs. Instead of relying on blurry semantic approximations caused by BPE tokenization, the model learns to internalize strict, deterministic state perception, allowing it to robustly process and manipulate token structures.\footnote{To illustrate this, we provide case studies in Appendix~\ref{appendix:case_study}}

\subsection{RQ3: Reasoning vs. Pattern Matching}
\label{subsec:rq3_presentation_format}

After discussing large-scale training results, we address \textbf{RQ3} by examining whether our synthesized data encourages reasoning rather than recall of memorized code templates. We conduct a preliminary study on \textsc{Qwen3-32B} (context length 16384) under four problem presentation formats: (1) \textbf{original}, the raw Codeforces statement; (2) \textbf{refined}, a minimal version with the narrative background removed; (3) \textbf{fused} (ours), where the task is embedded into a newly constructed natural-language context; and (4) \textbf{synthesized}, where the model directly generates the final test item without constraint. We then evaluate how these formats affect model behavior and perceived difficulty.

\begin{table}[t]
\centering

\small
\setlength{\tabcolsep}{3pt}
\begin{tabular}{lcccc}
\toprule
\textbf{Format} & \textbf{Acc. (\%)} & \makecell{\textbf{Code Rate} \\ \textbf{(\%)}} & \makecell{\textbf{Acc. (\%)} \\ \textbf{w/ Code}} & \makecell{\textbf{Acc. (\%)} \\ \textbf{w/o Code}} \\
\midrule
Original     & 23.8 & 38.7 & 18.4 & 27.2 \\
Refined      & 34.4 & 29.9 & 30.6 & 36.0 \\
Fused (Ours) & 42.2 & 0.7  & 15.0 & 42.4 \\
Synthesized  & 61.8 & 0.8  & 41.3 & 62.0 \\
\bottomrule
\end{tabular}
\caption{Preliminary comparison of different problem presentation formats on \textsc{Qwen3-8B}. ``Code Rate'' denotes the proportion of cases where the model attempted to write code during reasoning.}
\label{tab:format_preliminary}
\vspace{-3mm}
\end{table}

\begin{table*}[htbp]
\centering
\resizebox{\textwidth}{!}{
\begin{tabular}{@{}ll ccccccc ccccc cccc @{}}
\toprule
\multirow{2}{*}{\textbf{Model}} & \multirow{2}{*}{\textbf{Reward}}
  & \multicolumn{7}{c}{\textbf{General Reasoning}}
  & \multicolumn{5}{c}{\textbf{Math}}
  & \multicolumn{4}{c}{\textbf{Code}} \\
\cmidrule(lr){3-9} \cmidrule(lr){10-14} \cmidrule(lr){15-18}
& & ARC & BBH & MIMIC & GPQA & MMLU & DROP & \cellcolor{blue!8}\textbf{Avg}
  & GSM8K & GSM+ & MATH & AIME & \cellcolor{blue!8}\textbf{Avg}
  & HEval & CruxE & MBPP & \cellcolor{blue!8}\textbf{Avg} \\
\midrule
\multirow{5}{*}{Qwen3-8B}
 & Base               & 79.8 & 78.8 & 41.2 & 44.4 & 70.5 & 46.6 & \cellcolor{blue!8}60.2 & 90.6 & 73.7 & 79.2 & 23.3 & \cellcolor{blue!8}66.7 & 22.6 & 70.6 & 40.5 & \cellcolor{blue!8}44.6 \\
 & Binary (GRPO)      & 81.2 & \textbf{81.7} & 52.2 & 42.6 & \textbf{72.2} & \textbf{53.1} & \cellcolor{blue!8}63.8\up{3.6} & 91.7 & \textbf{78.1} & 81.0 & 26.7 & \cellcolor{blue!8}69.4\up{2.7} & 23.8 & \textbf{76.3} & 44.0 & \cellcolor{blue!8}\textbf{48.0}\up{3.4} \\
 & \textbf{CIR (Ours)} & \textbf{82.7} & 80.6 & \textbf{54.5} & \textbf{48.5} & 71.6 & 53.5 & \cellcolor{blue!8}\textbf{65.2}\up{5.0} & \textbf{92.6} & 78.0 & \textbf{81.4} & \textbf{30.0} & \cellcolor{blue!8}\textbf{70.5}\up{3.8} & \textbf{25.3} & 76.2 & 40.9 & \cellcolor{blue!8}47.5\up{2.9} \\
 & PRM800K            & 79.8 & 79.3 & 43.8 & 42.7 & 71.6 & 50.9 & \cellcolor{blue!8}61.4\up{1.2} & 91.1 & 74.7 & 80.5 & 23.3 & \cellcolor{blue!8}67.4\up{0.7} & 23.2 & 71.8 & 40.5 & \cellcolor{blue!8}45.2\up{0.6} \\
 & Math-Shepherd      & 81.4 & 79.8 & 43.9 & 43.1 & 71.4 & 52.7 & \cellcolor{blue!8}62.1\up{1.9} & 90.9 & 74.3 & 75.9 & 20.0 & \cellcolor{blue!8}65.3\dn{1.4} & 21.3 & 72.3 & \textbf{44.4} & \cellcolor{blue!8}46.0\up{1.4} \\
\bottomrule
\end{tabular}}
\caption{\textbf{Reward Signal Ablation.} Comparing reward strategies under GRPO training: (1) \textbf{Binary} outcome reward; (2) \textbf{CIR} (Code-Instrumented Reward), our approach leveraging execution checkpoints; (3) \textbf{PRM800K}; (4) \textbf{Math-Shepherd}. Subscripts show absolute change vs.\ Base.}
\label{tab:reward_ablation}
\vspace{-3mm}
\end{table*}

Table~\ref{tab:format_preliminary} provides a clear empirical answer to \textbf{RQ3} by isolating the impact of problem presentation. When processing the \textbf{original} Codeforces statements, the model exhibits a strong bias toward pattern matching rather than genuine reasoning, attempting to write raw code snippets in 38.7\% of the cases. This ``recall shortcut'' drastically harms performance (18.4\% accuracy with code vs.\ 27.2\% without), as the model wastes context capacity reconstructing memorized competitive-programming templates rather than analyzing the specific logic of the given test case. Merely stripping the narrative (the \textbf{refined} format) only partially mitigates this issue, leaving a high code generation rate of 29.9\%.

Crucially, our \textbf{fused} format fundamentally changes the model's behavior. By embedding the underlying algorithmic structure into novel, everyday natural-language scenarios, it removes the semantic cues that trigger template recall. As a result, the code rate drops sharply to 0.7\%, while reasoning accuracy rises to 42.2\% (vs.\ 23.8\% for the original format), indicating that the model is driven toward step-by-step deduction rather than shallow pattern matching. Although the fully \textbf{synthesized} format also yields a low code rate, its lack of constraint causes the generated test cases to be overly simple and stereotyped~(accuracy 61.8\%), failing to capture the adversarial hardness of rigorous benchmarks. Therefore, our fused format strikes a better balance: it effectively eliminates superficial memorization while preserving the intricate deductive complexity required for robust reasoning\footnote{More analyses, case studies, and results are provided in Appendix~\ref{appendix:data_analysis}.}.

\subsection{RQ4: Effectiveness of the Process Reward}

Finally, to answer \textbf{RQ4}, we ablate the effectiveness of our reward modeling strategy under the GRPO framework. Table~\ref{tab:reward_ablation} compares our Code-Instrumented Reward (CIR) against a sparse Binary outcome reward, as well as two state-of-the-art, learned Process Reward Models (PRM800K and Math-Shepherd). While a simple Binary reward provides a solid baseline improvement (+3.6 on General Avg, +2.7 on Math Avg for Qwen3-8B), it lacks the granularity to effectively penalize intermediate logical fallacies. On the other hand, injecting external learned PRMs yields surprisingly meager gains. For instance, PRM800K only improves the General Avg by 1.2 points, and Math-Shepherd actually degrades mathematical reasoning slightly relative to the base average (65.3 vs.\ 66.7). This underperformance stems from the fact that existing PRMs are heavily overfitted to their specific training distributions (typically step-by-step mathematical derivations) and actively struggle to reliably score the diverse, highly structured algorithmic reasoning trajectories generated by our pipeline.

In stark contrast, our CIR methodology bypasses the need for an external, fallible neural network by leveraging deterministic execution checkpoints natively embedded in the synthesized code. This guarantees perfectly accurate, dense, and graded reward signals at each logical step. Consequently, CIR delivers the best results on both general reasoning and mathematical reasoning, surpassing binary reward and existing PRMs by +5.0 and +3.8 points over the baseline, respectively. By verifying algorithmic progress domain-agnostically through direct computation, CIR proves a highly effective and robust mechanism for reinforcing genuine reasoning in LLMs.

\paragraph{Extended Investigations and Key Insights.}
We further conduct additional experiments on \TheName{} in the appendix. Appendix~\ref{appendix:data_analysis} presents analyses of the dataset distribution across categories. Appendix~\ref{token_ditribution} examines changes in output patterns and token probabilities before and after training, providing insight into how training on our data affects model behavior. Appendix~\ref{sec:training_dynamics} reports training dynamics to illustrate how the model evolves during optimization on our data. Appendix~\ref{sec:appendix_type_ablation} presents a task-type ablation study to analyze the contribution of each category during training.





\section{Discussion and Conclusion}
Beyond empirical gains, this work provokes critical reflections on LLM reasoning paradigms. First, it redefines code from a mere generation target into a rigorous \textit{cognitive scaffold}. The structural determinism of programs forces LLMs to ground linguistic reasoning in verifiable logic, effectively bridging symbolic execution with semantic understanding to mitigate hallucination. Second, this paradigm democratizes Process Reward Models (PRMs). Because intermediate programmatic execution states inherently provide cost-free, high-fidelity step-level verification, MIMIC unlocks a scalable pathway for process supervision without relying on prohibitive human annotations or closed-source teacher models.

In conclusion, we propose MIMIC, a framework that translates executable code and state transitions into high-quality natural language reasoning trajectories. By compelling LLMs to imitate the deterministic logic of programs, we significantly enhance their mathematical and logical capabilities. Ultimately, shifting the ``imitation game'' toward program execution establishes a self-verifiable, sustainable data engine, paving the way for more interpretable and reliable AI reasoning.


\section*{Limitations}
\label{sec:limitations}

\paragraph{Reward signal granularity.}
Our Code-Instrumented Reward uses a simple step-counting formulation ($k/K$), treating all execution checkpoints equally. The primary goal of this design is not to propose an optimal reward model, but rather to demonstrate that the rich intermediate results naturally available from program execution can serve as effective training signals, even in their simplest form. The fact that this straightforward approach already outperforms established process reward models (Table~\ref{tab:reward_ablation}) suggests that the deterministic, code-derived nature of the signal is itself the key ingredient.

\paragraph{Data source and scale.}
We currently use only Codeforces as the problem source, a subset of pretraining corpora for contextualization, and generate only $\sim$20 contextualized problems per seed problem, with all synthesis performed by a Qwen3.5-27B model rather than a frontier-scale system. While all of these~(problem sources, corpus coverage, per-seed generation volume, and synthesizer model capacity) can be trivially scaled, the current configuration is already more than sufficient, comfortably supporting both SFT ($\sim$114K samples) and RL ($\sim$10K prompts) training with consistent improvements across all evaluated dimensions.



\bibliography{main}

\clearpage
\appendix
\newtcolorbox[auto counter]{examplebox}{
    colback=yellow!10,  
    colframe=yellow!70!black,  
    arc=10pt,
    title={\textcolor{black}{\textbf{Example \thetcbcounter}}},
    colbacktitle=yellow!40,  
    breakable,
}
\newtcolorbox[auto counter]{promptbox}{
  colback=white,
  colframe=purple!80!black,
  arc=10pt,
  title={\textcolor{white}{\textbf{Prompt \thetcbcounter}}},
  colbacktitle=purple!80!black,
  breakable
}
\section{Data Synthesis Pipeline}
\label{appendix:pipeline}

This appendix describes the complete data synthesis pipeline used to transform competitive programming problems into story-based reasoning training data. The pipeline consists of four stages, each building upon the previous one.

We used only publicly available artifacts in this work, including datasets, benchmarks, and base models released through Hugging Face or other official repositories. These resources were used in accordance with their respective licenses and terms of use, and their original sources are cited in the paper. To the best of our knowledge, all used artifacts are openly released for research use.

We decompose our data synthesis pipeline into three parts:
\begin{enumerate}
    \item \textbf{Problem background synthesis}: The narrative context of each question is synthesized from pretraining corpora~(Appendix ~\ref{appendix:stage2}).
    \item \textbf{Test case synthesis}: When test cases are not available, they are generated from the code through guided random generation(Appendix ~\ref{appendix:stage3}).
    \item \textbf{Chain-of-thought synthesis}: The CoT is synthesized from execution traces obtained via code instrumentation, where intermediate checkpoints provide process-level reasoning signals (Appendix~\ref{appendix:stage4}).
\end{enumerate}
\subsection{Stage 1: Problem Classificaton and Mathematical Kernel Extraction}

Large language models (LLMs) still exhibit systematic weaknesses on tasks that require \emph{fine-grained symbolic fidelity}, such as character-level manipulation, exact numerical computation, and step-by-step state transition under strict rules. Canonical examples include counting the number of occurrences of a specific letter in a word (e.g., how many \texttt{r}'s appear in \texttt{strawberry}) or comparing decimal numbers such as \texttt{9.9} and \texttt{9.11}. While such tasks are often trivial for programs, they remain challenging for LLMs due to tokenization artifacts, approximate pattern matching, and brittle long-chain reasoning.

To address this gap, we propose \TheName{}, which transforms competitive programming problems into diverse, \emph{execution-verified} reasoning training data---injecting \emph{algorithmic reasoning primitives} that programs handle trivially but LLMs struggle with. Specifically, we collect 10,008 problems from the \textbf{open-r1/codeforces}\footnote{\url{https://huggingface.co/datasets/open-r1/codeforces}}, together with their accepted (AC) solutions. This source is particularly suitable for our purpose because each problem typically includes a well-specified natural language statement, detailed \texttt{input format} and \texttt{output format}, and official test cases used by the contest platform for evaluation~(some included). Moreover, AC solutions have already passed the hidden and public judge cases in official Codeforces contests, providing a strong correctness guarantee. This allows us to use code as a reliable intermediate representation for synthesizing high-quality training instances, without having to independently verify the semantic validity of each solution.

We categorize each problem according to the \emph{primary reasoning skill} required to solve it, aiming to capture those capability dimensions where code execution is naturally reliable but LLM reasoning is often error-prone. The resulting taxonomy contains five categories, shown in Table~\ref{tab:reasoning_taxonomy}. Problem classification is performed by Qwen3.5-27B using type-specific prompting as in Table~\ref{tab:reasoning_taxonomy}. To support type-aware data pairing in Stage~2, corpus passages from the pretraining data are independently assigned to compatible reasoning types after classification.

\begin{table*}[htbp]
\centering
\small
\begin{tabularx}{\textwidth}{p{0.08\textwidth} p{0.23\textwidth} p{0.16\textwidth} X}
\toprule
\textbf{Type} & \textbf{Category} & \textbf{Subtype} & \textbf{Description} \\
\midrule

\multirow{4}{*}{\textbf{Type 1}} 
& \multirow{4}{=}{\textbf{Text-Encoded Visual / Spatial Reasoning}} 
& 1.1 Grid / board / pixel encoding 
& Layouts represented as character matrices or ASCII art, where the solver must reconstruct a 2D structure from text and reason about adjacency, connectivity, or shape. \\
& & 1.2 Geometry / coordinates / directions 
& Spatial reasoning grounded in language, such as left/right/up/down relations, segment intersection, coordinate movement, or point-in-polygon logic. \\
& & 1.3 Glyph or compositional character structure 
& Reasoning based on visual properties of symbols or characters, including radicals, stroke counts, mirrored forms, rotated shapes, or internal composition. \\
& & 1.4 Textual charts / tabular layout understanding 
& Information encoded through alignment, indentation, spacing, or fixed-width formatting, such as tree structures or visually organized tables. \\
\midrule

\multirow{3}{*}{\textbf{Type 2}} 
& \multirow{3}{=}{\textbf{Sub-Token / Character Manipulation}} 
& 2.1 Character-level counting / extraction / replacement 
& Fine-grained operations on strings, including counting symbol frequency, extracting characters at specified positions, or performing local edits and substitutions. \\
& & 2.2 Character-wise mapping and ciphers 
& Per-character transformation rules such as Caesar shifts, ASCII-based conversion, or Base64-style encoding and decoding. \\
& & 2.3 Strict format validation 
& Verifying whether a string satisfies exact symbolic constraints, including delimiter placement, serial patterns, and regular-expression-like rules. \\
\midrule

\multirow{3}{*}{\textbf{Type 3}} 
& \multirow{3}{=}{\textbf{Precise Mathematical Computation}} 
& 3.1 Discrete and high-precision arithmetic 
& Exact computation over very large integers, long decimals, exponentiation, or arithmetic that cannot tolerate approximation. \\
& & 3.2 Algebraic and symbolic manipulation 
& Equation solving, matrix derivation, symbolic transformation, calculus-style reasoning, or other formula-driven procedures. \\
& & 3.3 Probability, statistics, and set operations 
& Computation involving variance, standard deviation, permutations, combinations, probability distributions, or exact set-based reasoning. \\
\midrule

\multirow{3}{*}{\textbf{Type 4}} 
& \multirow{3}{=}{\textbf{Large-Scale Data Processing}} 
& 4.1 Retrieval and filtering 
& Searching through large collections to identify records that satisfy specified conditions. \\
& & 4.2 Grouping, aggregation, and sorting 
& Reorganizing global data via sorting, grouping, counting, summation, or other aggregate operations. \\
& & 4.3 Structural mapping and transformation 
& Converting data from one representation to another, such as JSON-to-CSV conversion or flattening nested dictionaries. \\
\midrule

\multirow{3}{*}{\textbf{Type 5}} 
& \multirow{3}{=}{\textbf{Strict Rule / State Simulation}} 
& 5.1 Path finding and graph traversal 
& Reasoning over reachable states or paths, typically involving DFS, BFS, shortest-path search, or related graph procedures. \\
& & 5.2 Constraint satisfaction and backtracking 
& Searching for valid assignments under mutually dependent rules, such as Sudoku, N-Queens, or scheduling problems. \\
& & 5.3 Iterative and dynamical system simulation 
& Simulating systems whose state at step $t$ depends exactly on step $t-1$, including cellular automata, physical processes, or code-execution-style state updates. \\
\bottomrule
\end{tabularx}
\caption{Reasoning taxonomy used in Stage~1. Each problem is assigned to the category that best reflects its dominant reasoning bottleneck.}
\label{tab:reasoning_taxonomy}
\end{table*}

After classification, we further extract mathematical kernel of each problem by removing narrative background and contest-specific flavor text, while preserving the exact task semantics, constraints, and input/output requirements. The prompt used for extraction is shown below.

\begin{promptbox}
\textbf{Role: Competitive Programming Problem Refiner}

\textbf{Task Description.}
You are an efficient translator for competitive programming problems. Your goal is to remove all background stories, character names, narrative framing, and other non-essential text, and rewrite the problem as a concise and direct task description.

\textbf{Instructions.}
I will provide a \texttt{[Problem Description]}. You should retain only the problem kernel and express it in the most concise and explicit form possible.

\textbf{Constraints.}
\begin{itemize}
    \item \textbf{Core Only:} Do not include any preamble or explanation. Use a direct structure such as ``Given \ldots, find/calculate \ldots''
    \item \textbf{Plain Talk:} Use standard competitive-programming terminology that is immediately understandable, such as \emph{array}, \emph{subarray}, \emph{non-overlapping}, \emph{maximum sum}, \emph{path}, or \emph{tree}.
    \item \textbf{Output Format:} Return the result strictly in JSON format with a single \texttt{"text"} field. Do not include any extra explanation or markdown outside the JSON.
    \item \textbf{Preserve Semantics:} Do not omit or alter any logic, constraints, input/output requirements, or the intended solution structure of the original problem.
\end{itemize}

\textbf{Input Problem Description:} \\
\texttt{George wants to buy a new phone but doesn't have enough money, so he works as a programmer. His boss gives him a sequence of $n$ integers and asks him to choose $k$ pairs of integers $[l_i, r_i]$ such that they represent $k$ non-overlapping subarrays of length $m$ to maximize the total sum \ldots}\\

\textbf{Output:}

\texttt{"text": "Given a sequence of length n, choose k non-overlapping subarrays, each of length m, such that the sum of all elements in these subarrays is maximized."}\\

\textbf{Problem Description:} \\
\texttt{\{question\}}\\

\textbf{Output Format.}
After reasoning, return the extracted kernel directly in JSON format:
\begin{verbatim}
{
  "text": "extracted kernel"
}
\end{verbatim}
\end{promptbox}

We further illustrate the extraction step with a concrete example. As shown below, mathematic kernal extraction removes narrative framing and contest-specific wording while preserving the exact computational objective of the original problem. This helps isolate the problem kernel and makes the resulting instance better suited for reasoning-focused synthesis.

\begin{examplebox}
\textbf{Mathematic Kernal Extraction}
To reduce competitive programming-specific narrative cues, we refine each original problem into a concise kernel that preserves only the mathematical task. For example, the original statement of \textit{Codeforces 1304/A} (\textit{Two Rabbits}) contains an extended story about two rabbits in a park:
\\

\emph{``Being tired of participating in too many Codeforces rounds, Gildong decided to take some rest in a park \ldots He noticed that the two rabbits were hopping towards each other \ldots Will the two rabbits be at the same position at the same moment?''}
\\

After extraction, this narrative framing is removed and only the task-relevant kernel remains:
\\

\emph{``Given integers $x, y, a, b$ where $x < y$, determine the smallest non-negative integer $t$ such that $x + a t = y - b t$. If no such $t$ exists, output $-1$; otherwise, output $t$.''}
\\

This extraction step preserves the exact semantics, constraints, and intended solution structure, while removing background stories and stylistic contest phrasing that may otherwise trigger superficial dataset recall.
\end{examplebox}

\subsection{Stage 2: Problem Synthesis --- Story Wrapping}
\label{appendix:stage2}

Stage~2 rewrites each algorithmic problem into a realistic exam-style question by pairing it with an external real-world passage and using the passage as the narrative environment for the problem. The goal is to preserve the exact computational structure of the original task while removing contest-style presentation and recasting it as a self-contained question that looks like it was written for a human test.

The external passages are sampled from \textbf{FineWebEdu}\footnote{\url{https://huggingface.co/datasets/HuggingFaceFW/fineweb-edu}}, a large-scale educational web corpus containing expository and informational text across diverse domains. We use FineWebEdu because its topical breadth and great education value make it a good source of realistic background material for written exam questions.

A key challenge in this stage is that not every passage can serve as a natural wrapper for every problem. If the passage is incompatible with the problem's reasoning type, the resulting rewrite often becomes forced: the story is only loosely attached to the underlying logic, important quantities or structures must be invented, or the final question retains obvious traces of programming formulations. We therefore perform \emph{type-aware pairing}: before synthesis, we classify corpus passages by the kinds of reasoning structures they can support, and only pair a problem with passages compatible with its assigned type from Stage~1.

This filtering is especially important for the categories we target. Type~2 problems require coherent natural-language prose that can plausibly serve as the exact string being counted, transformed, or validated. Type~3 problems benefit from passages that already contain concrete extractable values, so that the synthesized question can reuse naturally occurring numbers or its occurring place rather than fabricated ones. Type~4 problems require structured entries or records that support filtering, grouping, sorting, or aggregation. Type~5 problems require passages with explicit sequential rules or state transitions, for example, procedural instructions, staged workflows, biological cycles, or turn-based processes, so that the narrative already contains a natural notion of stepwise evolution. For this reason, we classify FineWebEdu passages with a strict prompt and use the resulting labels to guide problem--passage matching.

The full prompt used for corpus classification is shown below:

\begin{promptbox}
\textbf{Strict Corpus Classification Prompt} \\[4pt]
You are a strict corpus classifier for exam problem synthesis. A text passage will be provided along with rule-based type hints from a fast heuristic scan. Your task is to determine which problem types (1--5) this text is \textbf{strictly} suitable for as background context or direct input.

\vspace{4pt}
\textbf{Type Definitions}

\begin{itemize}[leftmargin=*,itemsep=0pt, topsep=0.3em]
    \item \textbf{Type 1 -- Visual/Spatial Reasoning}: The text explicitly describes a grid, map, floor plan, coordinate system, or directional/positional relationships between objects (e.g., ``the room to the left of the hall'', ``cell (2,3) contains a wall''). The spatial structure must be clear enough to reconstruct a 2D layout.
    \item \textbf{Type 2 -- Character-Level Manipulation}: The text is coherent, continuous natural-language prose ($\geq 4$ complete sentences) with no dominant numeric or structural content. It is suitable as raw input for character-level operations: counting specific letters, detecting substrings, verifying text patterns, or string formatting tasks.
    \item \textbf{Type 3 -- Precise Mathematical Computation}: The text contains \textbf{at least 2 concrete numeric values} (measurements, statistics, prices, counts, dates, percentages) that are human-computable and can be directly extracted and used as parameters in a calculation problem.
    \item \textbf{Type 4 -- Data Processing}: The text contains a \textbf{structured dataset}: a table with rows and columns, or a list with \textbf{at least 3 entries} each having multiple named attributes (e.g., name + value + category). The data is suitable for filtering, sorting, grouping, or aggregation operations.
    \item \textbf{Type 5 -- State/Rule Simulation}: The text describes a \textbf{step-by-step process, iterative rules, or explicit state transitions} with enough detail to be modeled as a simulation (e.g., game rules with clear turn mechanics, a biological lifecycle with defined stages, a procedural algorithm described in prose).
\end{itemize}

\textbf{Strictness Rules}

\begin{itemize}[leftmargin=*,itemsep=0pt, topsep=0.3em]
    \item Mark a type as suitable \textbf{ONLY} if the text \textbf{clearly and unambiguously} provides what that type requires.
    \item A text may be suitable for \textbf{zero, one, or several} types.
    \item Rule-based hints (fast heuristic): \texttt{[rule\_hints]} --- treat as a starting point; override if the evidence in the text does not support a hint, or add a type the heuristic missed.
    \item \textbf{Do NOT} mark Type 1 unless the text contains actual spatial/positional structure, not just vague location words.
    \item \textbf{Do NOT} mark Type 2 if the text is primarily numeric, tabular, or highly fragmented.
    \item \textbf{Do NOT} mark Type 3 unless the numeric values are specific and extractable (e.g., ``37.5$^\circ$C'', ``142 patients'', ``\$2,380''), not vague estimates.
    \item \textbf{Do NOT} mark Type 4 unless the structured entries are clearly delineated with multiple attributes per entry.
    \item \textbf{Do NOT} mark Type 5 unless the sequential rules are explicit and complete enough to simulate step-by-step.
\end{itemize}

\textbf{Output Format}

Respond with \textbf{only} a valid JSON object, no other text:

\begin{verbatim}
{
  "suitable_types": [integers],
  "reasons": {
    "<type_number>": "<>"
  }
}
\end{verbatim}

\textbf{Text to Classify}

\texttt{\{text\}}
\end{promptbox}

Once a compatible passage is selected, we prompt the model to rewrite the original problem into a realistic exam question. The synthesis prompt is built around three core requirements. First, the \emph{corpus is the world}: the domain, entities, terminology, and contextual details should come from the paired FineWebEdu passage rather than from invented filler text. Second, the rewrite must contain \emph{zero programming traces}: variables, input/output phrasing, and algorithmic language are removed and replaced with natural prose. Third, the rewrite must preserve \emph{logical fidelity}: the mathematical structure, decision rules, and stopping conditions of the original problem remain unchanged, even though they are expressed through a realistic scenario.

Two additional constraints are particularly important for data quality. We require that \emph{all necessary information be embedded directly in the prose narrative}, so that the question is fully understandable without any contest-style input block. We also encourage \emph{rich contextual wrapping}: the passage should contribute substantial background detail, including irrelevant but realistic information, so that the result resembles a genuine exam reading task rather than a thinly disguised programming prompt. The full synthesis prompt is given below.

\begin{promptbox}
\textbf{Type-Aware Story-Wrapping Prompt} \\[4pt]
You are a top-tier exam question designer. Your task is to rewrite an algorithmic problem as a realistic written exam question that looks like it came from a human aptitude test, economics exam, or science assessment --- never from a programming contest.

You will receive:
\begin{itemize}
    \item A \textbf{corpus passage} (real-world text providing narrative context and potentially real data)
    \item An \textbf{original algorithmic problem} (defines the exact computation logic and I/O format)
    \item A \textbf{problem type} (tells you how to fuse the corpus with the logic)
\end{itemize}

\textbf{Universal Rules (apply to every type)}

\begin{enumerate}
    \item \textbf{The corpus is your world}: The scenario, characters, domain, and vocabulary of the exam question must come from the corpus. Never invent a fictional world when the corpus provides a real one.
    \item \textbf{Zero programming traces}: Ban all of --- variable names (\texttt{n}, \texttt{x}, \texttt{i}), ``input/output format'', ``algorithm'', ``time complexity'', ``given a positive integer'', ``print the answer''. Replace with natural prose.
    \item \textbf{Concrete nouns replace variables}: Instead of ``integer \texttt{n}'', write ``the number of monitoring stations'', ``the observation period in months'', ``the total number of affected residents'' --- nouns extracted or inspired by the corpus.
    \item \textbf{Logical fidelity}: The underlying mathematical model, decision rules, and stopping conditions must remain identical to the original problem. You may rephrase, never re-derive.
    \item \textbf{Self-contained question}: A test-taker with no programming knowledge should be able to read the question and understand exactly what is being asked.
    \item \textbf{ALL data embedded in prose (CRITICAL)}: Every numerical value, list, table, or parameter that the solver needs MUST appear directly within the problem narrative as natural text.
    \item \textbf{Rich context with distraction (CRITICAL)}: Incorporate a portion of the corpus passage into the problem, with some contextual detail not directly needed for solving.
    \item \textbf{Never reveal the mathematical model}: Do NOT write equivalences like ``this is equivalent to computing X'' or ``in other words, calculate $Y^n$''.
\end{enumerate}

\textbf{Type-Specific Fusion Strategy}

\textbf{Type \texttt{\{type\_id\}}} requires the following approach:

\texttt{\{type\_instruction\}}

\textbf{Output Format}

Respond with \textbf{only} a valid JSON object --- no markdown, no explanation:

\begin{verbatim}
{
  "problem": "<A realistic question>",
  "input": "<stdin for AC code>"
}
\end{verbatim}

IMPORTANT: The ``input'' field must contain values that are EXACTLY derivable from the ``problem'' text. A human reading only the ``problem'' field should be able to extract all the data needed and arrive at the same input values. The ``input'' field is only used internally for automated answer generation --- it is never shown to the test-taker.

\textbf{Inputs}

Corpus passage: \texttt{\{text\}}

Original algorithmic problem (logic + I/O format): \texttt{\{problem\}}
\end{promptbox}

Finally, for each source problem, we generate 20 story variants by pairing it with different compatible FineWebEdu passages, yielding roughly 200K synthesized questions. These variants share the same executable logic but differ in domain, discourse, and surface realization, which increases the diversity and generalizability of the resulting training data.

After the initial story-wrapping step, we perform a lightweight \emph{narrative refinement} pass to improve coherence. In early synthesis outputs, we frequently observe a ``two-block'' artifact: the model first reproduces a large chunk of corpus text, and then abruptly switches into the actual question through a transition such as ``consider the following case'' or ``in a separate task.'' Although the underlying logic is correct, this structure makes the result feel mechanically assembled rather than naturally written.

To address this, we apply a second-pass editing prompt that treats the initial synthesis as a draft and rewrites it into a more unified exam question. The goal is not to alter the computation, but to improve narrative flow: the background and the task should belong to the same story, each sentence should either motivate the question or provide relevant information, and the final wording should read like a coherent human-written exam item rather than a corpus passage followed by an attached exercise. Importantly, this refinement step is constrained to preserve the hidden \texttt{input} exactly, so the executable semantics and final answer remain unchanged.

Compared with the initial drafts, this refinement step substantially reduces redundant context and improves story--task integration. In our data, the average problem length decreases by approximately 35\% while narrative coherence improves noticeably. The full prompt is given below.

\begin{promptbox}
\textbf{Narrative Refinement Prompt} \\[4pt]
You are an expert exam question editor. You will receive a draft exam question that was machine-generated by fusing a real-world corpus passage with an algorithmic problem. Your job is to REFINE the draft into a polished, natural question where the background story and the computational task are seamlessly woven together.

You will receive:
\begin{itemize}
    \item The \textbf{draft problem} (the machine-generated question to improve)
    \item The \textbf{original input} (the stdin that produces the ground-truth answer --- you must preserve this exactly)
    \item The \textbf{corpus passage} (the original real-world text used for context)
\end{itemize}

\textbf{What's Wrong with Typical Drafts}

Most drafts suffer from a ``two-block'' structure:
\begin{enumerate}
    \item A long block of corpus text (copied almost verbatim)
    \item A sudden transition sentence (``In a separate task...'', ``Consider a specific...'', ``To better visualize...'') followed by the actual question
\end{enumerate}

This feels like two unrelated articles glued together. The reader can clearly see where the background ends and the real question begins.

\textbf{Your Refinement Goals}

\begin{itemize}[leftmargin=*,itemsep=0pt, topsep=0.3em]
    \item \textbf{Goal 1: Narrative Unity --- The story IS the problem.} The background and the question must be part of the same story. The reader should not feel a seam between context and task.
    \item \textbf{Goal 2: Causal Chain --- Every sentence earns its place.} Each piece of background should either explain why the problem exists or provide data needed to solve it. Remove irrelevant copied context.
    \item \textbf{Goal 3: Characters and Motivation --- Someone needs this answer.} You can introduce a concrete agent from the corpus world (e.g., researcher, inspector, technician, student, manager) who has a reason to need the result.
    \item \textbf{Goal 4: Appropriate Length.} The refined question should be 150--400 words. Do not pad with irrelevant detail.
    \item \textbf{Goal 5: Creative Integration over Literal Copying.} Rephrase, reorganize, and selectively use the corpus rather than copying large blocks verbatim.
\end{itemize}

\textbf{Hard Constraints (DO NOT VIOLATE)}

\begin{enumerate}[leftmargin=*,itemsep=0pt, topsep=0.3em]
    \item \textbf{The answer must not change.} The \texttt{input} must remain EXACTLY the same as the original.
    \item \textbf{All solving data must appear in the text.} Every value needed to derive the \texttt{input} must be explicitly stated in the problem narrative.
    \item \textbf{No programming traces.} No variable names, no ``input/output format'', no ``algorithm'', no ``given a positive integer''.
    \item \textbf{No revealing the math model.} Do not explain the hidden formula or say what the problem is ``equivalent to''.
\end{enumerate}

\textbf{Output Format}

Respond with \textbf{only} a valid JSON object --- no markdown, no explanation:

\begin{verbatim}
{
  "problem": "<refined question>",
}
\end{verbatim}

\textbf{Inputs}

Draft problem: \texttt{\{draft\_problem\}}

Corpus passage (for reference --- use selectively): \texttt{\{corpus\}}
\end{promptbox}

To illustrate the effect of Stage~2 synthesis and the subsequent refinement step, we present two representative examples below. We choose them because they capture two distinct but common failure modes in raw generation. The first is a Type~2 character-level problem, where naive synthesis often leaves a visible seam between a borrowed passage and the actual decoding task. The second is a Type~4 data-processing problem, where long descriptive context can easily overwhelm the operational question. In both cases, refinement preserves the original executable input while making the final question more coherent, natural, and exam-like.

\begin{examplebox}
\textbf{Case 1: Type~2 --- Character-Level Manipulation (Borze decoding)}\\
\textbf{Original problem.} Decode a Borze string, where \texttt{.}, \texttt{-.}, and \texttt{--} map to digits 0, 1, and 2.\\
\textbf{Corpus context.} Historical and industrial background of Épinal, France.

\medskip
\textbf{Before refinement.} The draft first presents a long encyclopedia-style description of Épinal, including its industries, wartime history, and cultural significance, and only then abruptly introduces a separate paragraph about archivists decoding an old telegraph log. Although the decoding task is correct, the story and the question feel like two unrelated texts pasted together.

\medskip
\textbf{After refinement.} The rewritten version turns the entire problem into a single archival scenario: a regional library in Épinal is digitizing early 20th-century telegraph records related to textile production, and the recovered string \texttt{-.--.-.-.} must be decoded using the local three-symbol code. This removes the visible transition between ``background'' and ``task'' and makes the character-level operation feel naturally motivated by the narrative.

\medskip
\textbf{Case 2: Type~4 --- Data Processing (frequency-based assignment)}\\
\textbf{Original problem.} Assign distinct prices to item types according to frequency so as to minimize and maximize the total cost.\\
\textbf{Corpus context.} Butterfly species records in Carson County, Texas.

\medskip
\textbf{Before refinement.} The draft contains a long ecological description of multiple butterfly species and habitats, followed by a sudden switch to a budgeting problem involving six grant amounts and a short observation log. The resulting question is solvable, but most of the biological background does not directly support the data-processing task, producing a clear narrative seam.

\medskip
\textbf{After refinement.} The refined version recasts the task as a conservation funding decision: county officials allocate distinct grant amounts to butterfly species based on how often each species appears in a field survey. The survey log and the six funding amounts are embedded directly in the narrative, and the computational objective---finding the minimum and maximum possible total allocation under different assignments---now arises naturally from the administrative scenario.

\medskip
In both examples, the refinement step preserves the hidden \texttt{input} exactly while substantially improving narrative unity and reducing the ``two-block'' artifact.
\end{examplebox}

\subsection{Stage 3: Test Case Synthesis and Narrative Input Fusion}
\label{appendix:stage3}

The story-wrapped questions generated in Stage 2 provide only one executable instance per problem, namely the input values embedded in the rewritten narrative. Test data obtained from real-world corpora is often overly simplistic; on the one hand, there is a lack of data within the corpus, and on the other, the model itself tends to favour simpler numbers.~(Details in Appendix~\ref{appendix:data_analysis}) This is insufficient for broad training coverage. In addition, while Codeforces problems are validated by official judge cases, those test cases are not always fully available in the released data, particularly the more challenging test cases. More importantly, many inputs that are useful for our setting~(such as adversarial symbolic strings, compact edge cases, or carefully structured numeric patterns) do not naturally appear in ordinary corpus passages. We therefore synthesize additional test inputs and then fuse them back into the existing story templates.

\paragraph{Guided random test synthesis.}
Our test generation process is guided by the accepted code rather than by uninformed random sampling. Specifically, we ask the LLM to read the AC solution and identify its key \emph{decision points}: branch conditions, comparison patterns, loop boundaries, special-case handling, and value configurations that may trigger qualitatively different execution paths. This analysis is performed by the LLM itself during generation. For each decision point, the model specifies both the triggering condition and a corresponding \emph{guided mode}, a structured pattern of inputs likely to hit that behavior reliably.

This setup is intended to uncover \emph{small but meaningful} logical vulnerabilities rather than to maximize raw computational burden. For example, the model may identify patterns such as singleton inputs, equal neighboring values, alternating structures, periodic layouts, or a single extreme outlier. These patterns are then turned into explicit test-generation strategies.

\paragraph{Three levels of test inputs.}
Using the decision-point analysis, we synthesize three levels of test inputs for each problem.

\textbf{Level A: Minimal sanity cases.}
These are small, mostly hard-coded cases designed to verify basic correctness. They typically use tiny sizes and simple values, and cover obvious edges such as singleton inputs, minimal valid structures, or degenerate small cases. We generate 5 such cases per problem.

\textbf{Level B: Hard-coded adversarial cases.}
These cases are constructed to hit the most important decision points directly. From the decision-point analysis, the model selects five critical branches and creates one deterministic case for each. These are intentionally non-random and are meant to trigger specific boundary behaviors, special cases, or branch-sensitive patterns. We generate 5 such cases per problem.

\textbf{Level C: Guided random cases.}
These cases provide broader coverage while remaining structured. The decision points are converted into a set of generation modes, such as \texttt{all\_equal}, \texttt{alternating}, \texttt{nearly\_sorted}, \texttt{large\_spike}, or \texttt{periodic}. Each mode has its own generator logic, and sampling rotates across modes to ensure diverse behavioral coverage. To keep execution efficient and avoid turning the task into unnecessarily heavy arithmetic, we cap input sizes at moderate ranges (e.g., $n \leq 50$, values $\leq 10^5$). We generate 10 guided-random cases per problem.

\textit{We omit the full prompt used for test synthesis. Unlike the other prompts in our pipeline, this prompt contains substantial program-specific content, including code formatting requirements, judge-dependent output templates, and execution-safety constraints, which make it considerably longer. As the central design principles have already been introduced in the preceding sections, reproducing the entire prompt here would add implementation detail without further clarifying the method.} To make the test-synthesis procedure more concrete, we provide one representative example below. This example illustrates how the model analyzes an accepted solution, identifies branch-sensitive decision points, and turns them into a small set of adversarial and guided-random test modes. Importantly, the goal is not to maximize raw input size, but to generate \emph{small, sharp} cases that are likely to expose logical mistakes in symbolic reasoning or state updates.

\begin{examplebox}
\textbf{Example: Decision-point-driven test synthesis for Codeforces 1426/F}

\textbf{Problem summary.} The input is a string over \texttt{a}, \texttt{b}, \texttt{c}, and \texttt{?}. Each \texttt{?} can be replaced by any of \texttt{a}/\texttt{b}/\texttt{c}. The task is to compute the total number of \texttt{abc} subsequences across all resulting strings, modulo $10^9+7$.

\medskip
\textbf{Core AC logic.} The accepted code maintains four DP states:
\begin{itemize}[leftmargin=*,itemsep=0pt, topsep=0.3em]
    \item \texttt{seq}: number of partial strings generated so far ($=3^{\#?}$),
    \item \texttt{a}: total number of \texttt{a} subsequences,
    \item \texttt{ab}: total number of \texttt{ab} subsequences,
    \item \texttt{out}: total number of \texttt{abc} subsequences.
\end{itemize}

For fixed characters, the update is straightforward:
\begin{itemize}[leftmargin=*,itemsep=0pt, topsep=0.3em]
    \item \texttt{'a'}: \texttt{a += seq}
    \item \texttt{'b'}: \texttt{ab += a}
    \item \texttt{'c'}: \texttt{out += ab}
\end{itemize}

The key difficulty lies in the \texttt{'?'} branch, where all three replacements must be merged carefully. In particular, the code must preserve the old value of \texttt{a} before scaling, and must use the old \texttt{ab} when updating \texttt{out}.

\medskip
\textbf{Identified decision points.} By reading the AC code, the model extracted five branch-sensitive behaviors:
\begin{enumerate}[leftmargin=*,itemsep=0pt, topsep=0.3em]
    \item the \texttt{'a'} branch, where \texttt{a += seq} must use the current number of partial strings;
    \item the \texttt{'b'} branch, where \texttt{ab += a} depends on the correct current value of \texttt{a};
    \item the \texttt{'c'} branch, where \texttt{out += ab};
    \item the \texttt{'?'} branch, where update order is crucial, especially the preservation of the old \texttt{a} and the use of the pre-scaled \texttt{ab};
    \item the initialization \texttt{seq = 1}, without which the entire DP collapses.
\end{enumerate}

\medskip
\textbf{Level A: minimal sanity-check cases.} The model first generated a small set of short inputs whose outputs are easy to verify by hand while still exercising distinct DP behaviors:
\begin{itemize}[leftmargin=*,itemsep=0pt, topsep=0.3em]
    \item \texttt{abc}: the simplest nonzero case, yielding exactly one \texttt{abc} subsequence,
    \item \texttt{acb}: a short reverse-order case, where the answer should be zero,
    \item \texttt{a?c}: checks whether a single wildcard can correctly complete the middle \texttt{b},
    \item \texttt{?bc}: checks whether a wildcard at the beginning can contribute the leading \texttt{a},
    \item \texttt{ab?}: checks whether a wildcard at the end can contribute the trailing \texttt{c}.
\end{itemize}

\medskip
\textbf{Level B: hard-coded adversarial cases.} Based on these decision points, the model constructed five deterministic test cases:
\begin{itemize}[leftmargin=*,itemsep=0pt, topsep=0.3em]
    \item \texttt{???}: pure wildcards; exposes incorrect update ordering in the \texttt{'?'} branch,
    \item \texttt{?a?}: checks whether \texttt{seq} is propagated correctly through wildcard expansion and then reused by a fixed \texttt{'a'},
    \item \texttt{a??}: verifies whether an early \texttt{a} is correctly expanded across multiple wildcard rounds,
    \item \texttt{?b?}: stresses the formation and accumulation of \texttt{ab},
    \item \texttt{?????}: tests multi-round wildcard updates and modular accumulation.
\end{itemize}

\medskip
\textbf{Level C: guided-random modes.} The model then converted the decision points into five reusable sampling modes~(The specific data is generated randomly under these patterns):
\begin{itemize}[leftmargin=*,itemsep=0pt, topsep=0.3em]
    \item \texttt{wildcard\_heavy}: about 70\% of positions are \texttt{'?'}, to stress repeated complex updates;
    \item \texttt{fixed\_sequence}: only \texttt{a}/\texttt{b}/\texttt{c}, to validate the base DP logic;
    \item \texttt{interleaved}: strict alternation between \texttt{'?'} and fixed characters, to force frequent switching between simple and complex branches;
    \item \texttt{prefix\_suffix}: wildcards concentrated near the beginning and end, to test boundary propagation;
    \item \texttt{reverse\_pattern}: strings dominated by \texttt{c}\,$\rightarrow$\,\texttt{b}\,$\rightarrow$\,\texttt{a} order, where the true count should remain very small and spurious overcounting is easier to detect.
\end{itemize}

\medskip
The generated lengths were intentionally restricted to roughly $3$--$50$, following our ``small but sharp'' principle: the synthesized tests are designed to stress logical correctness rather than brute-force numerical scale.
\end{examplebox}

\paragraph{Execution and gold outputs.}
Every synthesized input is executed against the original accepted code in a sandboxed environment to obtain the deterministic gold output. This guarantees that the supervision signal comes from verified program execution rather than from an LLM's own answer generation. We also retain available official examples or released test cases when present.

\paragraph{Narrative input fusion.}
After generating new inputs, we convert them into additional story-based questions by reusing the narrative templates from Stage~2. Given an existing story-wrapped problem and a newly synthesized input, we ask the model to rewrite the question so that the new values are naturally woven into the same scenario. This step is necessary because many synthesized inputs. For example, unusual symbol strings, boundary coordinates, or structured numeric sequences do not occur in the original corpus passage and must therefore be inserted explicitly into the prose.

The fusion prompt requires the model to identify which parts of the narrative correspond to the original executable input, replace them with the new values, and update all dependent textual details so that the story remains internally consistent. For example, if a new input changes the number of entities, then all related counts, names, and enumerations in the narrative must be updated as well. At the same time, the story world, characters, motivation, and question logic must remain unchanged. The result is a new natural-language question from which the updated executable input can again be fully recovered.

The full prompt used for this step is shown below.

\begin{promptbox}
\textbf{Narrative Input Fusion Prompt} \\[4pt]
You are an expert exam question editor. You will receive a story-based exam question (which already embeds some specific data values) and a NEW set of input data that must replace the original values. Your task is to rewrite the question so that the NEW data is naturally woven into the existing narrative, while preserving the story, characters, and scenario exactly as they are.

\vspace{4pt}
\textbf{What You Must Do}

\begin{enumerate}[leftmargin=*,itemsep=0pt, topsep=0.3em]
    \item \textbf{Identify the data points} in the original question that correspond to the original input (dimensions, counts, coordinates, strings, sequences, etc.)
    \item \textbf{Replace them} with the corresponding values from the new input, adjusting the narrative text naturally (e.g., ``a 6-meter by 6-meter room'' becomes ``a 10-meter by 8-meter room'' if the new input is \texttt{10 8 4})
    \item \textbf{Maintain narrative consistency}: when data changes affect quantities, update ALL related narrative elements accordingly. For example, if the number of items changes from 6 to 8, you must also update the count mentioned in prose, add or remove named entities to match, and ensure all enumerated items in the text match the new count.
    \item \textbf{Keep the core story unchanged}: the setting, characters' roles, scenario, motivations, and the question being asked must remain identical
    \item \textbf{Ensure all new data values appear explicitly in the text} --- a reader should be able to extract the new input from the rewritten question alone
\end{enumerate}

\textbf{Hard Constraints}

\begin{enumerate}[leftmargin=*,itemsep=0pt, topsep=0.3em]
    \item \textbf{The story and question logic must not change.} Only the embedded data values change.
    \item \textbf{Every value from the new input must appear in the output text.} Do not omit any data point.
    \item \textbf{No programming traces.} No variable names, no ``Input:'', no code formatting.
    \item \textbf{Preserve the narrative style and length} of the original question. Do not make it significantly shorter or longer.
    \item \textbf{If the new input contains strings or text}, embed them naturally as quotes, excerpts, or referenced content within the narrative.
\end{enumerate}

\textbf{Output Format}

Respond with \textbf{only} a valid JSON object:

\begin{verbatim}
{
  "problem": "<question with new data>"
}
\end{verbatim}

\textbf{Inputs}

Original story-based question: \texttt{\{problem\}}

Original input data (embedded in the question above): \texttt{\{original\_input\}}

NEW input data (must replace the original): {\{new\_input\}}
\end{promptbox}

By combining decision-point-driven test synthesis with narrative input, we expand each source problem from some simple story-based examples into a series of executable, more challenging and comprehensive test sets, thereby probing a wider range of logical behaviours.

\subsection{Stage 4: Code Instrumentation, Trace Collection, and CoT Synthesis}
\label{appendix:stage4}

The final stage of our pipeline constructs the \emph{output side} of the training data, i.e., high-quality reasoning trajectories aligned with the story-based questions produced earlier. A key observation is that accepted code naturally contains a form of process supervision: during execution, it passes through meaningful intermediate states, branch decisions, and variable updates that reflect how the problem is solved. Instead of relying on expensive human annotation, search-based reasoning distillation, or a much stronger teacher model to produce step-by-step solutions, we expose this latent process signal by instrumenting the code with sparse diagnostic breakpoints and then convert the resulting execution traces into natural-language chains of thought.

\paragraph{Sparse instrumentation over AC code.}
We begin from the original accepted solution and ask an LLM to insert purely observational diagnostic outputs. The instrumentation is strictly read-only: it may only add printing statements and must never modify, delete, or reorder any original code. All diagnostic outputs are prefixed with \texttt{[TRACE]} so that they can be separated from the program's official answer after execution.

The main design challenge is to expose enough intermediate structure to make the computation interpretable, while keeping trace volume manageable. We therefore encourage the model to insert trace points only at semantically meaningful locations:
\begin{itemize}
    \item \textbf{After preprocessing}: once sorting, grouping, graph construction, or other data preparation is complete;
    \item \textbf{When key variables are updated}: for example, when the current best value changes or when an important counter is increased;
    \item \textbf{At major branch decisions}: to record which path was taken and why;
    \item \textbf{At phase boundaries}: for multi-stage algorithms such as preprocess $\rightarrow$ optimize $\rightarrow$ aggregate;
    \item \textbf{Before final output}: to summarize the decisive quantities behind the answer.
\end{itemize}
Conversely, we avoid tracing every innermost-loop iteration, temporary helper variables, or computations that do not materially affect the final result. This makes the trace naturally sparse and more useful as a reasoning scaffold.

The full prompt used for code instrumentation is shown below.

\begin{promptbox}
\textbf{Code Instrumentation Prompt} \\[4pt]
You are a code analysis and debugging expert. Your task is to insert diagnostic output statements into a program that has passed all test cases (AC code), so that the program prints key intermediate computation results during execution. These intermediate results will help humans understand the program's complete problem-solving process.

You will receive:
\begin{itemize}[leftmargin=*,itemsep=0pt, topsep=0.3em]
    \item An \textbf{algorithm problem} (describing the problem background, input/output format, and examples)
    \item An \textbf{AC code} (a correct program that has passed all test cases)
\end{itemize}

\textbf{Core Principles}

\begin{enumerate}[leftmargin=*,itemsep=0pt, topsep=0.3em]
    \item \textbf{Read-only instrumentation}: You may only ADD output statements --- never modify, delete, or reorder any existing code. Every inserted statement must be purely observational.
    \item \textbf{[TRACE] prefix}: All inserted diagnostic output must start with \texttt{[TRACE]}.
    \item \textbf{Structured format}: Each TRACE line must follow the format \texttt{[TRACE] event\_label | key1=value1 | key2=value2}.
    \item \textbf{Output volume control}: No more than 20 TRACE lines total; never print unconditionally inside the innermost loop; prefer event-driven output.
    \item \textbf{Match the original code's output method}: use the same printing style as the original program whenever applicable.
\end{enumerate}

\textbf{Instrumentation Location Guidance}

The following are typically valuable observation points:
\begin{itemize}[leftmargin=*,itemsep=0pt,parsep=0.5em,topsep=0.3em,partopsep=0.3em]
    \item After input is fully read
    \item After data preprocessing
    \item When a key variable is updated
    \item At important branch decisions
    \item At algorithm phase transitions
    \item Before final output
\end{itemize}

You do not need to cover every category, select the 5--15 observation points most critical to understanding the specific algorithm.

\textbf{Output Format}

Output only a valid JSON object:

\begin{verbatim}
{
  "analysis": "<summarization>",
  "trace_plan": [
    {
      "location": "<insertion point>",
      "event": "<event label>",
      "variables": "<list of values>",
      "reason": ""
    }
  ],
  "instrumented_code": ""
}
\end{verbatim}

\textbf{Important:}
\begin{itemize}[leftmargin=*,itemsep=0pt, topsep=0.3em]
    \item instrumented\_code must be complete runnable code
    \item After filtering out all \texttt{[TRACE]} lines, the program's output must be identical to the original output
    \item Each entry in \texttt{trace\_plan} must correspond one-to-one with an actual TRACE statement in \texttt{instrumented\_code}
\end{itemize}

\textbf{Input}

Algorithm problem: \texttt{\{problem\}}

AC code:
\begin{verbatim}
{language}
{code}
\end{verbatim}
\end{promptbox}

\paragraph{Trace collection.}
We execute the instrumented programs on the verified test inputs synthesized in the previous stage. The raw stdout therefore contains two parts: the ordinary final answer and the inserted \texttt{[TRACE]} lines. Since instrumentation is read-only, filtering out all trace lines should recover exactly the original program output. This gives us execution-aligned intermediate states for the same inputs that were already used to construct story-based questions.

\paragraph{Trace-guided CoT synthesis.}
Given a story-based exam question, the gold answer, the instrumented code, and the raw execution output, we then synthesize a natural-language chain of thought. The process has two phases. First, we verify that the execution output is consistent with the known correct answer after removing the \texttt{[TRACE]} lines. If the answer does not match, we discard the sample rather than attempting to produce reasoning over an inconsistent trace. Second, if verification succeeds, we ask the model to generate a numbered reasoning chain in natural language.

The reasoning prompt explicitly requires the model to use the execution traces only as latent guidance, while expressing the final explanation entirely in the story's language. Each step should perform exactly one logical operation, such as identifying relevant quantities, making a local observation, carrying out a single calculation, or drawing an intermediate conclusion. The last step must state the final answer clearly. This format makes the resulting output suitable not only for SFT, but also for later RL use cases such as step-level reward assignment or process supervision.

The full prompt used for CoT synthesis is shown below.

\begin{promptbox}
\textbf{Trace-Guided CoT Synthesis Prompt} \\[4pt]
You are an expert reasoning assistant. You will receive a story-based exam question, the correct answer, the solution code (with diagnostic instrumentation), and the raw execution output from running that code on the given input.

Your job has two phases:

\vspace{4pt}
\textbf{Phase 1: Verify Output Correctness}

The raw execution output contains both \texttt{[TRACE]} diagnostic lines and the program's actual answer. Identify the actual answer by filtering out all \texttt{[TRACE]} lines.

Compare this actual answer against the correct answer:
\begin{itemize}
    \item For integers/strings: exact match after trimming whitespace
    \item For floating-point numbers: relative error $<1\%$
\end{itemize}

If they DO NOT match, respond with:
\begin{verbatim}
{
  "verified": false,
}
\end{verbatim}

Do not generate any reasoning in this case.

\vspace{4pt}
\textbf{Phase 2: Generate Step-by-Step Reasoning (only if verified)}

Write a reasoning chain list that explains how to solve the problem. Use the code logic and execution traces as your guide, but express everything in the story's language.

\textbf{Step Format}
\begin{itemize}[leftmargin=*,label=\textbullet,noitemsep,topsep=2pt]
    \item Each step performs exactly ONE logical operation
    \item Show the actual computation --- formulas, substitutions, arithmetic
    \item Every step must advance toward the final answer
\end{itemize}

\textbf{Using the Code and Traces}
\begin{itemize}[leftmargin=*,label=\textbullet,noitemsep,topsep=2pt]
    \item Read the code to understand the algorithm's strategy
    \item Use \texttt{[TRACE]} lines to identify key intermediate results
    \item Select the most informative checkpoints and fill in the reasoning between them
    \item Translate all technical concepts into the story's terminology
\end{itemize}

\textbf{Final Step}
The last step must clearly state the final answer.

\vspace{4pt}
\textbf{Output Format}

\begin{verbatim}
{
  "verified": true,
  "num_steps": <#reasoning steps>,
  "reasoning": ["", "", ""]
}
\end{verbatim}

\vspace{4pt}
\textbf{What NOT to Do}
\begin{itemize}
    \item No programming terminology
    \item No references to the code or traces
    \item No mentions of Codeforces or competitive programming
    \item No hedging
\end{itemize}

\vspace{4pt}
\textbf{Inputs}

Exam question: \texttt{\{problem\}}

Correct answer: \texttt{\{answer\}}

Instrumented solution code:
\begin{verbatim}
{instrumented_code}
\end{verbatim}

Raw execution output (includes \texttt{[TRACE]} lines and the actual answer): \texttt{\{raw\_stdout\}}
\end{promptbox}

\paragraph{Resulting supervision signal.}
This stage yields reasoning traces that are grounded simultaneously in three sources: the executable correctness of the AC solution, the intermediate structure exposed by instrumentation, and the natural-language scenario established by the story-wrapped problem. The resulting data are particularly useful for downstream supervised fine-tuning and reinforcement learning, because they provide explicit multi-step decompositions that can be treated as full targets, partial trajectories, or step-level supervision units.
To illustrate the final output of our trace-guided reasoning pipeline, we show a representative positive example below. This case is particularly suitable because the execution trace exposes a compact but informative intermediate state, the sorted order of the text fragments, and the synthesized reasoning successfully translates that procedural signal into a natural-language explanation grounded in the story. The result is a coherent multi-step solution that remains faithful to both the executable logic and the narrative setting.

\begin{examplebox}
\textbf{Example: Trace-guided CoT synthesis for a story-based string ordering problem}

\textbf{Story-based question.} The Interslavic Committee is cataloging manuscript fragments into a single archive identifier. Five recovered fragment codes --- \texttt{zb}, \texttt{f}, \texttt{z}, \texttt{qk}, and \texttt{q} --- must be arranged so that their concatenation forms the lexicographically smallest possible string.

\medskip
\textbf{Execution trace highlights.}
\begin{verbatim}
[TRACE] n=5 | S=[zb,f,z,qk,q]
[TRACE] S=[f,qk,q,zb,z]
[TRACE] result_string=fqkqzbz
[TRACE] answer=fqkqzbz
\end{verbatim}

\textbf{Synthesized CoT.}
\begin{itemize}[leftmargin=*,itemsep=0pt, topsep=0.3em]
    \item \textbf{Step 1:} The five fragment codes are \texttt{zb}, \texttt{f}, \texttt{z}, \texttt{qk}, and \texttt{q}, and they must be ordered into one final archive identifier.
    \item \textbf{Step 2:} The correct strategy is to decide the order of two fragments by comparing the two possible concatenations. For fragments \texttt{x} and \texttt{y}, if \texttt{xy} is lexicographically smaller than \texttt{yx}, then \texttt{x} should appear first.
    \item \textbf{Step 3:} Applying this rule to \texttt{f} and \texttt{qk}, we compare \texttt{fqk} with \texttt{qkf}. Because \texttt{fqk} is smaller, \texttt{f} comes before \texttt{qk}.
    \item \textbf{Step 4:} Next, compare \texttt{qk} with \texttt{q}. Here \texttt{qkq} is smaller than \texttt{qqk}, so \texttt{qk} must come before \texttt{q}.
    \item \textbf{Step 5:} Then compare \texttt{q} with \texttt{zb}. Since \texttt{qzb} is smaller than \texttt{zbq}, \texttt{q} should come before \texttt{zb}.
    \item \textbf{Step 6:} Finally, compare \texttt{zb} with \texttt{z}. Because \texttt{zbz} is smaller than \texttt{zzb}, \texttt{zb} should come before \texttt{z}. These pairwise decisions determine the full order: \texttt{f}, \texttt{qk}, \texttt{q}, \texttt{zb}, \texttt{z}.
    \item \textbf{Step 7:} Joining the fragments in this order gives the smallest possible archive identifier: \texttt{fqkqzbz}.
\end{itemize}

This example shows how sparse execution checkpoints can be converted into a human-readable reasoning chain without exposing programming terminology. The trace provides the hidden procedural scaffold, while the final CoT remains fully grounded in the archive-management narrative.
\end{examplebox}


\section{Evaluation Setting}
\label{appendix:eval_setting}

\subsection{Evaluation Framework}
We conduct all evaluations using the \texttt{lm-evaluation-harness} framework~\cite{eval-harness}, which provides standardized task implementations, prompt templates, answer extraction, and metric aggregation. All experiments are executed via the \texttt{vLLM}~\citep{vllm} backend to ensure efficient batched generation and consistent decoding behavior across models. For code benchmarks (HumanEval, MBPP, and CRUXEval), we use a separate execution-based evaluation pipeline with sandboxed code execution and unit-test verification according to their official repository.

\subsection{Benchmarks}
To comprehensively evaluate both the reasoning abilities targeted by our method and potential side effects on other capabilities, we select benchmarks spanning three categories: \emph{mathematical reasoning}, \emph{general reasoning and knowledge} and \emph{code generation}. Table~\ref{tab:bench_hf} provides a summary of all benchmarks.

\newcommand{\hfdata}[2]{\href{https://huggingface.co/datasets/#1}{\texttt{#2}}}

\newcolumntype{C}[1]{>{\centering\arraybackslash}p{#1}}

\begin{table*}[htbp]
\centering
\small
\setlength{\tabcolsep}{5pt}
\renewcommand{\arraystretch}{1.05}
\begin{tabular}{C{1.6cm} C{2.4cm} p{6.2cm} p{4.6cm}}
\toprule
\textbf{Benchmark} & \textbf{HF Dataset} & \textbf{Dataset Summary} & \textbf{Primary Abilities Tested} \\

\midrule
\multicolumn{4}{l}{\textit{Mathematical Reasoning}} \\
\midrule

GSM8K &
\hfdata{openai/gsm8k}{gsm8k} &
Grade-school math word problems (1,319 test) requiring 2--8 steps of elementary arithmetic~\cite{cobbe2021gsm8k}. &
Multi-step arithmetic reasoning; text-to-equation mapping. \\

GSM-Plus &
\hfdata{qintongli/GSM-Plus}{GSM-Plus} &
Adversarially augmented GSM8K~\cite{li2024gsmPlus} with numerical perturbation, structural rephrasing, and distractors to reduce shortcut biases. &
Robust numerical reasoning under perturbation. \\

MATH &
\hfdata{hendrycks/competition_math}{competition\_math} &
5,000 competition-level problems across 7 subjects (algebra, geometry, number theory, etc.) drawn from AMC/AIME contests~\cite{hendrycksmath2021}. &
Formal math reasoning; symbolic manipulation; multi-step proofs. \\

AIME &
\hfdata{math-ai/aime24}{AIME\_2024} \hfdata{math-ai/aime25}{AIME\_2025} &
60 problems from AIME 2024 and AIME2025 (I+II)~\cite{zhang2023aime}; integer answers in 0--999 requiring creative multi-step reasoning in algebra, combinatorics, and geometry. &
Competition-level mathematical problem solving. \\

\midrule
\multicolumn{4}{l}{\textit{General Reasoning \& Knowledge}} \\
\midrule

BBH &
\hfdata{lukaemon/bbh}{bbh} &
23 challenging BIG-Bench tasks where LMs previously fell below human performance~\cite{suzgun2023BBH}, spanning logical, symbolic, linguistic, and algorithmic problem types. &
Logical reasoning; compositional generalization; symbolic manipulation. \\

ARC-C &
\hfdata{allenai/ai2_arc}{ai2\_arc} &
Challenge split (1,172 questions) of the AI2 science exam benchmark~\cite{arc_challenge}, filtering for questions that defeat retrieval-based methods. &
Scientific reasoning; commonsense knowledge; multi-hop inference. \\

DROP &
\hfdata{EleutherAI/drop}{drop} &
Reading comprehension requiring discrete operations (addition, counting, sorting) over paragraphs~\cite{dua2019dropreadingcomprehensionbenchmark}. &
Numerical reasoning over text; discrete arithmetic over passages. \\

GPQA &
\hfdata{Idavidrein/gpqa}{gpqa} &
Graduate-level ``Google-proof'' science questions (physics, chemistry, biology) written by domain experts~\cite{rein2023gpqa}. &
Expert-level scientific reasoning; deep domain knowledge. \\

MMLU-Pro &
\hfdata{TIGER-Lab/MMLU-Pro}{MMLU-Pro} &
Enhanced MMLU with 10 options and stronger contamination resistance across academic and professional subjects~\cite{wang2024mmlupro}. &
Broad knowledge; domain generalization; fine-grained discrimination. \\

MIMIC &
Ours &
4,695 held-out problems from our pipeline, stratified by difficulty rating, problem type (5 categories), and source level (A/B/C tiers). &
Character-level operations; precise computation; rule-based simulation. \\

\midrule
\multicolumn{4}{l}{\textit{Code Generation}} \\
\midrule

HumanEval &
\hfdata{openai/openai_humaneval}{humaneval} &
164 hand-crafted Python problems with docstring specifications~\cite{humaneval}; correctness verified by executing against hidden unit tests. &
Code generation; algorithmic problem solving. \\

MBPP &
\hfdata{google-research-datasets/mbpp}{mbpp} &
Sanitized split of crowd-sourced Python programming tasks with test cases~\cite{austin2021program_mbpp}. &
Code synthesis from natural language; test-driven correctness. \\

CRUXEval &
\hfdata{cruxeval-org/cruxeval}{cruxeval} &
800 Python functions with input prediction and output prediction tasks~\cite{gu2024cruxeval}. &
Code comprehension; execution tracing; I/O reasoning. 
\textbf{We use its input predition problem set which is the most difficult} \\

\midrule
\multicolumn{4}{l}{\textit{Character-Level \& Spatial Reasoning}} \\
\midrule

CharBench &
\hfdata{omriuz/CharBench}{CharBench} &
A character-level benchmark with 175K instances across four tasks: character counting, frequency counting, and first/last occurrence finding~\cite{uzan2026charbench}. &
Sub-token character perception; counting under tokenization; positional indexing. \\

StringBench &
\href{https://github.com/wxl-lxw/StringLLM}{\texttt{StringLLM}} &
A string manipulation benchmark with 49 atomic tasks derived from Python and 1,462 composite tasks combining multiple operations~\cite{wang2025stringllm}. &
String-level manipulation; character mapping; slicing and indexing; compositional string operations. \\

MazeBench &
\hfdata{Menlo/Maze-Bench-v0.2}{Maze-Bench} &
A spatial reasoning benchmark of 100 tokenized 5$\times$5 mazes~\cite{dao2025alphamaze}. Models must output a valid move list from origin to target without traversing walls. &
Spatial reasoning; path planning; grid navigation; sequential action generation under constraints. \\

\bottomrule
\end{tabular}

\caption{Benchmarks used in our evaluation, grouped into four categories.}
\label{tab:bench_hf}
\end{table*}

\subsection{Metric Definition and Subset Aggregation}
\paragraph{Per-task accuracy.}
For each task (or subtask) with $N$ evaluation instances, we compute accuracy as:
\begin{equation}
\mathrm{Acc} = \frac{1}{N}\sum_{i=1}^{N}\mathbb{I}\left[\hat{y}_i = y_i\right],
\label{eq:acc}
\end{equation}
where $y_i$ is the ground-truth answer, $\hat{y}_i$ is the extracted model answer, and $\mathbb{I}[\cdot]$ is the indicator function.

\paragraph{\texttt{weight\_by\_size} aggregation for benchmarks with subsets.}
Some benchmarks (e.g., BBH, MMLU-Pro) consist of multiple subsets $\{s\}_{s=1}^{S}$ with different sizes $N_s$.
Following lm-evaluation-harness's \texttt{weight\_by\_size} strategy, we aggregate subset accuracies by weighting each subset by its number of examples:
\begin{equation}
\mathrm{Acc}_{\mathrm{weighted}} = \sum_{s=1}^{S} w_s \cdot \mathrm{Acc}_s,
\quad
w_s = \frac{N_s}{\sum_{j=1}^{S} N_j},
\label{eq:weight_by_size}
\end{equation}
where $\mathrm{Acc}_s$ is computed using Eq.~\eqref{eq:acc} on subset $s$.

\subsection{Inference Configuration}
We use \texttt{temperature=0.6}, \texttt{top\_p=0.95}, \texttt{top\_k=20}, \texttt{repetition\_penalty=1.05}, \texttt{min\_p=0}, and \texttt{max\_gen\_toks=4096}. These values follow the officially recommended decoding configuration for the \texttt{Qwen3} series \emph{think} models. For \texttt{Qwen3} series \emph{non-think} models, we use \texttt{temperature=0.7}, \texttt{top\_p=0.8}, \texttt{top\_k=20},  \texttt{repetition\_penalty=1.5}, and \texttt{max\_gen\_toks=4096}.
Since the \texttt{DeepSeek-R1-Distill-LLaMA} family does not provide an officially recommended inference setting for reasoning-focused decoding, we apply the same decoding configuration to ensure consistent evaluation across model families. As the tests involve randomness, all the results we present are the mean of six independent tests~(mean@6).

\section{Training Setting}
\label{sec:appendix_training}

This section provides comprehensive details of our training setup, including model specifications, hyperparameters, and dataset configurations for both the Supervised Fine-Tuning (SFT) and Reinforcement Learning (RL) stages.

\subsection{Base Models}

We conduct experiments on three instruction-tuned models of comparable scale, spanning two distinct reasoning paradigms: hybrid thinking/non-thinking models and reasoning-distilled models.

\paragraph{Model Selection.}
To evaluate the generalizability of our method across different model families and its scalability within a specific architecture, we select two representative dense models: \textbf{Qwen3}~\citep{yang2025qwen3} nonthink models and \textbf{DeepSeek-R1-Distill-Llama-8B}~\citep{guo2025deepseek}. 
Specifically, we employ two variants from the Qwen3 family (4B and 8B) to investigate the impact of model scaling. 
Meanwhile, DeepSeek-R1-Distill-Llama-8B, built on the Llama-3.1 architecture with distilled reasoning capabilities, serves as a distinct baseline to assess cross-architecture robustness. 
For consistent evaluation, we disable Qwen3's native ``thinking mode'' to focus on direct response generation, while retaining DeepSeek's inherent chain-of-thought behavior to compare different reasoning paradigms.

\subsection{Training Datasets}
\label{sec:appendix_datasets}

We construct three categories of training data for both SFT and RL stages:

\begin{enumerate}
    \item \textbf{\TheName{}}: Our core synthetic dataset derived from competitive programming problems. Each sample consists of a story-wrapped reasoning problem paired with a step-by-step chain-of-thought (CoT) solution. The training data is curated through our multi-stage pipeline (Section~\ref{appendix:pipeline}) and balanced across problem types, source difficulty levels, and problem identities. For SFT, this yields approximately 114K verified CoT samples; for RL, we select approximately 10K ``useful'' samples where the base model achieves partial correctness (1--7 out of 8 rollout samples correct) and balanced again by codeforces problem\_id, capability requirement type
 and test case source, ensuring the problems are neither too easy nor too hard for effective policy gradient learning.

    \item \textbf{Math}: For the SFT stage, we incorporate a widely-used open-source mathematical reasoning corpus, NuminaMath-CoT~\citep{numina_math_datasets}, which provides high-quality chain-of-thought solutions for mathematical problems across diverse difficulty levels. For the RL stage, we use the official mathematical reasoning datasets provided by the Verl~\citep{verl} framework, which include curated prompts from GSM8K~\citep{cobbe2021gsm8k} and MATH~\citep{hendrycksmath2021} training sets formatted for policy gradient training.

    \item \textbf{Mixed}: A combination of \TheName{} and Math data, constructed by randomly sampling equal portions from each source. This configuration is designed to study the complementary effects of code-based reasoning data and conventional mathematical reasoning data. As SFT suffers from significant catastrophic~\citep{sft_diversity, chu2025sft} forgetting when using single-source data, we use the mixed dataset by default in SFT.
\end{enumerate}

\subsection{Supervised Fine-Tuning (SFT)}
\label{sec:appendix_sft}

\paragraph{Framework.} We use LLaMA-Factory~\citep{zheng2024llamafactory} for all SFT experiments. Training is distributed across 8 GPUs using DeepSpeed ZeRO Stage-3 for memory-efficient full-parameter fine-tuning.

\paragraph{Hyperparameters.} Table~\ref{tab:sft_hyperparams} summarizes the SFT training hyperparameters. All models share the same configuration except for model-specific chat templates (Qwen3 template for Qwen models, DeepSeek3 template for DeepSeek-R1-Distill-Llama-8B). For a fair comparison, we use checkpoints from approximately the same training step, around 1,500 steps, for evaluation.

\begin{table}[H]
\centering
\small
\begin{tabular}{lc}
\toprule
\textbf{Hyperparameter} & \textbf{Value} \\
\midrule
Fine-tuning type & Full parameter \\
Parallelism & DeepSpeed ZeRO-3 \\
Number of GPUs & 8 \\
Per-device batch size & 2 \\
Gradient accumulation steps & 8 \\
Effective batch size & 128 \\
Learning rate & $5 \times 10^{-6}$ \\
LR scheduler & Cosine \\
Warmup ratio & 0.1 \\
Number of epochs & 2 \\
Max sequence length & 8192 \\
Precision & BF16 \\
Gradient checkpointing & Enabled \\
Validation split & 1\% \\
Eval / save interval & 200 steps \\
\bottomrule
\end{tabular}
\caption{SFT training hyperparameters.}
\label{tab:sft_hyperparams}
\end{table}

\paragraph{Data Configurations.} Recent studies have shown that SFT on single-domain data leads to catastrophic forgetting of general capabilities and overfitting to narrow output formats~\citep{sft_diversity,chu2025sft}. To mitigate this, we adopt a mixed training strategy rather than training exclusively on \TheName{} data. Specifically, we randomly sample 114K samples from NuminaMath-CoT~\citep{numina_math_datasets} as mathematical reasoning data, matching the scale of our 114K \TheName{} samples. Our primary training configuration (\textbf{Mixed}) combines equal portions from both sources, yielding approximately 114K training instances in total. This balanced mixture ensures that the model benefits from our code-mediated reasoning data while maintaining robust mathematical reasoning and general instruction-following capabilities. In ablation studies, we also train on Math-only configurations to verify that the mixed data outperforms single math domain source alone. And We also trained on original Codeforces problems (using LLM to replace the variables with real test cases, but without the story context) to illustrate the significance of our diverse contexts and synthesis.

\subsection{Reinforcement Learning (GRPO)}
\label{sec:appendix_grpo}

\paragraph{Framework.} We use Verl~\citep{verl} (Volcano Engine Reinforcement Learning for LLMs) for all RL experiments with Group Relative Policy with Optimization (GRPO)~\citep{shao2024deepseekmath}.

\paragraph{GRPO binary Reward Function.} We use a rule-based reward function with flexible answer matching. For each prompt, the model generates $n=8$ rollout responses. Each response is scored 1 (correct) or 0 (incorrect) by comparing the extracted answer against the ground-truth output produced by the AC code. The answer extraction pipeline applies three strategies in sequence: (1) exact string match, (2) substring containment check, and (3) numeric equivalence comparison (for mathematical answers).

\paragraph{Code-Instrumented Process Reward (CIR).}
\label{sec:appendix_cir}
In addition to the binary reward described above, we investigate a denser reward variant that leverages the multi-step structure of our code-synthesized problems. Since each training problem is derived from a competitive programming solution with well-defined intermediate execution checkpoints (e.g., input parsing, constraint validation, core algorithm steps, output formatting), we can assess \emph{how far} a model's response progresses through the intended reasoning chain, even when the final answer is incorrect.

Concretely, we prompt a lightweight judge model (Qwen3-4B) with the model's response alongside the reference solution's intermediate outputs, and ask it to determine which step the response has reached. The reward is then computed as:
\begin{equation}
    r = \frac{k}{K},
\end{equation}
where $k$ is the number of correctly completed steps identified by the judge and $K$ is the total number of steps in the reference solution. This provides a graded signal in $[0, 1]$ rather than a binary $\{0, 1\}$, enabling the policy to receive partial credit for partially correct reasoning trajectories.

To facilitate step-level evaluation across all reward strategies, we instruct the model to delimit its reasoning steps with explicit \texttt{[STEP]} markers during generation. This structured output format allows external PRMs to identify and score individual reasoning steps consistently.

\paragraph{Baseline PRM Configurations.}
For comparison, we evaluate two established process reward models under the same GRPO training setup:
\begin{itemize}[leftmargin=*,noitemsep,topsep=2pt]
    \item \textbf{PRM800K}~\cite{processbench,prmlessons}: We use Qwen2.5-Math-7B fine-tuned on the PRM800K dataset, which contains approximately 800K step-level correctness annotations derived from GSM8K and MATH problems. At each \texttt{[STEP]} boundary, the model scores the reasoning step and we take the minimum step-level score as the trajectory reward, following standard PRM practice.
    \item \textbf{Math-Shepherd}~\cite{wang2024math}: We use the official \texttt{math-shepherd-mistral-7b-prm} checkpoint based on Mistral-7B. This model was trained on automatically generated step-level labels for mathematical reasoning. It scores each step by predicting the probability of a positive label (\texttt{+}) versus negative (\texttt{-}) at designated step boundaries (marked by a special Cyrillic delimiter token).
\end{itemize}
Both PRM models are hosted as inference services on a dedicated GPU and queried via HTTP during training. Importantly, both models were trained exclusively on mathematical reasoning data, which limits their ability to provide meaningful reward signals for the diverse, non-mathematical reasoning tasks in our \TheName{} training set (see Table~\ref{tab:reward_ablation} for results).

\paragraph{Hyperparameters.} Table~\ref{tab:grpo_hyperparams} summarizes the GRPO training configuration. All experiments use on-policy rollouts generated by the vLLM engine~\citep{vllm} with the same sampling parameters and we use the same step 190 for our evaluation for fair comparison.

\begin{table}[H]
\centering
\small
\begin{tabular}{lc}
\toprule
\textbf{Hyperparameter} & \textbf{Value} \\
\midrule
Algorithm & GRPO \\
Parallelism & FSDP \\
Number of GPUs & 8 \\
Rollout engine & vLLM \\
Rollout samples per prompt ($n$) & 8 \\
Train batch size & 128 \\
PPO mini-batch size & 32 \\
PPO micro-batch size per GPU & 1 \\
Actor learning rate & $1 \times 10^{-6}$ \\
KL loss coefficient & 0.001 \\
KL loss type & Low-variance KL \\
Max prompt length & 4096 \\
Max response length & 4096 \\
Precision & BF16 \\
Gradient checkpointing & Enabled \\
Rollout GPU memory utilization & 0.6 \\
\bottomrule
\end{tabular}
\caption{GRPO training hyperparameters.}
\label{tab:grpo_hyperparams}
\end{table}

\paragraph{RL Dataset Configurations.} We construct three RL dataset variants using the verl parquet format:


\paragraph{Data Selection for RL.}
We construct the \TheName{} RL set from the synthesized data under a lexicographic objective. We first prioritize balance across \texttt{problem\_id}, reasoning \texttt{type}, and test-case source, so that the selected set is as uniform as possible along these axes. Subject to this diversity constraint, we run 8 rollouts per instance and preferentially retain those whose empirical correctness is closest to 50\%, since such cases maximize reward variance and yield the strongest policy gradient signal. We further encourage source diversity by preferring instances from different difficulty levels of the same problem.

\section{Distribution of Question Types in the Original Dataset}
\label{appendix:data_analysis}
Table~\ref{tab:stage1_distribution_cleaned} summarizes the distribution of Codeforces problems across the five top-level categories in our first-stage taxonomy. The cleaned dataset contains 10,008 instances in total, and the distribution is markedly imbalanced. In particular, Type 5 (\textit{Strict Rule \& State Simulation}) and Type 3 (\textit{Precise Mathematical Computation}) dominate the dataset, accounting for 46.02\% and 40.64\% of all instances, respectively; together, they cover 86.66\% of the dataset. This indicates that Codeforces problems are primarily characterized by deterministic procedural reasoning and mathematically precise problem solving. By contrast, Type 2 (\textit{Sub-Token/Character Manipulation}) represents a much smaller but still nontrivial portion (6.47\%), while Type 4 (\textit{Large-Scale Data Processing}, 4.80\%) and Type 1 (\textit{Text-Encoded Visual/Spatial Reasoning}, 2.07\%) remain relatively underrepresented.

At the subtype level, several categories exhibit a clear head-heavy structure. Within Type 5, \textit{Iterative State Simulation} is the largest subtype, whereas Type 3 is comparatively more balanced across arithmetic, algebraic, and set/statistical reasoning. Type 2 is strongly dominated by \textit{Character Counting/Extraction/Replace}, and Type 4 is almost entirely concentrated in \textit{Aggregation \& Sorting}. Overall, these results suggest that Codeforces provides broad and dense coverage of symbolic, procedural, and mathematically grounded reasoning, but offers substantially less diversity in visual/spatial and layout-sensitive problem types.

This distribution is beneficial for both training and evaluation because it combines strong coverage of major reasoning types with explicit inclusion of smaller but capability-critical categories. In particular, although current models often underperform on string manipulation, precise computation, and text-encoded visual/spatial reasoning, our taxonomy ensures that each pre-defined fine-grained category is represented rather than omitted by the natural dominance of a few major types. As a result, the synthesized dataset provides broad capability coverage while still maintaining sufficient density in core categories such as mathematical computation and rule-based simulation, making it well suited for both scalable training and fine-grained diagnosis of model weaknesses.

\begin{figure}[t]
    \centering
    \includegraphics[width=\columnwidth]{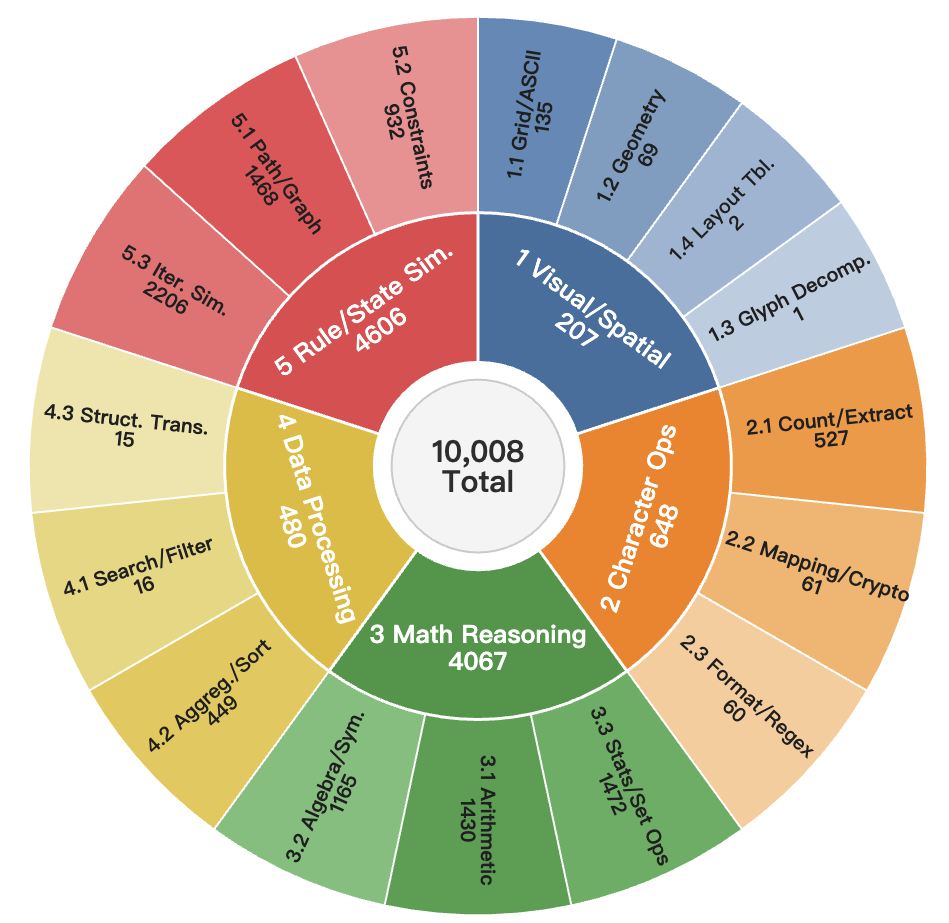}
    \caption{Hierarchical distribution of Codeforces problems in the Stage-I taxonomy. The inner ring denotes the five top-level categories and the outer ring denotes the corresponding fine-grained subtypes. Colors indicate top-level categories, while lighter shades represent their subtypes. Sector sizes are normalized within each level for visual clarity, and the actual numbers of instances are shown inside the segments.}
    \label{fig:stage1_taxonomy_coverage}
\end{figure}

To provide a more intuitive view of the scale and composition of our synthesized data, Figure~\ref{fig:stage1_taxonomy_coverage} visualizes the hierarchical distribution of Codeforces problems in the Stage-I taxonomy. Although the distribution is naturally skewed toward several major reasoning types, the figure shows that all pre-defined categories are covered, including smaller but capability-critical ones. This broad coverage helps ensure that the synthesized training and test sets span diverse reasoning abilities rather than concentrating only on a few dominant domains. Notably, such coverage is achieved using only a single publicly available source, Codeforces, which suggests that our taxonomy and synthesis framework are practical under current web-scale data conditions. In other words, the fact that one easily accessible dataset already provides near-complete coverage highlights both the resource efficiency and the scalability of our approach, while relatively sparse categories can be further supplemented when actually needed.

\begin{table*}[t]
\centering
\small
\setlength{\tabcolsep}{5pt}
\begin{tabular}{llrrr}
\toprule
\textbf{Top-level Category} & \textbf{Subtype (abbrev.)} & \textbf{Count} & \textbf{Top-level Share (\%)} & \textbf{Subtype Share (Parent, \%)} \\
\midrule

\multirow{4}{*}{\makecell[l]{Type 1: Text-Encoded\\ Visual/Spatial}} 
& 1.1 Grid/ASCII Parsing & 135 & \multirow{4}{*}{2.07} & 65.2 \\
& 1.2 Geom./Rel. Grounding & 69 &  & 33.3 \\
& 1.4 Layout-aware Tables/Diag. & 2 &  & 1.0 \\
& 1.3 Glyph Decomp. & 1 &  & 0.5 \\
\midrule

\multirow{3}{*}{\makecell[l]{Type 2: Sub-Token/\\ Character}} 
& 2.1 Char Count/Extract/Replace & 527 & \multirow{3}{*}{6.47} & 81.3 \\
& 2.2 Char Mapping/Crypto. & 61 &  & 9.4 \\
& 2.3 Formatting/Regex & 60 &  & 9.3 \\
\midrule

\multirow{3}{*}{\makecell[l]{Type 3: Precise\\ Math Computation}} 
& 3.3 Stats/Set Ops & 1472 & \multirow{3}{*}{40.64} & 36.2 \\
& 3.1 High-Prec. Arithmetic & 1430 &  & 35.2 \\
& 3.2 Algebra/Symbolic & 1165 &  & 28.6 \\
\midrule

\multirow{3}{*}{\makecell[l]{Type 4: Large-Scale\\ Data Processing}} 
& 4.2 Aggregation/Sorting & 449 & \multirow{3}{*}{4.80} & 93.5 \\
& 4.1 Search/Filter & 16 &  & 3.3 \\
& 4.3 Struct. Transform. & 15 &  & 3.1 \\
\midrule

\multirow{3}{*}{\makecell[l]{Type 5: Rule \&\\ State Simulation}} 
& 5.3 Iterative Simulation & 2206 & \multirow{3}{*}{46.02} & 47.9 \\
& 5.1 Pathfinding/Graphs & 1468 &  & 31.9 \\
& 5.2 Constraints/Backtracking & 932 &  & 20.2 \\
\midrule
\textbf{Total} & -- & \textbf{10,008} & \textbf{100.00} & -- \\
\bottomrule
\end{tabular}
\caption{Hierarchical distribution of the cleaned Stage-I dataset. ``Top-level Share'' is computed over all 10,008 instances, and ``Subtype Share (Parent)'' denotes the proportion within each top-level category. Abbreviated subtype names are used for space efficiency.}
\label{tab:stage1_distribution_cleaned}
\end{table*}

\section{Preliminary Study of Problem's Difficulty and Code Rate}
\label{subsec:preliminary_format_study}

\subsection{Difficulty-Dependent Effects Across Ratings}
\begin{table}[H]
\centering
\small
\setlength{\tabcolsep}{2pt}
\begin{tabular}{lcccc}
\toprule
\textbf{Rating} & \textbf{Original} & \textbf{Refined} & \textbf{Fused} & \textbf{Synthesized} \\
\midrule
$<1200$       & 27.7 & 20.2 & 0.4 & 0.6 \\
1200--1599    & 46.3 & 34.9 & 0.6 & 0.8 \\
1600--1999    & 49.2 & 38.1 & 0.8 & 0.9 \\
$\geq 2000$   & 44.9 & 35.5 & 0.8 & 0.7 \\
\bottomrule
\end{tabular}
\caption{Code-writing rate in model reasoning across different Codeforces rating bands under different problem presentation formats. Lower is better.}
\label{tab:rating_code_rate}
\end{table}

\begin{table}[H]
\centering
\small
\setlength{\tabcolsep}{2pt}
\begin{tabular}{lccccc}
\toprule
\textbf{Rating} & \textbf{Original} & \textbf{Refined} & \textbf{Fused} & \textbf{Synthesized} & \textbf{$\Delta$} \\
\midrule
$<1200$       & 46.6 & 55.1 & 48.6 & 72.3 & $-6.6$ \\
1200--1599    & 32.2 & 44.4 & 45.8 & 68.6 & $+1.4$ \\
1600--1999    & 19.3 & 30.6 & 41.4 & 65.1 & $+10.8$ \\
$\geq 2000$   & 13.0 & 24.4 & 40.6 & 56.9 & $+16.2$ \\
\bottomrule
\end{tabular}
\caption{Accuracy (\%) across different Codeforces rating bands under different problem presentation formats. The last column $\Delta$ reports the accuracy gain of \textbf{Fused} over \textbf{Refined}.}
\label{tab:rating_accuracy}
\end{table}

We further analyze this effect across Codeforces difficulty ratings. Table~\ref{tab:rating_code_rate} reports the proportion of cases in which the model attempted to write or reconstruct code during reasoning, and Table~\ref{tab:rating_accuracy} reports the corresponding accuracy under different presentation formats. A clear trend emerges: for the \textbf{original} and \textbf{refined} formats, code-oriented behavior becomes much more frequent on harder problems, reaching nearly half of all cases in the 1200--1999 rating range. This suggests that as task difficulty increases, the model is increasingly likely to fall back on memorized competitive-programming patterns instead of directly reasoning about the answer. By contrast, both \textbf{fused} and \textbf{synthesized} formats keep the code rate consistently below 1\% across all rating bands, showing that the reformulated presentation effectively suppresses this coding-oriented shortcut regardless of problem difficulty.

More importantly, the effect of reformatting is not uniform across difficulty levels. On lower-rated problems, the model's native reasoning ability is often already sufficient, so additional narrative packaging may introduce some noise and slightly reduce performance. However, on medium-to-hard problems, the benefit becomes increasingly significant: the performance gap between \textbf{fused} and \textbf{refined} grows steadily with rating, from negative on problems below 1200 to a substantial gain on problems above 2000. This pattern indicates that for difficult problems, the bottleneck is not simply insufficient reasoning capacity, but rather the model's tendency to choose the wrong strategy---namely, attempting to recall source problems or reconstruct code. Once this strategy is suppressed, the model performs markedly better by returning to direct reasoning. In this sense, our story-style reformulation acts as a form of \emph{reasoning strategy correction}: instead of allowing the model to exploit coding priors, it forces the model onto a more faithful natural-language reasoning path.

These findings further clarify the role of our synthesis framework. The goal is not merely to paraphrase programming problems, but to redirect optimization away from competitive-programming recall and toward pure logical reasoning. This distinction is especially important for training: if original or lightly edited coding tasks are used directly, the model may improve mainly by strengthening coding-related retrieval and solution-template behavior. In contrast, our fused-and-structured synthesis pipeline encourages the model to solve problems through explicit reasoning over natural-language scenarios and carefully constructed test cases, which is better aligned with the capabilities we aim to evaluate and improve.

\subsection{Performance Across Taxonomy Types}
\begin{table}[H]
\centering
\small
\setlength{\tabcolsep}{4pt}
\begin{tabular}{lccc}
\toprule
\textbf{Type} & \textbf{Accuracy (\%)} & \textbf{Total} & \textbf{Code Rate (\%)} \\
\midrule
Visual/Spatial & 37.2 & 2328  & 0.5 \\
Char-Level     & 47.2 & 8045  & 1.2 \\
Math           & 44.7 & 54121 & 0.9 \\
Data Process   & 53.5 & 6011  & 0.5 \\
Rule/State     & 37.2 & 57090 & 0.6 \\
\bottomrule
\end{tabular}
\caption{Model performance across top-level taxonomy types under the fused setting.}
\label{tab:type_performance}
\end{table}

We further analyze model performance across the major taxonomy types. As shown in Table~\ref{tab:type_performance}, the model performs poorly across all categories, with especially low accuracy on \textbf{Visual/Spatial} and \textbf{Rule/State} reasoning, both at only 37.2\%. These categories correspond to precisely the kinds of capability gaps that have repeatedly appeared in public discussions of strong language models: tasks such as determining left versus right in a mirror, comparing numbers like \texttt{9.9} and \texttt{9.11}, or counting the number of \texttt{r}'s in \texttt{strawberry}. What makes these problems difficult is not the need for broad world knowledge, but the need for \emph{fine-grained symbolic fidelity} as careful structural interpretation, exact local manipulation, and faithful multi-step state tracking. Our results suggest that these weaknesses are not isolated anecdotes, but reflect broader and still unresolved deficiencies in current models.

To make these type-level difficulty patterns more concrete, we show three representative examples from the kinds of data synthesized by our pipeline~(simplied for presentation in our paper). Although each example is simple for original program, all of them require the kind of fine-grained symbolic fidelity that remains fragile for current LLMs.

\begin{examplebox}
\textbf{Representative Examples of Fine-Grained Reasoning Tasks.}

\textbf{(1) Visual/Spatial Reasoning.}
A biologist tracks a lizard on a $2 \times 2$ grid whose cells are indexed by $(\text{row}, \text{column})$. The lizard starts at $(1,2)$. The grid is then transformed in sequence: (i) rotate clockwise by $90^\circ$ zero times, (ii) flip horizontally once, and (iii) rotate counterclockwise by $90^\circ$ once. The question asks for the lizard's final position. This kind of task is closely related to popular failure cases such as mirror-based left/right reasoning: the challenge is not high-level knowledge, but exact spatial transformation under symbolic descriptions.

\medskip
\textbf{(2) Precise Numerical Comparison.}
A shop lists five prices:
\[
17.02,\quad 7.18,\quad 7.8,\quad 10.1,\quad 10.05.
\]
For each of the following budgets,
\[
7.09,\quad 7.9,\quad 10.0,\quad 10.04,
\]
the model must determine how many items have price less than or equal to the budget. This is a representative variant of the well-known \texttt{9.9} vs.\ \texttt{9.11} phenomenon: although the required reasoning is elementary, it depends on exact digit-level numerical comparison rather than approximate magnitude matching.

\medskip
\textbf{(3) Character-Level String Reasoning.}
A security system uses the trigger word \texttt{silent} and scans the log entry \texttt{they are listening to the music}. The task is to determine whether the log contains a \emph{contiguous substring} that is a permutation of \texttt{silent}. If such a substring exists, the correct output is \texttt{YES}; otherwise, \texttt{NO}. This type of example captures the same core weakness as counting letters in words like \texttt{strawberry}: the model must reason over exact character composition and local string structure, rather than relying on semantic similarity or approximate lexical cues.
\end{examplebox}

This observation is exactly what motivates our data construction strategy. Rather than generating generic reasoning data, our method intentionally transforms competitive-programming problems into natural-language training instances that concentrate on those capability dimensions where programs are naturally reliable but language models remain fragile. In particular, visual/spatial tasks require models to reconstruct and manipulate text-encoded structures, while rule/state tasks require precise sequential simulation under strict constraints. Even categories such as \textbf{Math} and \textbf{Char-Level} remain far from saturated, showing that the resulting data targets fundamental reasoning weaknesses rather than narrow corner cases. Therefore, the type-level analysis supports the value of our approach as a \emph{training data synthesis method}: it can systematically produce supervision signals for the very forms of precise reasoning that current LLMs still struggle to master.

\subsection{Quality of Random Guided Synthetic Test Cases}
\begin{table}[htbp]
\centering
\small
\setlength{\tabcolsep}{7pt}
\begin{tabular}{lcccc}
\toprule
\textbf{Rating} & \textbf{Official} & \textbf{Level A} & \textbf{Level B} & \textbf{Level C} \\
\midrule
$<1200$       & 52.4 & 51.8 & 48.3 & 42.8 \\
1200--1599    & 44.8 & 51.9 & 48.0 & 38.4 \\
1600--1999    & 36.1 & 50.7 & 45.1 & 33.1 \\
2000--2399    & 31.5 & 52.0 & 46.0 & 32.2 \\
$\geq 2400$   & 24.1 & 48.4 & 42.2 & 28.0 \\
\bottomrule
\end{tabular}
\caption{Accuracy (\%) across different test sources and Codeforces rating bands.}
\label{tab:test_level_vs_official}
\end{table}

We also evaluate the quality of our automatically synthesized test cases produced by the code-to-text and key-point-guided test generation pipeline. Table~\ref{tab:test_level_vs_official} compares model accuracy on official tests and on our three synthesized levels. The results show that our generated test cases are consistently meaningful and non-trivial. In particular, \textbf{Level-C} often approaches the difficulty of the official tests and, for some rating ranges, remains broadly comparable to them. This is notable because our generated test cases are constructed under much more restricted settings as we only use relatively small constraints (e.g., $n \leq 50$ and values within $10^5$), whereas official tests are not limited in this way and may include more adversarial large-scale cases. Despite this limitation, the generated tests still induce similar failure patterns, demonstrating that our method captures critical reasoning bottlenecks rather than relying solely on scale or brute-force stress. These findings support the practical value of our pipeline: even in the absence of official tests, code-guided synthesis can produce test cases that are sufficiently challenging for evaluation and training. This makes our synthesis pipeline more scalable. 

\section{Token Distribution Shift}
\label{token_ditribution}
Following the methodology proposed by \cite{lin2023}, we conducted a comprehensive analysis of token distribution shifts between the base and models after GRPO and SFT with our reasoning and math benchmarks. The analysis focuses on identifying and quantifying the changes in token prediction patterns that occur during the alignment process.

Our analysis procedure consists of the following steps:

1) For each position in the input text, we use the aligned model with greedy decoding to generate the output token $o_t$.

2) We then examine how this token is ranked in the base model's probability distribution $P_{base}$. This ranking, denoted as $\eta$, serves as our primary metric for categorizing token shifts.

3) Based on the base ranking $\eta$, we classify each token position into three categories:
   \begin{itemize}
   \item Unshifted positions ($\eta = 1$): The token is top-ranked in both base and trained models
   \item Marginal positions ($1 < \eta \leq 3$): The token has a relatively high probability in the base model
   \item Shifted positions ($\eta > 3$): The token is unlikely to be sampled by the base model
   \end{itemize}

4) For shifted tokens, we calculate \textit{Rank Improvement Ratio}: $\frac{\text{base\_rank}}{\text{trained\_rank}}$

\begin{table}[H]
\centering
\small
\setlength{\tabcolsep}{7pt}
\begin{tabular}{lccc}
\toprule
\textbf{Method} & \textbf{Unshifted} & \textbf{Marginal} & \textbf{Shifted} \\
\midrule
SFT vs Base  & 78.9\% & 15.6\% & 5.5\% \\
GRPO vs Base & 95.1\% & 4.8\%  & 0.2\% \\
\bottomrule
\end{tabular}
\caption{Shifted token distribution under different alignment methods, measured against the base model. We report the proportion of unshifted, marginal, and shifted positions.}
\label{tab:shifted_token_distribution}
\end{table}

\begin{figure}[htbp]
\centering
\includegraphics[width=0.5\textwidth]{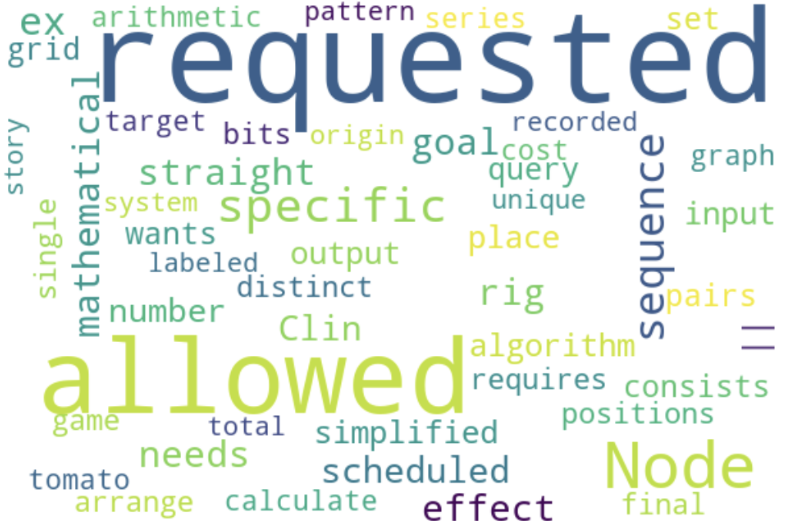}
\caption{Word cloud of shifted tokens under SFT, measured relative to the base model. The shifted vocabulary is dominated by instruction- and description-level terms such as \textit{requested}, \textit{allowed}, \textit{sequence}, \textit{algorithm}, and \textit{grid}, suggesting that SFT learns substantial surface-level regularities of the training data in addition to task behavior.}
\label{fig:sft_shifted_wordcloud}
\end{figure}

\begin{figure}[htbp]
\centering
\includegraphics[width=0.5\textwidth]{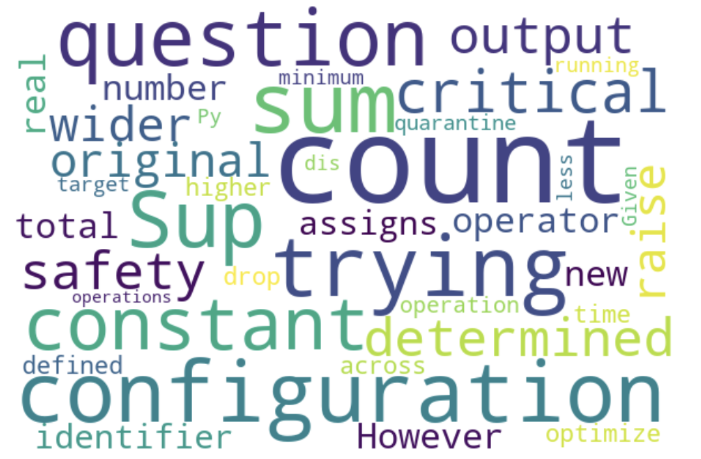}
\caption{Word cloud of shifted tokens under GRPO, measured relative to the base model. Although the number of shifted positions is much smaller, the shifted vocabulary is more concentrated on reasoning-relevant terms such as \textit{count}, \textit{sum}, \textit{configuration}, \textit{operation}, and \textit{determined}, indicating more targeted changes to computation and decision-making behavior.}
\label{fig:grpo_shifted_wordcloud}
\end{figure}

Table~\ref{tab:shifted_token_distribution} shows that both \textbf{SFT} and \textbf{GRPO} successfully modify the model's behavior on our synthesized training data, but they do so in very different ways. SFT causes a much broader redistribution of token preferences: only 78.9\% of positions remain unshifted relative to the base model, while 5.5\% become shifted. In contrast, GRPO is much more conservative, keeping 95.1\% of positions unshifted and changing only 0.2\% of positions in the shifted category. This indicates that both methods are effective, but GRPO achieves alignment with substantially smaller perturbations to the model's original token distribution.

The shifted vocabulary further reveals a qualitative difference between the two methods. As shown in Figure~\ref{fig:sft_shifted_wordcloud},~\ref{fig:grpo_shifted_wordcloud}, the shifted tokens under \textbf{SFT} are dominated by task-description and instruction-related words such as \textit{requested} and \textit{allowed}, as well as algorithmic or formatting-oriented terms such as \textit{Node}, \textit{sequence}, \textit{algorithm}, \textit{grid}, \textit{bits}, \textit{pairs}, \textit{mathematical}, \textit{simplified}, \textit{output}, and \textit{input}. This pattern suggests that SFT learns not only task-relevant behavior, but also the surface style of the training corpus---in particular, how to phrase solutions in benchmark-specific or instruction-like language. In other words, a substantial portion of SFT's token shift appears to reflect \emph{what to say} and \emph{how to present it}, rather than only \emph{how to reason}.

By contrast, the very small set of shifted tokens under \textbf{GRPO} is much more tightly concentrated on computation and reasoning-critical vocabulary. Frequent examples include arithmetic and counting terms such as \textit{count}, \textit{sum}, \textit{total}, and \textit{number}; operational terms such as \textit{configuration}, \textit{operation}, \textit{operator}, and \textit{optimize}; and logical or mathematical descriptors such as \textit{constant}, \textit{determined}, \textit{defined}, \textit{critical}, \textit{original}, and \textit{real}. These are not merely stylistic markers: they are closely tied to the internal structure of calculation, constraint tracking, and decision-making. This suggests that GRPO is not primarily teaching the model \emph{how to sound like} the training data, but rather making targeted adjustments to \emph{how the model carries out reasoning and computation}.

This distinction helps explain the different downstream behaviors of the two alignment methods. Because SFT induces broader token-level shifts, including many shifts on formatting-sensitive and description-level vocabulary, it is more prone to over-adapting to the surface regularities of the training distribution. Such over-specialization can improve in-domain imitation while reducing flexibility on evaluation sets that differ in phrasing or problem presentation. GRPO, in contrast, achieves gains through minimal but strategically placed changes, preserving the base model's general language competence while sharpening a small set of tokens that are disproportionately important for correct reasoning. In this sense, GRPO provides a more precise form of alignment: it does not broadly rewrite the model's language distribution, but selectively corrects the parts most relevant to mathematical and procedural reasoning.

\section{Training Dynamics}
\label{sec:training_dynamics}
\subsection{RL Training Dynamics Analysis}

To understand how GRPO training with \TheName{} data shapes model behavior, we analyze the training dynamics of our one of best-performing configuration (Qwen3-8B, $\mathcal{D}_\text{mixed}$) across 190 training steps. We examine four complementary perspectives: reward progression, policy divergence, response characteristics, and downstream benchmark evolution.

\begin{figure*}[t]
    \centering
    \begin{subfigure}[b]{0.32\textwidth}
        \includegraphics[width=\textwidth]{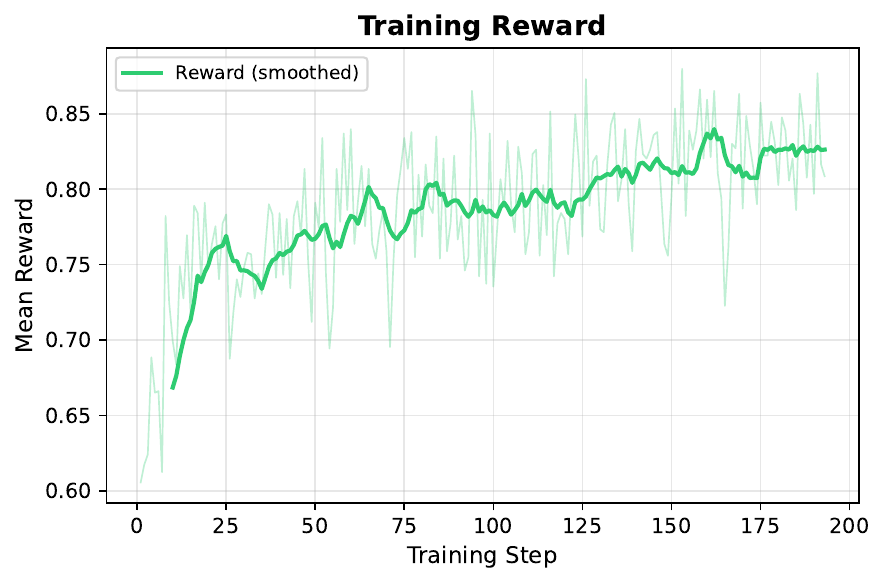}
        \caption{Training reward}
        \label{fig:dynamics_reward}
    \end{subfigure}
    \hfill
    \begin{subfigure}[b]{0.32\textwidth}
        \includegraphics[width=\textwidth]{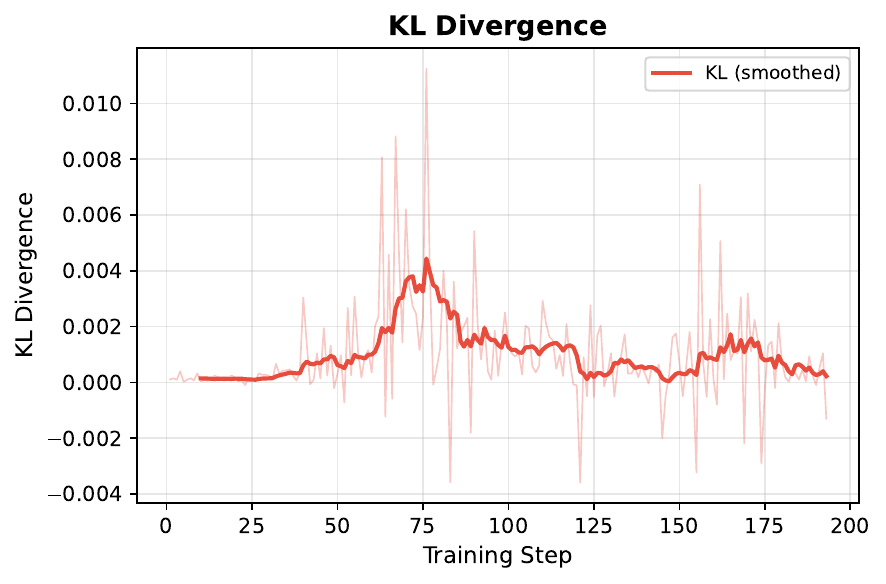}
        \caption{KL divergence}
        \label{fig:dynamics_kl}
    \end{subfigure}
    \hfill
    \begin{subfigure}[b]{0.32\textwidth}
        \includegraphics[width=\textwidth]{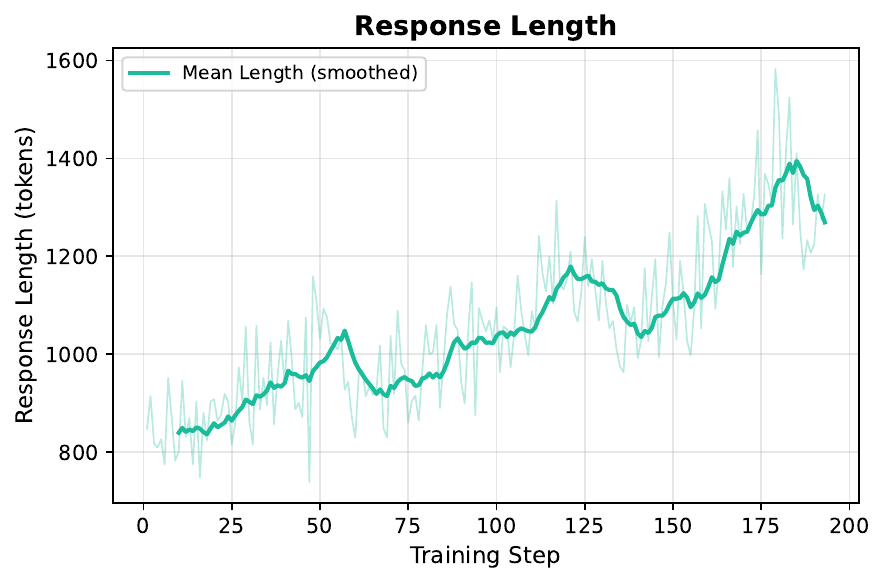}
        \caption{Response length}
        \label{fig:dynamics_length}
    \end{subfigure}
    \caption{\textbf{GRPO training dynamics} for Qwen3-8B on $\mathcal{D}_\text{mixed}$. (a)~Mean reward increases steadily from 0.61 to 0.81, indicating consistent learning. (b)~KL divergence remains bounded below 0.004 (smoothed), confirming stable policy updates without reward hacking. (c)~Mean response length grows from $\sim$850 to $\sim$1{,}330 tokens, reflecting the emergence of more thorough reasoning chains.}
    \label{fig:training_dynamics}
\end{figure*}

\begin{figure*}[t]
    \centering
    \includegraphics[width=\textwidth, trim=0 0 0 1cm, clip]{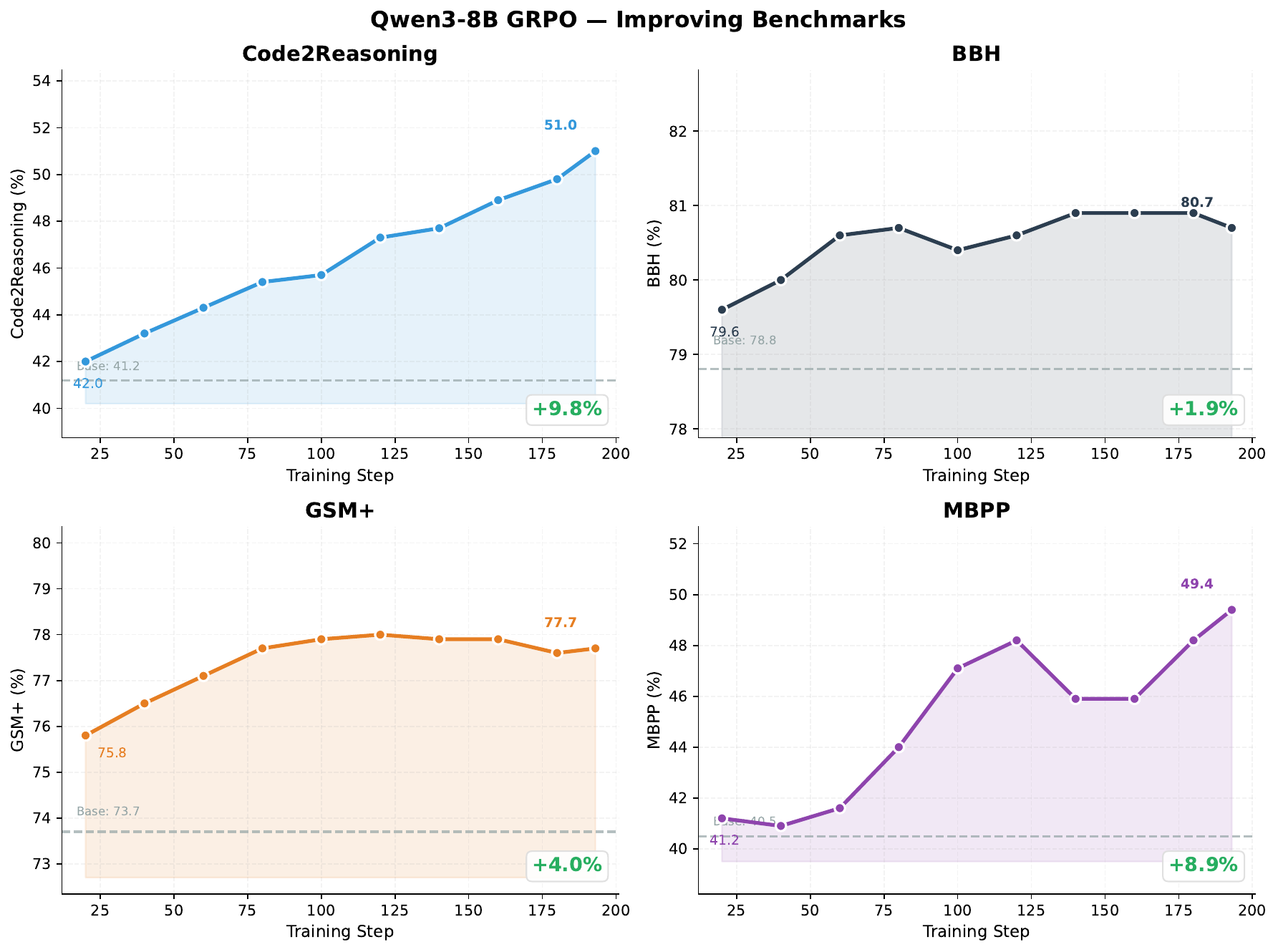}
    \caption{\textbf{Benchmark scaling during GRPO training} (Qwen3-8B, $\mathcal{D}_\text{mixed}$). All four benchmarks show consistent improvement over 190 training steps. Dashed lines indicate base model performance. \TheName{} improves by +9.8\%, demonstrating effective transfer from code-mediated reasoning to general problem-solving. Notably, MBPP (+8.9\%) also improves substantially, suggesting that reasoning-focused training can reciprocally enhance code generation capabilities.}
    \label{fig:eval_scaling}
\end{figure*}

\paragraph{Reward progression.}
Figure~\ref{fig:dynamics_reward} shows the mean reward steadily increasing from 0.61 to 0.81 over 190 steps. The learning curve exhibits two phases: a rapid improvement phase (steps 1--40) where the model quickly learns to solve easier problems, followed by a gradual refinement phase (steps 40--190) where gains become incremental as the model tackles harder problems. The absence of sudden jumps or collapses indicates stable optimization throughout training.

\paragraph{Policy stability.}
The KL divergence between the current policy and the reference model (Figure~\ref{fig:dynamics_kl}) remains remarkably small throughout training, staying below 0.004 in smoothed values. A brief exploratory spike around steps 60--80 is followed by a natural decline, suggesting that the model temporarily explored alternative reasoning strategies before settling into a stable policy. This bounded KL divergence, combined with our low KL penalty coefficient ($\beta = 0.001$), confirms that the model achieves meaningful capability gains without deviating excessively from its pretrained distribution---a property further validated by our token-level shift analysis (Section~\ref{token_ditribution}).

\paragraph{Response length evolution.}
The mean response length increases from approximately 850 to 1{,}330 tokens during training (Figure~\ref{fig:dynamics_length}). This growth is not indicative of \emph{length hacking} which is a known failure mode where models generate verbose outputs to exploit reward signals, since our reward function evaluates only the final answer against a ground-truth output, rather than explicitly rewarding longer responses. Moreover, the increase in response length is gradual and stable over training, and is accompanied by a consistent rise in reward. Taken together, these trends suggest that the model progressively learns to produce more elaborate reasoning chains, allocating additional tokens to intermediate computation steps rather than superficial elaboration.

\paragraph{Downstream benchmark evolution.}
Figure~\ref{fig:eval_scaling} tracks four representative benchmarks across training steps using external evaluation. \TheName{} demonstrates the most pronounced and consistent improvement (+9.8\%), rising from 41.2\% to 51.0\% without signs of saturation. This steady climb, in contrast to the rapid convergence of mathematical benchmarks observed during validation, suggests that code-mediated reasoning abilities develop gradually through sustained policy refinement rather than early memorization.

Beyond the target task, three additional benchmarks exhibit clear positive transfer: BBH improves by +1.9\%, reflecting enhanced complex reasoning; GSM-Plus gains +4.0\%, indicating better mathematical generalization; and MBPP increases by +8.9\%, demonstrating that strengthening reasoning capabilities can reciprocally benefit code generation. This cross-domain transfer supports our central thesis that code-mediated reasoning training produces broadly transferable cognitive improvements rather than narrow task-specific gains.

\subsection{SFT Training Dynamics Analysis}
\label{sec:sft_dynamics}

To contextualize the advantages of reinforcement learning for code-mediated reasoning, we analyze the SFT training dynamics under the similar configuration and model configuration (Qwen3-8B, $\mathcal{D}_\text{mixed}$).

\begin{figure}[t]
    \centering
    \includegraphics[width=0.5\textwidth]{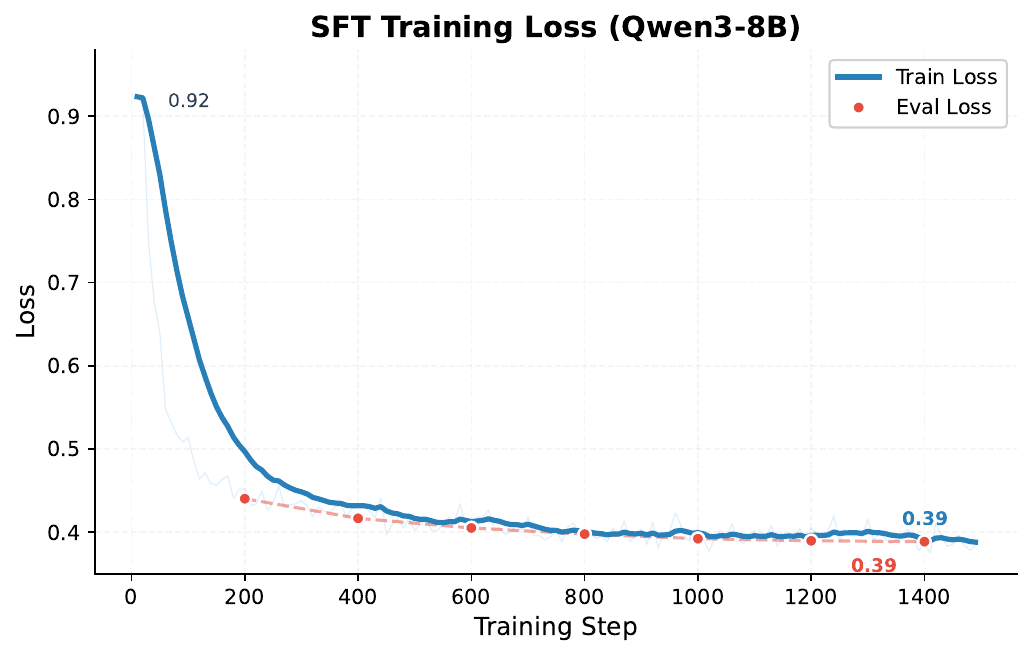}
    \caption{\textbf{SFT training and validation loss} for Qwen3-8B on $\mathcal{D}_\text{mixed}$. The training loss drops sharply from 0.92 to $\sim$0.45 within the first 400 steps, then gradually converges to 0.39. The validation loss closely follows, plateauing at 0.39 with minimal train-validation gap.}
    \label{fig:sft_loss}
\end{figure}

\begin{figure*}[t]
    \centering
    \includegraphics[width=\textwidth, trim=0 0 0 1cm, clip]{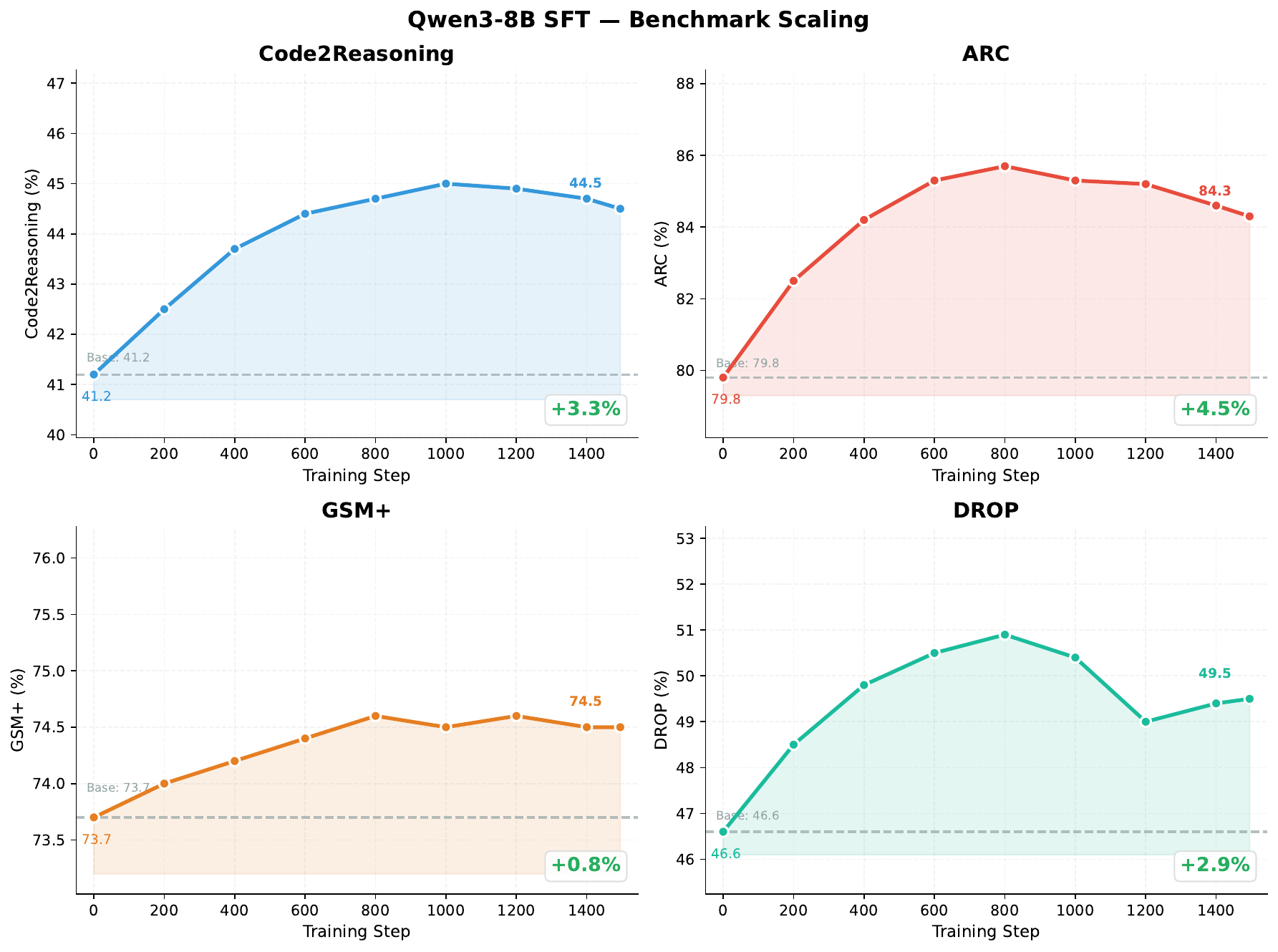}
    \caption{\textbf{Benchmark evolution during SFT} (Qwen3-8B, $\mathcal{D}_\text{mixed}$). Downstream benchmarks show quick initial gains that plateau or slightly decline after step 800, contrasting sharply with the continuous improvement observed under GRPO (Figure~\ref{fig:eval_scaling}).}
    \label{fig:sft_scaling}
\end{figure*}

\paragraph{Rapid convergence and early saturation.}
Figure~\ref{fig:sft_loss} reveals that SFT training loss drops sharply from 0.92 to approximately 0.45 within the first 400 steps, then gradually converges to 0.38 by the end of training (1{,}495 steps). The validation loss closely tracks the training loss, decreasing from 0.44 to 0.39, with minimal train-validation gap, indicating that the model is not heavily overfitting in the traditional sense. However, this rapid convergence masks a subtler issue: the model primarily learns to replicate the \emph{surface format} of the training data rather than acquiring deeper reasoning capabilities, as evidenced by the benchmark dynamics.

\paragraph{Benchmark dynamics reveal shallow learning.}
Figure~\ref{fig:sft_scaling} tracks four representative benchmarks across SFT training steps. A consistent pattern emerges: rapid initial improvement within the first $\sim$800 steps, followed by plateaus or mild regression. ARC-Challenge gains +5.5\% quickly as the model adapts to the answer format, while \TheName{}---our primary target---improves by only +3.3\% before saturating entirely.

This early saturation contrasts sharply with GRPO, where MIMIC accuracy rises continuously throughout training (+9.8\%, Figure~\ref{fig:eval_scaling}) without signs of plateauing. The gap is even more striking on code benchmarks: GRPO lifts MBPP by +8.9\%, whereas SFT provides negligible code improvement. These results suggest that SFT primarily teaches the model to mimic the \emph{surface format} of chain-of-thought demonstrations, while GRPO enables the model to discover and internalize genuine reasoning strategies that transfer across domains by optimizing for outcome correctness.

\paragraph{Why GRPO outperforms SFT for reasoning.}
The divergent training dynamics between SFT and GRPO illuminate a fundamental distinction in their learning mechanisms:

\begin{itemize}[leftmargin=1.5em,topsep=2pt,itemsep=1pt]
    \item \textbf{SFT saturates early}: all benchmarks converge within $\sim$800 steps, and the remaining training provides negligible improvement. The model learns the output distribution of the training data but does not develop novel reasoning strategies beyond what is explicitly demonstrated in the supervision signal.
    \item \textbf{GRPO improves continuously}: MIMIC accuracy rises monotonically throughout all GRPO steps without saturation (Figure~\ref{fig:eval_scaling}). Because GRPO optimizes for \emph{outcome correctness} rather than \emph{token-level imitation}, the model is free to explore diverse reasoning paths, retaining only those that yield correct answers.
    \item \textbf{Transfer breadth}: SFT improvements are concentrated in format-sensitive benchmarks (ARC: +5.5\%) while reasoning-heavy benchmarks improve modestly (MIMIC: +3.3\%). GRPO shows balanced gains across both reasoning (MIMIC: +9.8\%, BBH: +1.9\%) and mathematical benchmarks (GSM+: +4.0\%), alongside a surprising +8.9\% improvement in code generation (MBPP), indicating deeper capability transfer.
\end{itemize}

These observations align with recent findings that SFT tends to memorize output formats while RL methods elicit genuine reasoning improvements~\citep{chu2025sft}. Our results further demonstrate that this distinction is particularly pronounced for code-mediated reasoning tasks, where the solution space is too diverse for format imitation to be sufficient.

\section{Task Type Ablation Analysis}
\label{sec:appendix_type_ablation}


\begin{table*}[htbp]
\centering
\resizebox{\textwidth}{!}{
\begin{tabular}{@{}ll ccccccc ccccc cccc @{}}
\toprule
\multirow{2}{*}{\textbf{Model}} & \multirow{2}{*}{\textbf{Setting}}
  & \multicolumn{7}{c}{\textbf{General Reasoning}}
  & \multicolumn{5}{c}{\textbf{Math}}
  & \multicolumn{4}{c}{\textbf{Code}} \\
\cmidrule(lr){3-9} \cmidrule(lr){10-14} \cmidrule(lr){15-18}
& & ARC & BBH & MIMIC & GPQA & MMLU & DROP & \cellcolor{blue!8}\textbf{Avg}
  & GSM8K & GSM+ & MATH & AIME & \cellcolor{blue!8}\textbf{Avg}
  & HEval & CruxE & MBPP & \cellcolor{blue!8}\textbf{Avg} \\
\midrule

Qwen3-8B & Base & 79.8 & 78.8 & 41.2 & 44.4 & 70.5 & 46.6 & \cellcolor{blue!8}60.2 & 90.6 & 73.7 & 79.2 & 23.3 & \cellcolor{blue!8}66.7 & 22.6 & 70.6 & 40.5 & \cellcolor{blue!8}44.6 \\
Qwen3-8B & Ours (All) & 81.2 & 81.7 & 52.2 & 42.6 & 72.2 & 53.1 & \cellcolor{blue!8}63.8\up{3.6} & 91.7 & 78.1 & 81.0 & 26.7 & \cellcolor{blue!8}69.4\up{2.7} & 23.8 & 76.3 & 44.0 & \cellcolor{blue!8}48.0\up{3.4} \\

\midrule
\multicolumn{18}{c}{\textit{(A) Single-Type Training: GRPO with only one task category}} \\
\midrule

\multirow{5}{*}{Qwen3-8B}
 & Type 1 (Visual)      & 79.3 & 78.8 & 43.0 & \underline{45.2} & 71.1 & 50.8 & \cellcolor{blue!8}61.4\up{1.2} & 90.3 & 74.6 & 79.9 & \underline{26.7} & \cellcolor{blue!8}67.9\up{1.2} & \textbf{22.9} & 71.4 & 41.6 & \cellcolor{blue!8}45.3\up{0.7} \\
 & Type 2 (Char)        & 80.0 & 79.7 & 42.9 & \textbf{46.1} & \underline{71.4} & 51.5 & \cellcolor{blue!8}61.9\up{1.7} & \underline{92.0} & 75.4 & \underline{81.1} & \textbf{28.3} & \cellcolor{blue!8}\underline{69.2}\up{2.5} & \underline{22.9} & 70.4 & \underline{43.6} & \cellcolor{blue!8}45.6\up{1.0} \\
 & Type 3 (Math)        & 79.8 & \textbf{81.8} & \underline{52.2} & 43.2 & 71.1 & \textbf{52.9} & \cellcolor{blue!8}\underline{63.5}\up{3.3} & 90.8 & \underline{78.6} & \underline{81.1} & 23.3 & \cellcolor{blue!8}68.5\up{1.8} & 22.3 & \textbf{75.0} & \underline{43.6} & \cellcolor{blue!8}\textbf{47.0}\up{2.4} \\
 & Type 4 (Data)        & \underline{81.7} & 80.1 & 42.9 & 43.6 & 70.3 & 51.0 & \cellcolor{blue!8}61.6\up{1.4} & 90.5 & 75.3 & 80.8 & 16.7 & \cellcolor{blue!8}65.8\dn{0.9} & 21.7 & 69.8 & 41.6 & \cellcolor{blue!8}44.4\dn{0.2} \\
 & Type 5 (Rule)        & \textbf{82.6} & \underline{81.6} & \textbf{53.8} & 39.4 & \textbf{71.8} & \underline{52.4} & \cellcolor{blue!8}\textbf{63.6}\up{3.4} & \textbf{92.3} & \textbf{78.8} & \textbf{81.2} & 26.7 & \cellcolor{blue!8}\textbf{69.7}\up{3.0} & 22.3 & \underline{72.0} & \textbf{44.4} & \cellcolor{blue!8}\underline{46.2}\up{1.6} \\

\midrule
\multicolumn{18}{c}{\textit{(B) Leave-One-Out: GRPO with all categories except one}} \\
\midrule

\multirow{5}{*}{Qwen3-8B}
 & w/o Type 1           & 79.8 & 81.6 & \textbf{52.6} & 40.1 & \textbf{72.1} & 51.9 & \cellcolor{blue!8}63.0\up{2.8} & \textbf{93.3} & \textbf{78.7} & 79.6 & 20.0 & \cellcolor{blue!8}67.9\up{1.2} & \underline{22.9} & 72.5 & \underline{43.6} & \cellcolor{blue!8}\underline{46.4}\up{1.8} \\
 & w/o Type 2           & \underline{81.1} & \textbf{82.2} & 51.5 & \underline{43.0} & 71.5 & \underline{52.1} & \cellcolor{blue!8}\textbf{63.6}\up{3.4} & \underline{91.7} & 78.1 & \underline{81.0} & 23.3 & \cellcolor{blue!8}\underline{68.6}\up{1.9} & 22.9 & \textbf{75.9} & \textbf{44.0} & \cellcolor{blue!8}\textbf{47.6}\up{3.0} \\
 & w/o Type 3           & 77.8 & 80.9 & 49.2 & 41.4 & 70.9 & 51.8 & \cellcolor{blue!8}62.0\up{1.8} & 91.7 & 77.4 & \textbf{82.0} & 16.7 & \cellcolor{blue!8}67.0\up{0.3} & 21.1 & \underline{73.7} & 41.6 & \cellcolor{blue!8}45.5\up{0.9} \\
 & w/o Type 4           & \textbf{81.6} & \underline{82.0} & \underline{51.8} & 40.7 & 70.6 & \textbf{52.7} & \cellcolor{blue!8}\underline{63.2}\up{3.0} & 91.7 & \underline{78.5} & 80.7 & \textbf{30.0} & \cellcolor{blue!8}\textbf{70.2}\up{3.5} & 22.3 & 71.5 & 40.5 & \cellcolor{blue!8}44.8\up{0.2} \\
 & w/o Type 5           & 79.4 & 80.0 & 45.7 & \textbf{43.4} & \underline{71.8} & 51.7 & \cellcolor{blue!8}62.0\up{1.8} & 90.9 & 76.1 & 80.3 & \underline{26.7} & \cellcolor{blue!8}68.5\up{1.8} & \textbf{23.6} & 72.3 & 41.2 & \cellcolor{blue!8}45.7\up{1.1} \\

\bottomrule
\end{tabular}}
\caption{\textbf{Task Type Ablation (Qwen3-8B).} (A)~\textit{Single-type training}: GRPO trained with only one task category. (B)~\textit{Leave-one-out}: GRPO trained with all categories except one. Subscripts on \textbf{Avg} show absolute change vs.\ Base. \textbf{Bold} = best, \underline{underline} = second best within each section.}
\label{tab:type_ablation}
\end{table*}

To understand how each of the five task categories contributes to the final model performance, we conduct a systematic ablation study on Qwen3-8B using GRPO training. We design two complementary experiments:

\begin{itemize}[leftmargin=*,noitemsep,topsep=2pt]
    \item \textbf{Single-type training}: Train GRPO using only data from one task category (Type $i$ only), producing five models.
    \item \textbf{Leave-one-out (LOO)}: Train GRPO using data from all categories \emph{except} one (w/o Type $i$), producing five additional models.
\end{itemize}

All models are trained under identical hyperparameters and evaluated on the same benchmark suite. Full results are presented in Table~\ref{tab:type_ablation}.

\begin{figure}[H]
    \centering
    \includegraphics[width=0.5\textwidth]{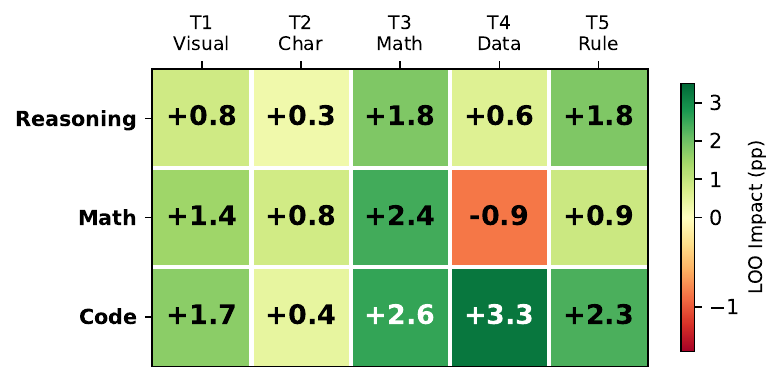}
    \caption{LOO Impact on category averages (Reasoning, Math, Code). Positive values (green) indicate beneficial contribution; negative values (red) indicate interference when included.}
    \label{fig:heatmap_avg}
\end{figure}

\begin{figure}[htbp]
    \centering
    \includegraphics[width=0.95\linewidth]{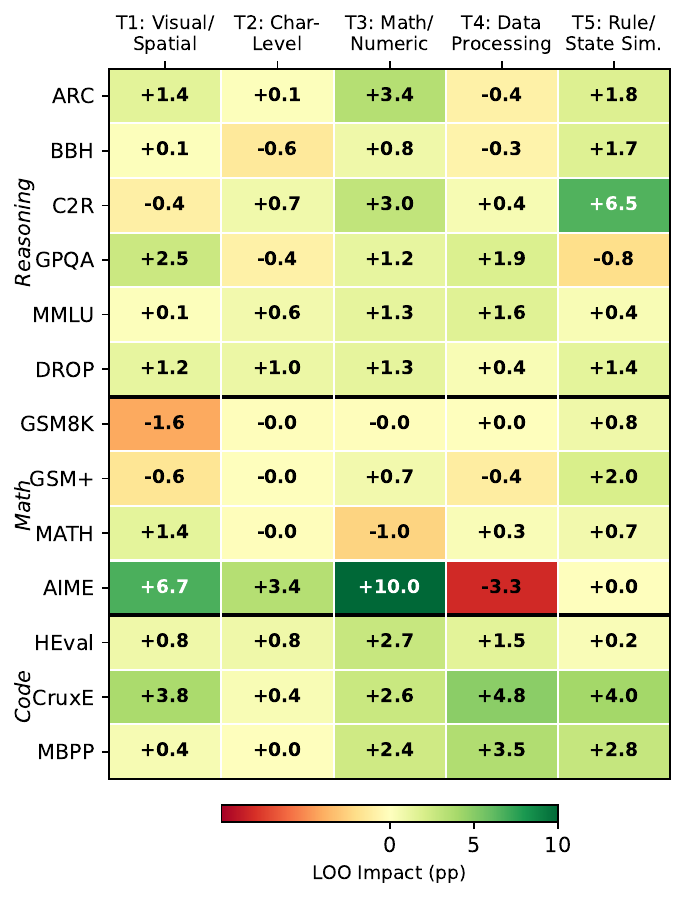}
    \caption{Per-benchmark LOO Impact. Each cell shows the accuracy drop (in percentage points) when the corresponding task type is removed from training. Horizontal lines separate the three evaluation categories: Reasoning, Math, and Code.}
    \label{fig:heatmap_full}
\end{figure}

\paragraph{LOO Impact metric.}
To quantify the marginal contribution of each task category, we define the \emph{LOO Impact} as the performance difference between the full model (trained on all five categories) and the leave-one-out model:
\begin{equation}
\small
    \Delta_{\text{LOO}}(\text{Type}_i, b) = \text{Score}_{\text{All}}(b) - \text{Score}_{\text{w/o Type}_i}(b),
\end{equation}
where $b$ is a benchmark or category average. A positive $\Delta_{\text{LOO}}$ indicates that removing Type $i$ hurts performance---i.e., Type $i$ is beneficial. A negative value indicates that removing it \emph{improves} performance, suggesting potential interference. We visualize the LOO Impact across all benchmarks in Figure~\ref{fig:heatmap_full} and across category averages in Figure~\ref{fig:heatmap_avg}.

\subsection{Per-Type Contribution Analysis}

\paragraph{Type 3 (Precise Mathematical Computation) is the most broadly impactful category.}
Across all three evaluation dimensions, Type 3 shows the highest or near-highest LOO Impact: Reasoning +1.8, Math +2.4, Code +2.6 (Figure~\ref{fig:heatmap_avg}). The per-benchmark breakdown (Figure~\ref{fig:heatmap_full}) reveals that this contribution is driven by large gains on AIME (+10.0 pp), ARC (+3.4), MIMIC (+3.0), and CruxEval (+2.6). This aligns with our expectation: tasks involving high-precision arithmetic, algebraic manipulation, and numerical verification develop reasoning primitives (careful step tracking, exact computation) that transfer broadly across evaluation benchmarks. Notably, the +10.0 pp impact on AIME confirms that competition-level mathematics particularly benefits from training on precise computation tasks. In the single-type experiment, Type 3 alone achieves Reasoning Avg 63.5 and Math Avg 68.5, closely approaching the full model's 63.8 and 69.4, respectively.

\paragraph{Type 5 (Strict Rule \& State Simulation) is the strongest single-type contributor to reasoning.}
When trained in isolation, Type 5 achieves the highest Reasoning Avg (63.6) and Math Avg (69.7) among all single-type models, and the highest single-type MIMIC score (53.8---surpassing even the full model's 52.2). The LOO analysis corroborates this: removing Type 5 causes the largest drop in MIMIC (-6.5 pp) and substantial drops in BBH (-1.7) and GSM+ (-2.0). Tasks in this category~(graph traversal, constraint satisfaction, iterative state simulation) require the model to maintain and update complex state across multiple reasoning steps, a skill directly tested by multi-step reasoning benchmarks like BBH (tracking shuffled objects, logical deduction) and our MIMIC benchmark. The strong MIMIC drop is particularly meaningful since MIMIC specifically targets the abilities our data aims to develop.

\paragraph{Type 1 (Visual/Spatial Reasoning) has an outsized impact on AIME.}
While Type 1's average contributions are moderate (LOO: R+0.8, M+1.4, C+1.7), the per-benchmark view reveals a striking +6.7 pp impact on AIME---second only to Type 3. AIME competition problems frequently involve geometric reasoning, coordinate manipulation, and spatial visualization, which directly align with Type 1's training tasks (grid parsing, coordinate movement, geometric constraints). Type 1 also shows +3.8 pp on CruxEval, suggesting that spatial reasoning skills help with code execution tracing. The +2.5 pp impact on GPQA (graduate-level science questions) is also notable, likely reflecting the spatial reasoning demands of physics and chemistry problems.

\paragraph{Type 2 (Character-Level Operations) provides consistent but modest gains.}
Type 2 shows the smallest LOO Impact across all dimensions (R+0.3, M+0.8, C+0.4), suggesting that character-level manipulation (counting, mapping, format validation) is the least critical category for standard benchmark performance. However, in the single-type experiment, Type 2 achieves the highest AIME score (28.3) among all single types, indicating that careful character-level attention may help with the precise symbolic manipulation required in competition mathematics. This category's contribution may also manifest more strongly on the specialized benchmarks (CharBench, StringBench) analyzed in Section~\ref{sec:experiment}.

\paragraph{Type 4 (Large-Scale Data Processing) shows a unique dual character.}
Type 4 exhibits the most interesting pattern: it has the \emph{highest} LOO Impact on Code Avg (+3.3 pp) but a \emph{negative} LOO Impact on Math Avg (-0.9 pp). The per-benchmark analysis reveals that this dual character stems from strong positive contributions to CruxEval (+4.8) and MBPP (+3.5), contrasted with a -3.3 pp impact on AIME. Tasks in this category involve search, filtering, aggregation, and structural transformation over large data, which are highly relevant to code comprehension and generation (CruxEval, MBPP) but may introduce noise for mathematical competition problems that require fundamentally different reasoning strategies. In the single-type experiment, Type 4 is the weakest single-type model (R: 61.6, M: 65.8, C: 44.4), performing below the base model on Math and Code averages, confirming that this category alone provides insufficient training signal. However, when mixed with other data, type 4 is crucial for coding ability.

\subsection{Cross-Type Interactions and Complementarity}

\paragraph{No single type matches the full mixture.}
While Types 3 and 5 individually approach the full model's Reasoning and Math averages, no single type matches the full model across \emph{all three} dimensions simultaneously. The full model achieves (R: 63.8, M: 69.4, C: 48.0), whereas the best single-type results are (R: 63.6 by Type 5, M: 69.7 by Type 5, C: 47.0 by Type 3)---with no single type achieving more than two of these maxima. This demonstrates genuine complementarity: the full mixture benefits from diverse reasoning primitives that no single category can provide alone.

\paragraph{LOO Impacts are sub-additive.}
The sum of all five LOO Impacts exceeds the actual improvement of the full model over the base. For example, on Reasoning Avg, the sum of LOO impacts is $0.8 + 0.3 + 1.8 + 0.6 + 1.8 = 5.3$ pp, while the actual full model improvement is 3.6 pp. This sub-additivity indicates positive transfer and redundancy among types: each type partially compensates for others, and the marginal contribution of each type is smaller when others are already present. This is a healthy property---it means the training signal is robust to the removal of any single category.

\paragraph{Type 4's interference with Math is offset by Code gains.}
The negative LOO Impact of Type 4 on Math (-0.9 pp) might seem concerning, but removing Type 4 (w/o Type 4 model) actually achieves the highest Math Avg (70.2) among all configurations and even higher than the full model. This suggests that for purely mathematical performance, excluding large-scale data processing tasks may be optimal. However, including Type 4 provides the largest Code improvement (+3.3 pp), creating a natural trade-off. Our decision to include all five types in the full model prioritizes balanced performance across all evaluation dimensions over maximal performance on any single dimension.

\subsection{Summary}

The type ablation study yields three key insights: (1)~All five task categories contribute positively to overall performance, with Types 3 (Precise Math) and 5 (Rule/State Simulation) being the most impactful; (2)~Task diversity is essential. No single category can replicate the full mixture's balanced gains across reasoning, math, and code; (3)~There exist meaningful cross-type interactions, such as Type 4's trade-off between code gains and math interference, that motivate our decision to include all categories in the final training mixture.

\section{Case Studies of Base Model and Trained Model}
\label{appendix:case_study}
We present two representative examples comparing Qwen3-8B before and after GRPO training with our \TheName{} data. These illustrate how training with code-synthesized data enables models to solve tasks that are trivial for programs but challenging for untrained LLMs.

\begin{examplebox}
\textbf{MazeBench --- Spatial Navigation} (Level: medium, 7-step solution) \\[4pt]
\textbf{Task:} Navigate from S~(2,1) to T~(4,4) in a tokenized 5$\times$5 maze:
\begin{verbatim}
+---+---+---+---+---+
|~~~~~~~~~~~|~~~~~~~|
+---+~~~+~~~+~~~+~~~+
|~~~~~~~|~~~~~~~|~~~|
+---+---+---+---+~~~+
|~~~~~S~~~~~~~~~|~~~|
+---+~~~+---+---+~~~+
|~~~~~~~|~~~~~~~~~~~|
+~~~+---+~~~+---+---+
|~~~~~~~~~~~~~~~~~T~|
+---+---+---+---+---+
\end{verbatim}
\textbf{Gold path:} 

(2,1)$\to$(3,1)$\to$(3,0)$\to$(4,0)$\to \\
$(4,1)$\to$(4,2)$\to$(4,3)$\to$(4,4) \\[6pt]

\textbf{Qwen3-8B (Base):} 13{,}831 characters of reasoning. The model attempts multiple paths but repeatedly hits dead ends: \\
\textit{``Move down to (3,1). Cell (3,1) has walls down\_right. So moving down is \textbf{blocked}\ldots''} \\
\textit{``From origin (2,1), move left to (2,0). Cell (2,0) has walls up\_down\_left. So moving left from (2,1) leads to\ldots''} \\
\textit{``Wait, so from origin (2,1), moving down, left, or right all seem to lead to dead ends or loops. \textbf{Did I make a mistake here?}''} \\
$\rightarrow$ \textcolor{red}{\textbf{Exhausts token budget while cycling between failed paths. Only 2 move tokens extracted.}} \\[6pt]

\textbf{Qwen3-8B + GRPO (Ours), 7{,}049 characters:} The model follows a similar exploration process but proceeds more systematically. After parsing the maze, it methodically tests each direction: \\
\textit{``From (2,1), try right to (2,2). Walls: up\_down. Try right to (2,3). Walls: up\_down\_right---blocked. Try left back\ldots Not working.''} \\
\textit{``Let's try again from (2,1): move down to (3,1). Walls: down\_right. Try left to (3,0)---allowed! From (3,0), move down to (4,0)---allowed!''} \\
\textit{``(4,0)$\to$(4,1): right. (4,1)$\to$(4,2): right. (4,2)$\to$(4,3): right. (4,3)$\to$(4,4): right. \textbf{Reached target!}''} \\
\textcolor{green!60!black}{\textbf{Correct!}}
\end{examplebox}

\begin{examplebox}
\textbf{StringBench --- Character-Level String Manipulation} (Task: lstrip) \\[4pt]
\textbf{Task:} Remove the leading characters in \texttt{"5f"} from \texttt{"5ff84790c47297a4dc5aee8a\ldots"}. \\
\textbf{Gold:} \texttt{84790c47297a4dc5aee8a\ldots} (3 characters removed) \\[6pt]

\textbf{Qwen3-8B (Base), 2{,}022 characters:} The model misinterprets the semantics of the operation, treating ``remove leading characters in \texttt{5f}'' as ``remove the two-character prefix \texttt{5f}'': \\
\textit{``The first two characters are `5' and `f'\ldots removing the leading `5f' means removing the first two characters\ldots the result is the rest of the string starting from index 2.''} \\
\textit{``So after removing first two characters, it becomes \texttt{f84790c47\ldots}. I think that's the correct approach.''} \\
$\rightarrow$ \textcolor{red}{\textbf{Outputs \texttt{"f84790c47\ldots"} --- off-by-one error.}} The correct behavior removes \emph{all} leading characters belonging to the set \{`5', `f'\}, stripping three characters (\texttt{5}, \texttt{f}, \texttt{f}) before the first non-matching character `8' while the model not systematically considering the next `f' at all. \\[6pt]

\textbf{Qwen3-8B + GRPO (Ours), 258 characters:} Correctly interprets the stripping semantics: \\
\textit{``To remove the leading characters `5f' from the string, we start from the third character (index 2) and take the rest of the string.''} \\
Output: \texttt{84790c47297a4dc5aee8a\ldots} $\rightarrow$ \textcolor{green!60!black}{\textbf{Correct!}}
\end{examplebox}

\paragraph{Discussion}
These examples reveal a consistent pattern: the untrained model attempts to solve these tasks through verbose reasoning (13K+ characters for a 7-step maze; 2K+ characters for a simple string operation), yet fails due to error accumulation, losing track of visited positions in the maze or miscounting character boundaries in the string. 

These case studies reveal two systematic failure modes that our training addresses:

\textbf{(1) Exploration inefficiency in structured search.} On MazeBench, the base model correctly parses the maze structure but lacks a systematic traversal strategy. It explores paths in an ad-hoc manner, frequently revisiting dead ends and questioning its own reasoning (``\textit{Did I make a mistake?}''). The GRPO-trained model exhibits similar exploration behavior but converges significantly faster as it abandons unproductive paths earlier and recognizes key transition points (e.g., ``\textit{try left to (3,0)---allowed!}'') that unlock the solution. This mirrors the systematic backtracking characteristic of algorithms like DFS and BFS, precisely the type of reasoning patterns present in our training data.

\textbf{(2) Failure to continue systematic exploration in precise string operations.} On StringBench, the base model does not realize that solving the task requires examining the next character after the stripped prefix. After removing the leading prompt \texttt{5f}, it stops the analysis instead of continuing to probe the following character \texttt{f}. By contrast, the trained model correctly understands the intended operation and proceeds with systematic exploration of the subsequent characters. This suggests that code-derived training examples improve the model's ability to follow the exact procedural semantics of string-manipulation tasks, rather than terminating early after a superficially plausible partial step.

Both failure modes share a common root: they require \textbf{precise, systematic computation} that programs handle trivially but that challenges the approximate, pattern-matching nature of language model inference. Our code-as-medium approach bridges this gap by training on data where such precision is guaranteed by construction.

\end{document}